\pdfoutput=1
\documentclass[letterpaper]{article} 
\usepackage{aaai23}  
\usepackage{times}  
\usepackage{helvet}  
\usepackage{courier}  
\usepackage[hyphens]{url}  
\usepackage{graphicx} 
\usepackage{natbib}  
\usepackage{caption} 
\usepackage[linesnumbered,ruled,vlined]{algorithm2e}
\SetKwInput{KwInput}{Input}    
\SetKwInput{KwOutput}{Output}    
\SetKwInput{KwParameter}{Parameter} 

\usepackage{placeins}
\usepackage{aecompl}
\usepackage{amsmath,amsfonts}   
\usepackage{bbding}
\usepackage{booktabs}
\usepackage{enumerate}
\usepackage{makecell}
\usepackage{multirow}
\usepackage{pifont}
\usepackage{subfigure}
\usepackage{threeparttable}
\usepackage[T1]{fontenc}
\usepackage{xcolor}
\usepackage{colortbl}

\def\ie{\text{i.e.}}

\def\eg{\text{e.g.}}
\def\etal{\text{et al.}}

\DeclareMathOperator{\st}{s.t.}

\title{PointCA: Evaluating the Robustness of 3D Point Cloud Completion Models Against Adversarial Examples}

\author {
    Shengshan Hu,\textsuperscript{\rm 1,5,6,7,8}
    Junwei Zhang,\textsuperscript{\rm 1,5,6,7,8}
    Wei Liu,\textsuperscript{\rm 1,5,6,7,8}
    Junhui Hou,\textsuperscript{\rm 9}
    Minghui Li,\textsuperscript{\rm 3}\thanks{Corresponding Author.}
    Leo Yu Zhang,\textsuperscript{\rm 10}
    Hai Jin,\textsuperscript{\rm 2,4,5,8}
    Lichao Sun\textsuperscript{\rm 11}
}
\affiliations {
    \textsuperscript{\rm 1} School of Cyber Science and Engineering, Huazhong University of Science and Technology\\
    \textsuperscript{\rm 2} School of Computer Science and Technology, Huazhong University of Science and Technology\\
    \textsuperscript{\rm 3} School of Software Engineering, Huazhong University of Science and Technology\\
    \textsuperscript{\rm 4} Cluster and Grid Computing Lab
    \textsuperscript{\rm 5} National Engineering Research Center for Big Data Technology and System\\
    \textsuperscript{\rm 6} Hubei Engineering Research Center on Big Data Security
    \textsuperscript{\rm 7} Hubei Key Laboratory of Distributed System Security\\
    \textsuperscript{\rm 8} Services Computing Technology and System Lab 
    \textsuperscript{\rm 9} City University of Hong Kong
    \textsuperscript{\rm 10} Deakin University
    \textsuperscript{\rm 11} Lehigh University\\
$\{$hushengshan,jwzh,weiliu73,minghuili,hjin$\}$@hust.edu.cn, jh.hou@cityu.edu.hk, leo.zhang@deakin.edu.au, lis221@lehigh.edu
}

\usepackage{bibentry}

\begin{document}

\maketitle

\begin{abstract}
Point cloud completion, as the upstream procedure of 3D recognition and segmentation, has become an essential part of many tasks such as navigation and scene understanding. While various point cloud completion models have demonstrated their powerful capabilities, their robustness against adversarial attacks, which have been proven to be fatally malicious towards deep neural networks, remains unknown. 
In addition, existing attack approaches towards point cloud classifiers cannot be applied to the completion models due to different output forms and attack purposes.
In order to evaluate the robustness of the completion models, we propose PointCA, \textit{the first adversarial attack against 3D point cloud completion models}. 
PointCA can generate adversarial point clouds that maintain high similarity with the original ones, while being completed as another object with totally different semantic information.
Specifically, we minimize the representation discrepancy between the adversarial example and the target point set to jointly explore the adversarial point clouds in the geometry space and the feature space.
Furthermore, to launch a stealthier attack, we innovatively employ the neighbourhood density information to tailor the perturbation constraint, leading to geometry-aware and distribution-adaptive modification for each point.
Extensive experiments against different premier point cloud completion networks show that PointCA can cause a performance degradation from $77.9\%$ to $16.7\%$, with the structure chamfer distance kept  below $0.01$. We conclude that existing completion models are severely vulnerable to adversarial examples, and state-of-the-art defenses for point cloud classification will be partially invalid when applied to incomplete and uneven point cloud data. 
\end{abstract}

\section{Introduction}\label{introduction}
With the flourishing development of diverse 3D sensors like LiDAR, RADAR, and depth camera, point cloud data are utilized among many safety-critical fields such as
autonomous driving, augmented reality, and robotics. 
Due to its great success in the computer vision area~\cite{he2016deep}, deep learning techniques have been widely applied to point cloud tasks as well. Recent studies show that neural networks are vulnerable to adversarial attacks, where misclassifications can be easily triggered when facing adversarial examples~\cite{goodfellow2014explaining,hu2022protecting}. 
Therefore, more and more research efforts are devoted to exploring the security and robustness of deep learning-based 3D point cloud systems.
\begin{figure}[t]
    \centering
    \includegraphics[width=0.45\textwidth]{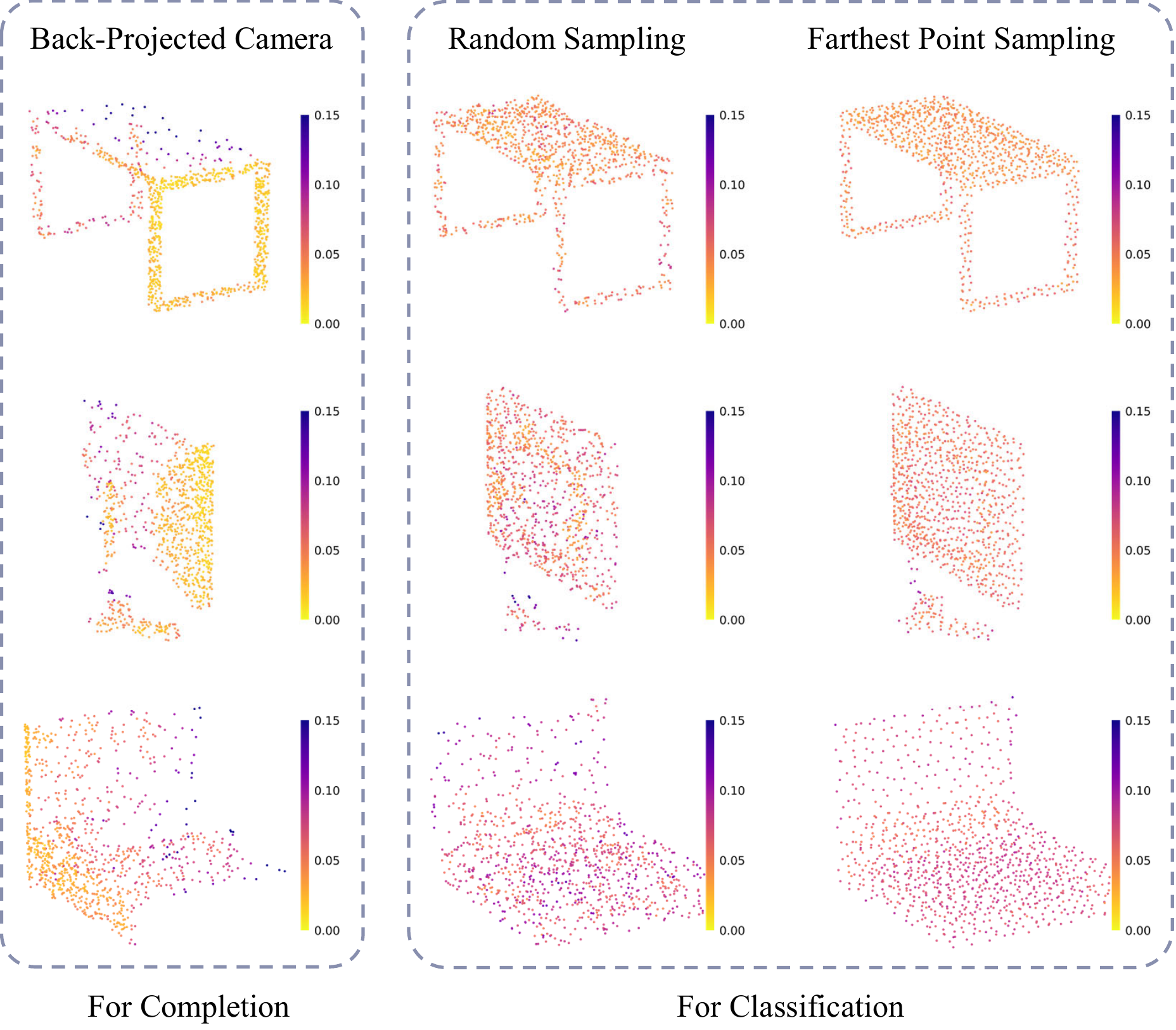}
    \vspace{-0.8em}
    \caption{Geometric distribution of partial point cloud data for completion and complete point cloud data for classification. Partial point clouds have a significantly more complex and locally diverse geometric distribution.}
    \label{fig:camera}
    \vspace{-2em}
\end{figure}


Unfortunately, existing works only concentrate on the point cloud classification  scenario, whereas  point cloud completion, another important task for point cloud systems, has received no attention. 
The completion model is designed to restore the incomplete point cloud data induced by various real-world circumstances, such as limited sensor resolution, and thus it has become a necessary upstream procedure in point cloud processing.

As opposed to point cloud classification, adversarial attack of point cloud completion is more challenging since it requires the manipulation of the geometric shapes rather than the object labels. Specifically, the output of the completion model is an instantial point cloud with semantic and geometric shape, instead of a hard label that directly indicates which class the object belongs to. No classification score or cross-entropy loss can be exploited to generate an adversarial point cloud. 
In other words, the goal of adversarial attacks on the point cloud completion models is to generate a misleading geometric shape rather than a false class label, thus a totally different loss function is needed to measure the similarity between the adversarial example and the target.
Secondly, point cloud classification generally generates adversarial examples over synthetic datasets~\cite{wu20153d}, whose point clouds are complete and uniform with continuous geometric manifolds, whereas the incomplete partial point clouds in the completion tasks have a more realistic and uneven geometric distribution caused by the occlusion and limited sensor viewing angle.
To verify this, Fig.~\ref{fig:camera} visualizes the difference in geometric distributions of the input point clouds from two tasks, assigning a score to each point depending on the density of its surrounding region through Eq.~(\ref{eq:local density}). The data distribution is more fluctuating and inconsistent in partial point clouds. How to efficiently and stealthily generate adversarial examples on more nonuniformly distributed data is a challenging problem that has been ignored by previous works in the classification, but is crucial for point cloud completion attacks.


\begin{figure*}[t]
    \centering
    \includegraphics[width=0.95\textwidth]{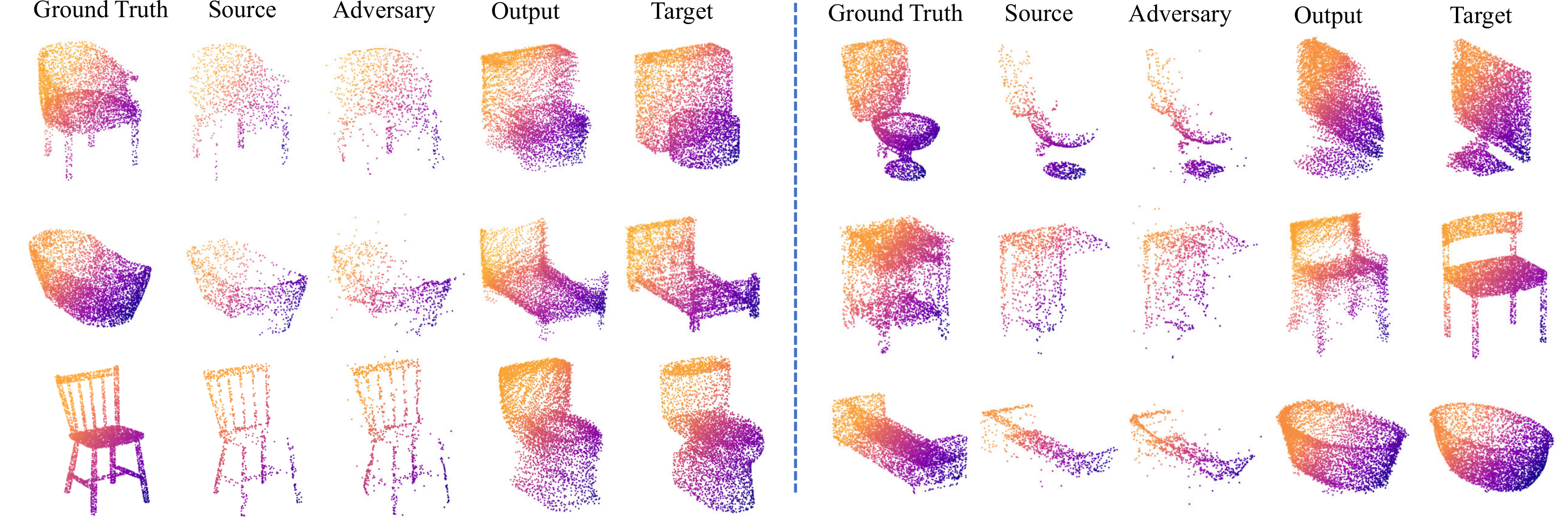}
    \caption{An illustration of targeted attack results. \textit{Source} is the partial point cloud generated on \textit{Ground Truth} from one view point. Tiny perturbations are added to \textit{Source} to obtain the \textit{Adversary}, whose completion \textit{Output} is very close to the \textit{target}.}
    \label{fig:illustration}
    \vspace{-1em}
\end{figure*}

In this paper, we propose PointCA, the first adversarial attack on 3D point cloud completion models to systematically evaluate their robustness. As shown in Fig.~\ref{fig:illustration}, the adversarial example generated by PointCA is visually similar to the original one, but completed as another target object with totally different semantic information. 
To overcome the above challenges, PointCA first explores the similarity measurement in the geometry space and the feature space respectively, formulating the adversarial example generation  as an appropriate optimization problem which can be solved by the gradient-based algorithm.
In addition, considering the unevenness and incompleteness of the partial point cloud data, we define and allocate local geometric neighborhood for each input point through the \textit{k-Nearest Neighbor} ($k$NN) algorithm and evaluate the distribution density in the neighbor point sets, based on which the perturbation constraint will be adaptively tailored for each point to achieve a more imperceptible and efficient attack.

To summarize, the contributions of our work are as follows:
\begin{itemize}
\item We propose PointCA, \textit{the first adversarial attack on 3D point cloud completion models}. By investigating the characteristics of the geometry space and the feature space, an appropriate optimization problem is formulated to measure the similarity between geometric shapes and find adversarial examples.

\item We innovatively employ the neighborhood density information to tailor a perturbation constraint and design the geometry-aware and distribution-adaptive modification for each point to achieve a stealthy attack under the locally diverse geometric distribution of partial point clouds. 

\item Our experiments show that PointCA can cause at least 60\% performance degradation over different completion models, while keeping a low perceptibility of adversarial perturbations. We verify that the state-of-the-art defense methods based on statistic outlier removal cannot fully guard the point cloud completion model.
\end{itemize}

\section{Related Work}\label{related}

\subsection{Attacks on Point Cloud Classification}
In the point cloud domain, Xiang \etal~\shortcite{xiang2019generating} proposed the first point cloud attack algorithm based on the C\&W framework~\cite{carlini2017towards}.
Hamdi \etal~\shortcite{hamdi2020advpc} then employed an autoencoder loss to strengthen the transferability of the attack across multiple  classification models.
To reduce the perceptibility of adversarial point clouds, Kim \etal~\shortcite{kim2021minimal} and Shi \etal~\shortcite{ShiCX0YH22} explored perturbing only a minimal number of points and preserving the original geometric shape as much as possible.
LG-GAN was introduced by Zhou \etal~\shortcite{zhou2020lg} for a more efficient and flexible point cloud attack.


Although widely studied in the literature, existing adversarial attacks focus on point cloud classification tasks, which mainly concentrate on the artificial data in virtual scenarios and ignore many realistic properties of point clouds such as nonuniform distribution and structure damage~\cite{sun2022benchmarking,ren2022benchmarking}, making them difficult to apply to point completion tasks.

\subsection{Attacks on Point Cloud Generation}
Attacks on generative models have also received attention~\cite{kos2018adversarial,willetts2021improving}.
GeoAdv~\cite{lang2021geometric} first launched the attack against point cloud reconstruction model based on autoencoder structure. However, our work is significantly different from it.
Compared with reconstruction, point cloud completion is a much more different and difficult task. The completion model is not explicitly enforced to retain the input in its output like an autoencoder. Instead, it needs to infer the complete structure from the partial observation with less prior knowledge. 
Besides, the higher data complexity in point cloud completion induces a higher standard of perturbation constraint (Fig.~\ref{fig:camera}). 
In order to prevent excessive perturbations that do not match the local geometry manifolds and easily recognizable attacks, we create the Adaptive Geometric Constraint for each point rather than employing a globally consistent constraint threshold. 

\subsection{Defenses on Point Clouds}
Disparate methods have been developed to defend against adversarial attacks for point cloud classification. Among existing defenses including adversarial training, Gaussian noise perturbation~\cite{yang2019adversarial}, certified robustness~\cite{liu2021pointguard}, and point removal~\cite{liu2019extending}, the \textit{statistical outlier removal} (SOR)~\cite{zhou2019dup} achieves the best performance due to its effectiveness and efficiency.  Based on the fact that perturbed points in attacks are likely to become outliers off the manifold of the point cloud surface, SOR can remove adversarial points in a statistical manner.
As shown in our experiments (Table~\ref{tab:defense}), although SOR performs well in the classification task, it fails to fully guarantee robustness when facing incomplete partial point clouds.

\section{Methodology}
\label{method}

\subsection{Problem Formulation}
\label{sec:preliminaries}

\textit{Point cloud} (PC) completion models aim to predict the complete structure of a given incomplete input. Generally, a well-trained PC completion model can provide a one-way geometric mapping: $\mathbf{X}^P\in\mathbb{R}^{m\times3}\rightarrow\mathbf{X}\in\mathbb{R}^{n\times3}$, where $\mathbf{X}^P$ represents a partial point cloud, usually captured by the 3D sensor from a single observation angle of a real object. $\mathbf{X}$ is the ground-truth point cloud of $\mathbf{X}^P$, which can be considered as the point set uniformly scanned from the whole surface of the same original object associated with $\mathbf{X}^P$. 
An effective PC completion model $f_{\theta}$ should satisfy:
\begin{equation}\label{eq:Evaluation}
\mathit{d}\left(f_{\theta}\left(\mathbf{X}^P\right), \mathbf{X}\right)\leq  \gamma,
\end{equation}
where $d(S_1, S_2)$ represents an appropriate metric that measures the discrepancy between two point clouds, $\gamma$ is the evaluating threshold with a small value.

In this paper, we propose PointCA, the first adversarial attack towards point cloud completion models. The goal of PointCA is to delicately construct an adversarial version of $\mathbf{X}^P$, denoted as $\mathbf{X}^{P^{\prime}}$, which can lead to a false completion result from the PC completion model $f_{\theta}$. Formally, we have:
\begin{equation}\label{eq:Attack Evaluation}
\mathit{d}\left(f_{\theta}\left(\mathbf{X}^{P^{\prime}}\right),\mathbf{Y}\right)\leq  \gamma,
\end{equation}
where $\mathbf{Y}$ is the target point cloud from a different category.

According to the knowledge we know about the representations of the source and target point clouds, PointCA will be investigated in the geometry space and the latent space, respectively. Here, we denote them as Geometry PointCA and Latent PointCA.


\subsection{Geometry PointCA}
\label{sec:Formulating PC Completion Attack}

A straightforward solution for solving Eq.~(\ref{eq:Attack Evaluation}) is the brute-force search for $\mathbf{X}^{P^{\prime}}$, which is apparently infeasible in practice. We thus reformulate the  problem as a rational optimization instance that can be efficiently solved by existing optimization algorithms. We denote the adversarial perturbation added to the original partial input as $\boldsymbol{\delta} = \mathbf{X}^{P^{\prime}}-\mathbf{X}^{P}$. Therefore Eq.~(\ref{eq:Attack Evaluation}) can be formulated as:
\begin{equation}\label{eq:basic attack}
\mathop{\arg\min}\limits_{\boldsymbol{\delta}}\;
\mathit{d}\left(f_{\theta}\left(\mathbf{X}^P + \boldsymbol{\delta}\right), \mathbf{Y}\right), \enspace 
\st \;\left\| \boldsymbol{\delta} \right\|_{p} \leq  \epsilon,
\end{equation}
where $\left\| \cdot \right\|_{p}$ is a distance metric to measure the perceptibility of adversarial perturbation. Normally, $\left\| \cdot \right\|_{p}$ is instantiated as $\ell_{p}$-norm $(p\in\{1,2,\infty\})$.



In Geometry PointCA, we assume that the adversary knows the exact ground-truth of the target point cloud. As most  PC completion networks are built on the encoder-decoder structure~\cite{yang2018foldingnet,yuan2018pcn,xie2020grnet,zhang2020detail}, the model $f_\theta$ can be formulated as: 
$f_\theta(\cdot) = De(En(\cdot))$. 
Therefore the optimized similarity loss 
based on geometry information can be defined as:
\begin{equation}\label{eq:geometry similarity loss}
\begin{aligned}
\mathit{d}_{similarity} = 
\mathit{D}_{chamfer}\left(De\left(En\left(\mathbf{X}^{P} +\boldsymbol{\delta}\right)\right),\mathbf{Y}\right),
\end{aligned}
\end{equation}
where $En$ is the encoder of the completion model, $De$ is the decoder part, $Y$ is the target complete point cloud. ${D}_{chamfer}$ represents the Chamfer distance metric. 

In summary, the adversarial perturbations in Geometry PointCA are generated according to the Chamfer distance between the output completed geometry structures of the adversarial examples and the target point clouds. More details on point cloud distance metrics are attached in the supplementary.

\subsection{Latent PointCA}
In some strict scenarios, the point clouds of real-world objects suffer from diverse corruptions ~\cite{sun2022benchmarking}, where we cannot obtain complete and detailed ground-truth samples~\cite{yu2021pointr}. It is necessary to further explore how to generate adversarial examples that get rid of the knowledge of the target complete point cloud data.
Hence we propose the Latent PointCA where the adversary only obtains the partial target point cloud. 

Inspired by variational autoencoder attacks and the latent variable model~\cite{kos2018adversarial}, inputs whose distribution are similar to each other in the representation of the latent layer will also get similar outputs with a high probability~\cite{hu2022Badhash}.
Instead of using the target point data straightly in Eq.~(\ref{eq:geometry similarity loss}), we leverage the similarity metric in the feature space and regard the learned latent feature vector as an approximate representation of the shape manifold. To launch this latent-based attack, we combine Euclidean distance~\cite{kos2018adversarial} and Distribution distance~\cite{willetts2021improving} to measure the difference between features and reformulate the optimized similarity loss as: 
\begin{equation}\label{eq:latent similarity loss}
\begin{aligned}
\mathit{d}_{similarity}&=
\left\|En(\mathbf{X}^{P} +\boldsymbol{\delta}) - En(\mathbf{Y}^{P})\right\|_{2} \\
&+\lambda D_{KL}\left(En(\mathbf{Y}^{P})\|En(\mathbf{X}^{P} +\boldsymbol{\delta})\right),
\end{aligned}
\end{equation}
where $\mathbf {Y}^{P}$ is the partial point cloud in the target category different from $\mathbf {X}^{P}$, $\ell_{2}$-norm $\left\| \cdot \right\|_{2}$ and Kullback–Leibler divergence $D_{KL}(\cdot\|\cdot)$ are jointly used as the similarity metric between the features of inputs, $\lambda$ is the hyperparameter.

\subsection{Adaptive Geometric Constraint}
Note that the optimation formula in Eq.~(\ref{eq:basic attack}) has an important perturbation constraint: $\left\| \boldsymbol{\delta} \right\|_{p}$. 
It is obvious that a smaller perturbation metric $\left\| \boldsymbol{\delta} \right\|_{p}$ leads to a stealthier attack. In this paper, we design a new method to provide adaptive constraints for PointCA to further improve the imperceptibility of adversarial point clouds and the attack efficiency. Here we first describe the intuition behind the adaptive constraint  and then give the detailed algorithm.

\noindent\textbf{Intuition.}
For traditional 2D images, adversarial attacks usually use $\ell_{\infty}$-norm to limit the variation degree on each channel, which clips the redundant perturbation to limit the maximum value  as:
$\left\| \boldsymbol{\delta} \right\|_{\infty}\leq\epsilon$.
Similarly, Hamdi \etal~\shortcite{hamdi2020advpc} and Liu \etal~\shortcite{liu2019extending} inherit this strategy to limit the perturbation on each 3D coordinate in point cloud adversarial attack as:
\label{sec:vanlia channelwise epsilon}
\begin{equation}\label{eq:vanlia channelwise epsilon}
\begin{aligned}
-\,\epsilon \leq \delta_{i,j} \leq \epsilon,\quad
i={1,\cdots, m},\quad j={1,2,3},
\end{aligned}
\end{equation}
where $i$ is the sequence number of the point $x_{i}$  with the size of $m$, $j$ is the coordinate parameter corresponding to one element of $(x,y,z)$. 
Due to the highly structured property of point clouds, JGBA~\cite{ma2020efficient} considered the spatial geometry  and designed a pointwise limitation to clip the perturbation amplitude in the Euclidean space: $\left\| \delta_{i} \right\|_{2}\leq \epsilon$. 

Nevertheless, these works still treat all the points equally with the same restricting threshold. According to recent studies~\cite{sun2021local} and our experiments shown in Fig.~\ref{fig:comparasion}, the distributions of points at different portions are distinct, not all points are equally essential in the optimization for the construction of an adversarial point cloud. The same limitation could be excessively strict for certain points while being insufficient for others. Additionally, after applying too much perturbation, points in uniformly and tightly dispersed areas are more likely to produce outliers that do not match the local geometry, lowering the imperceptibility of adversarial point clouds~\cite{zhou2019dup}.

In light of this, instead of treating each independent point identically, we explore the local geometric relationship of points to tailor adversarial examples on partial point clouds.


\begin{figure}[t]

\centering   
\includegraphics[width=0.35\textwidth]{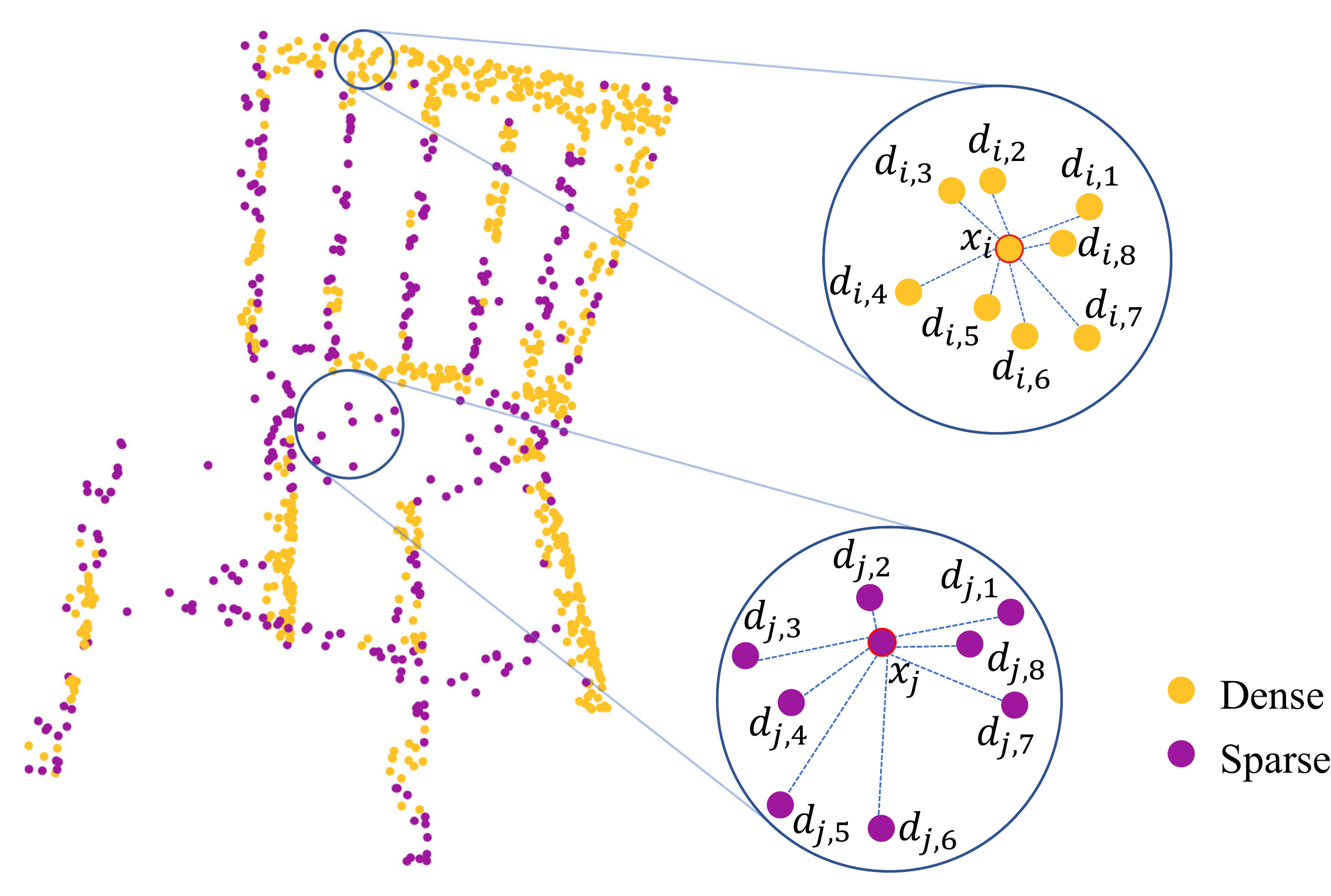}   

\vspace{-1em}
\caption{The illustration of local neighborhood geometry. For ease of presentation, we simply partition the points into two categories according to each point's score $\rho$. Points whose score $\rho$ is below a threshold $T$ are marked in yellow, otherwise marked in purple. 
The purple centroid point $x_{j}$ has a sparse local neighborhood, and the yellow centroid point $x_{i}$ is located in a dense local neighbor points set.}
\label{fig:local geometry}
\vspace{-1em}
\end{figure}


\begin{algorithm}[t] 
\label{alg1}
\caption{Point Cloud Completion Attack}
\KwInput{Model $f_\theta$; benign partial PC $\mathbf{X}^P$; target partial PC $\mathbf{Y}^P$ or target complete PC $\mathbf{Y}$. }
\KwParameter{Iterations $n$; nearest neighbor range $k$; step size $\beta$.}
\KwOutput{Completion adversarial example $\mathbf{X}^{P^{\prime}}$.}

Initialize $\mathbf{X}^{P^{\prime}}$ with random noise $\boldsymbol{\delta}$.\\

Obtain the local neighbor point set $\mathcal{S}(x_{i},\,k)$ that each point $x_{i}$ in $\mathbf{X}^P$ has by $k$NN;\\


Compute the geometric adaptive restraint $\epsilon_{i}$ for each point $x_{i}$ through Eq.~(\ref{eq:adaptive epsilon});

\For{$q = 1$ to $n$ }
{
    \If {Geometry PointCA}
    { 
        calculate ${d}_{similarity}$ through Eq.~(\ref{eq:geometry similarity loss})
    }
    \ElseIf {Latent PointCA}
    { 
        calculate ${d}_{similarity}$ through Eq.~(\ref{eq:latent similarity loss})
    }

    Compute the gradient $\Delta= \nabla_{\mathbf{X}^{P^{\prime}}}{d}_{similarity}$;

    Update the point cloud as $\mathbf{X}^{P^{\prime}}\leftarrow \mathbf{X}^{P^{\prime}} -\beta \cdot sign(\Delta)$;
    
    Compute the overall perturbation $\boldsymbol{\delta} \leftarrow \mathbf{X}^{P^{\prime}} - \mathbf{X}^{P}$;
    
    For each point in $\mathbf{X}^{P^{\prime}}$:
    
    \If {$\left\| {\delta}_{i} \right\|_{2} > \epsilon_{i}$}
    {
    Clip the ${\delta}_{i}\leftarrow \delta_{i}\cdot\frac{{\epsilon}_{i}}{\left\|{\delta}_{i} \right\|_{2}}$;
    }
    Update the point cloud as $\mathbf{X}^{P^{\prime}}\leftarrow \mathbf{X}^{P}+\boldsymbol{\delta}$ ;
}
\textbf{Return} Adversarial partial point cloud $\mathbf{X}^{P^{\prime}}$.
\end{algorithm}

\noindent\textbf{Local geometric density.}
To give a more precise analysis of the geometric structure, we partition the local neighborhood set for each point through $k$NN algorithm. Let $\mathcal{S}(x_{i},\,k)$ denote the local $k$NN point set of point $x_{i}$ in partial point cloud $\mathbf{X}^{P}$. The pairwise distance $\mathit{d}_{i,l}$ between the set centroid point $x_{i}$ and its local neighbor point $x_{l}$ is defined as:
\begin{equation}\label{eq:Local neighbor distance}
\begin{aligned}
\mathit{d}_{i,l} = \left\| x_{i} - x_{l} \right\|_{2},
\;{x_{l}\in\mathcal{S}\left(x_{i},\,k\right)},
\quad{l} = 1,\cdots,k.
\end{aligned}
\end{equation}

We measure the \textit{distribution sparsity} of different local point sets $\mathcal{S}(x_{i},\,k)$ by calculating the average pairwise distance:
\vspace{-0.5em}
\begin{equation}\label{eq:Local average distance}
\begin{aligned}
\mathit{d}_{i} = \frac{1}{k}\,\sum\limits_{l=1}^{k}\,{d}_{i,l}\;, \quad{i} = 1,\cdots,m.
\end{aligned}
\end{equation}

Meanwhile, we evaluate the \textit{distribution uniformity} of the local point set $\mathcal{S}(x_{i},\,k)$ by the standard deviation of $\mathit{d}_{i,l}$\;:
\vspace{-0.5em}
\begin{equation}\label{eq:standard deviation}
\begin{aligned}
{\sigma}_{i} = \sqrt{\frac{1}{k-1}\,\sum\limits_{l=1}^{k}\,
\left({d}_{i,l} - {d}_{i}\right)^{2}}\;.
\end{aligned}
\vspace{-0.5em}
\end{equation}

These two metrics together constitute a description score $\rho$ about the local point cloud density for each point:
\vspace{-0.5em}
\begin{equation}\label{eq:local density}
\begin{aligned}
\rho_{i} =\mathit{d}_{i} + t \cdot \sigma_{i},
\end{aligned}
\vspace{-0.5em}
\end{equation}
where $t$ is an auxiliary parameter that further refines the constraint with local uniformity information. Fig.~\ref{fig:local geometry} gives a brief illustration for the local geometry density. 

\noindent\textbf{Adaptive constraint.}
Based on the above analysis of local geometric density, we can jointly exploit the \textit{sparsity} and \textit{uniformity} to customize an adaptive geometric perturbation threshold $\epsilon_{i}$ independently for each point:
\vspace{-0.5em}
\begin{equation}\label{eq:adaptive epsilon}
\begin{aligned}
\epsilon_{i} &=\eta\,\cdot \rho_{i} \\
&=
\frac{\eta}{k}\,\sum\limits_{l=1}^{k}\,{d}_{i,l} + \eta \cdot t \cdot \sqrt{\frac{1}{k-1}\,\sum\limits_{l=1}^{k}\,
\left({d}_{i,l} - {d}_{i}\right)^{2}}\,,\\
&\quad{i} = 1,\cdots,m,\;
\end{aligned}
\vspace{-0.5em}
\end{equation}
where $\eta$ is a scaling coefficient to set a flexible perturbation  limitation. As a result, the  adaptive geometric constraint will attach a larger perturbation to points which originally have a sparse distribution, and the points with a denser local neighborhood will be perturbed slightly. Thus the adversarial point cloud can maintain a similar 
distribution and fit the surface manifold of the clean partial point cloud better.

Finally, combining with the adaptive geometric constraint, Geometry PointCA is reformulated as:
\begin{equation}\label{eq:final outspace attack}
\begin{aligned}
\mathop{\arg\min}\limits_{\boldsymbol{\delta}}\;
&\mathit{D}_{chamfer}\left(De\left(En\left(\mathbf{X}^{P} +\boldsymbol{\delta}\right)\right),\mathbf{Y}\right),\\
&\st \;\left\| {\delta}_{i} \right\|_{2} \leq  \epsilon_{i}, \quad{i} = 1,\cdots,m,\;
\end{aligned}
\end{equation}
\noindent and Latent PointCA is reformulated as:
\begin{equation}\label{eq:final latent attack}
\begin{aligned}
\mathop{\arg\min}\limits_{\boldsymbol{\delta}}\;
&\left\|En\left(\mathbf{X}^{P} +\boldsymbol{\delta}\right)- En\left(\mathbf{Y}^{P}\right)\right\|_{2}\\
&+\lambda D_{KL}\left(En(\mathbf{Y}^{P})\|En(\mathbf{X}^{P} +\boldsymbol{\delta})\right),\\
&\enspace\st \;\left\| \delta_{i} \right\|_{2} \leq \epsilon_{i},\quad{i} = 1,\cdots,m.\;
\end{aligned}
\end{equation}

Note that we use an iterative gradient-based strategy to minimize these two similarity losses and update the perturbation $\boldsymbol{\delta}$. A complete description of our point cloud completion attack is shown in Algorithm 1.

\section{Experiments}\label{experiment}

\subsection{Experimental Setup}
\label{sec:Experimental Setup}

\noindent\textbf{Dataset and victim completion models.} According to prevailing methods~\cite{xie2021style,wang2021unsupervised}, we employ the back-projected depth camera~\cite{yuan2018pcn} to create partial point clouds on Modelnet10~\cite{wu20153d} because of the dataset‘s excellent versatility in various point cloud tasks.
Four well-trained point cloud completion models: PCN~\cite{yuan2018pcn}, RFA(Zhang \etal 2020), GRNet~\cite{xie2020grnet}, and VRCNet~\cite{pan2021variational} are the target models of our attacks.
More detailed dataset setting, model training parameters, and completion evaluations appear in our supplementary.

\begin{table*}[t]
\centering\Large
\renewcommand\arraystretch{1.16}
\resizebox{\textwidth}{!}{%
\begin{tabular}{cccccccccccccc}
\toprule[1.5pt]
\multirow{2}{*}{\textbf{Method}} & \multirow{2}{*}{$\boldsymbol{\eta}$} & \multicolumn{3}{c}{\textbf{PCN}} & \multicolumn{3}{c}{\textbf{RFA}} & \multicolumn{3}{c}{\textbf{GRNet}} & \multicolumn{3}{c}{\textbf{VRCNet}} \\ \cmidrule(r){3-5} \cmidrule(r){6-8} \cmidrule(r){9-11} \cmidrule(r){12-14}
 &  & $\text{T-RE}_{emd}$ & $\text{T-RE}_{cd}$ & $\text{T-NRE}_{cd}$ & $\text{T-RE}_{emd}$ & $\text{T-RE}_{cd}$ & $\text{T-NRE}_{cd}$ & $\text{T-RE}_{emd}$ & $\text{T-RE}_{cd}$ & $\text{T-NRE}_{cd}$ & $\text{T-RE}_{emd}$ & $\text{T-RE}_{cd}$ & $\text{T-NRE}_{cd}$ \\ \hline
\multirow{3}{*}{Random Noise} 
 & 1.5 & 0.1837 & 0.0560 & 4.4062 & 0.2098 & 0.0538 & 4.8919 & 0.1819 & 0.0545 & 4.6387 & 0.1721 & 0.0527 & 4.8132 \\
 & 2.5 & 0.1837 & 0.0558 & 4.3914 & 0.2103 & 0.0535 & \textbf{4.8705} & 0.1824 & 0.0542 & 4.6185 & 0.1724 & 0.0525 & 4.7957 \\
 & 5 & 0.1837 & 0.0558 & \textbf{4.3875} & 0.2112 & 0.0536 & 4.8782 & 0.1825 & 0.0542 & \textbf{4.6171} & 0.1724 & 0.0524 & \textbf{4.7889} \\ \hline
\multirow{3}{*}{Classification Noise}
 & 1.5 & 0.1815 & 0.0578 & 4.5446 & 0.2001 & 0.0568 & 5.1521 & 0.1819 & 0.0576 & 4.8934 & 0.1675 & 0.0555 & \textbf{5.0684} \\
 & 2.5 & 0.1815 & 0.0578 & 4.5437 & 0.1998 & 0.0567 & \textbf{5.1467} & 0.1819 & 0.0576 & 4.8932 & 0.1675 & 0.0556 & 5.0685 \\
 & 5 & 0.1815 & 0.0578 & \textbf{4.5427} & 0.1999 & 0.0567 & 5.1484 & 0.1819 & 0.0575 & \textbf{4.8912} & 0.1675 & 0.0555 & 5.0696 \\ \hline
\multirow{3}{*}{Geometry PointCA} 
 & 1.5 & 0.1255 & 0.0233 & 1.8129 & 0.1588 & 0.0246 & 2.2379 & 0.1342 & 0.0228 & 1.9423 & 0.1145 & 0.0248 & 2.2638 \\
 & 2.5 & 0.1186 & 0.0195 & 1.5174 & 0.1540 & 0.0211 & 1.9159 & 0.1283 & 0.0203 & 1.7253 & 0.1080 & 0.0220 & 2.0085 \\
 & 5 & 0.1148 & 0.0175 & \textbf{1.3552} & 0.1506 & 0.0192 & \textbf{1.7430} & 0.1257 & 0.0193 & \textbf{1.6381} & 0.1057 & 0.0211 & \textbf{1.9242} \\ \hline
\multirow{3}{*}{Latent PointCA}
 & 1.5 & 0.1459 & 0.0331 & 2.5523 & 0.1873 & 0.0378 & 3.4212 & 0.1762 & 0.0471 & 4.0215 & 0.1214 & 0.0313 & 2.8253 \\
 & 2.5 & 0.1372 & 0.0270 & 2.0651 & 0.1815 & 0.0333 & 2.9960 & 0.1760 & 0.0465 & 3.9701 & 0.1078 & 0.0255 & 2.2765 \\
 & 5 & 0.1318 & 0.0231 & \textbf{1.7449} & 0.1750 & 0.0296 & \textbf{2.6490} & 0.1761 & 0.0463 & \textbf{3.9582} & 0.0988 & 0.0220 & \textbf{1.9425} \\
 \bottomrule[1.5pt]
\end{tabular}%
}
\vspace{-0.5em}
\caption{Results of attack methods under different perturbation constraints $\eta$. A smaller value indicates a better attack effect. 
}
\vspace{-0.5em}
\label{tab:performance}
\end{table*}

\noindent\textbf{Evaluation metrics.}
PointCA aims to alter the model's output to a target geometric shape. In order to evaluate the attack performance,  we thus use  \textit{target reconstruction error} T-RE = $d(f_\theta(\mathbf{X}^{P^{\prime}}),\mathbf{Y})$ to measure the similarity between outputs and targets, where $d(\cdot)$ indicates \textit{chamfer distance} (CD) or \textit{earth mover's distance} (EMD)~\cite{rubner2000earth}. Considering the completion models usually have an inherent error, we also investigate the relative attack effect through \textit{target normalized reconstruction error}~\cite{lang2020samplenet} as:
\vspace{-0.5em}
\begin{equation}\label{eq:t-nre}
\begin{aligned}
\text{T-NRE} = \frac{d(f_\theta(\mathbf{X}^{P^{\prime}}),\mathbf{Y})}{d(f_\theta(\mathbf{Y}^{P}),\mathbf{Y})}.
\end{aligned}
\vspace{-0.5em}
\end{equation}

If $\text{T-NRE}=1$, the attack effect of the adversarial point cloud can be roughly equated to that of the target object's original partial point cloud.
 
Besides, the \textit{perturbation budget} $d(\mathbf{X}^{P^{\prime}},\mathbf{X}^{P})$ and \textit{outliers number} under SOR algorithm~\cite{zhou2019dup} are calculated to further evaluate the stealthiness of our method. 

For defense countermeasure assessments on our adversarial point clouds, we exploit the \textit{source reconstruction error} S-RE = $d(f_\theta(\mathbf{X}^{P^{\prime}}),\mathbf{X})$ and \textit{source normalized reconstruction error}~\cite{lang2021geometric}:
\vspace{-0.5em}
\begin{equation}\label{eq:s-nre}
\begin{aligned}
\text{S-NRE} = \frac{d(f_\theta(\mathbf{X}^{P^{\prime}}),\mathbf{X})}{d(f_\theta(\mathbf{X}^{P}),\mathbf{X})}.
\end{aligned}
\vspace{-0.5em}
\end{equation}

\subsection{Attack Performance}
\label{sec:Attack Performance}
\textbf{Implementation details.} 
Different from the object-to-label attack in the classification, the adversarial attack in point cloud completion is object-to-object. One target class may include many different attacking objects for each source example, which results in an enormous expense. Similar to previous research~\cite{lang2021geometric}, for each object class, we randomly select 20 point cloud pairs in the test set as source examples and each source pair will attack the other 9 object classes. For a given source pair and target label, we take 5 pairs from the target class whose ground truths are top-5 geometric neighbors nearest with source's in terms of Chamfer distance. To sum up, there are 20$\times$10$\times$9$\times5$=9000 source-target pairs in PointCA. All the evaluation metrics are calculated on the average of these 9000 attacks. 

\noindent\textbf{Baselines.} 
\textit{Random noise}: We add random Gaussian noise to the input partial point clouds and investigate the outputs to exclude the interference of general noise. 
\textit{Classification noise}: We also verify whether the adversarial noises created on classifiers can be transferred to attack point cloud completion models.
More attack and baseline setting details are reported in our supplementary.


\begin{table*}[t]
\centering
\renewcommand\arraystretch{1.2}
\resizebox{\textwidth}{!}{%
\begin{tabular}{ccccccccccccccc}
\toprule[1.5pt]
\textbf{Classifier} & &\multicolumn{4}{c}{\textbf{PointNet}} & \multicolumn{4}{c}{\textbf{PointNet++}} & \multicolumn{4}{c}{\textbf{DGCNN}} \\ 
\cmidrule(r){3-6} \cmidrule(r){7-10} \cmidrule(r){11-14}
Completion Model & & PCN & RFA & GRNet & VRCNet & PCN & RFA & GRNet & VRCNet & PCN & RFA & GRNet & VRCNet \\ \hline
Benign Completion Output & ACC & 78.00 & 77.50 & 79.44 & 79.96 & 77.94 & 64.61 & 75.01 & 80.26 & 71.00 & 61.00 & 67.51 & 67.07 \\ \hline
\multirow{2}{*}{\makecell[c]{Geometry PointCA \\ ($\boldsymbol{\eta}$=5)}} & ACC$(\downarrow)$ & 18.31 & 18.54 & \textbf{21.30} & 31.80 & 16.72 & 19.93 & \textbf{20.68} & 31.69 & 16.73 & 17.37 & \textbf{19.36} & 27.29 \\
 & ASR$(\uparrow)$ & 42.48 & 35.40 & \textbf{33.61} & 28.49 & 43.16 & 28.51 & \textbf{31.01} & 27.76 & \textbf{35.61} & 23.30 & \textbf{24.72} & 20.78 \\ \hline
\multirow{2}{*}{\makecell[c]{Latent PointCA \\($\boldsymbol{\eta}$=5)}} & ACC$(\downarrow)$ & \textbf{16.43} & \textbf{16.51} & 41.81 & \textbf{12.87} & \textbf{15.14} & \textbf{17.29} & 36.00 & \textbf{11.71} & \textbf{15.49} & \textbf{14.42} & 30.91 & \textbf{13.87} \\
 & ASR$(\uparrow)$ & \textbf{44.54} & \textbf{39.30} & 14.56 & \textbf{50.67} & \textbf{46.67} & \textbf{29.59} & 12.48 & \textbf{50.71} & 35.14 & \textbf{23.79} & 12.48 & \textbf{37.00} \\ 
 \bottomrule[1.5pt]
\end{tabular}%
}
\vspace{-0.5em}
\caption{Semantic evaluation of the outputs reconstructed on our adversarial point clouds. Lower ACC ($\downarrow$) and higher ASR ($\uparrow$) indicate stronger semantic information of our attack.}
\label{tab:semantic}
\vspace{-0.5em}
\end{table*}

\noindent\textbf{Analysis.}
The detailed attack results are exhibited in Table~\ref{tab:performance}. 
Firstly, the low results under three scaling coefficients $\eta$ imply a satisfactory attack performance of our geometry attack and latent attack against all four completion models. The overall attack performance of geometry PointCA is even close to the completion effect of the original clean samples. Secondly, the high T-NRE of \textit{random noise} shows that general noises cannot easily disturb the completion procedure towards targets. Thirdly, the adversarial perturbation created on classification model is useless when attacking completion models. Because the purpose of classification adversarial attack is merely to change the output's label, in most cases, the adversarial point clouds 
still resemble the original structures, making it impossible for them to successfully deceive other point cloud processing missions.


\noindent\textbf{Relative attack success rate.}
Geometric distance metrics can be employed to measure the attack effectiveness of adversarial examples. Nevertheless, there is no precise definition of \textit{attack success rate} (ASR) in our point cloud completion attacks. Therefore, we leverage T-NRE as a reference to dynamically assess the \textit{relative attack success rate} (Relative-ASR). We set a success threshold $\tau$, 
if one attack's T-NRE is below this threshold, we can roughly consider the attack is successful. Then the overall Relative-ASR can be calculated on these 9000 source-target pairs.

\begin{figure}[t] 
\centering
\subfigure[Relative-ASR on PCN] 
{
    \label{RASR1}
	\centering         
	\includegraphics[width=0.22\textwidth]{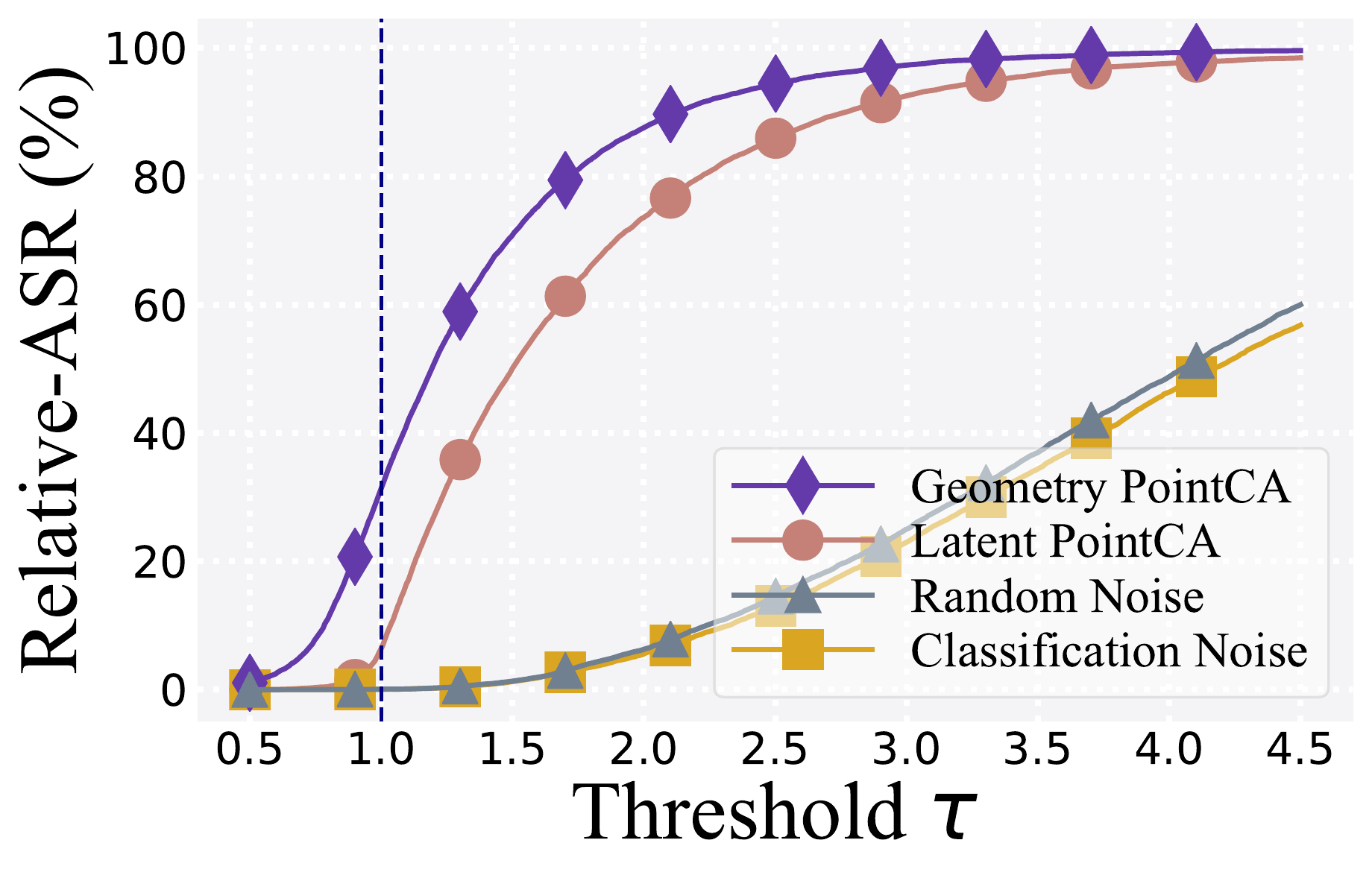}   
}
\subfigure[Relative-ASR on RFA] 
{
    \label{RASR2}
	\centering     
	\includegraphics[width=0.22\textwidth]{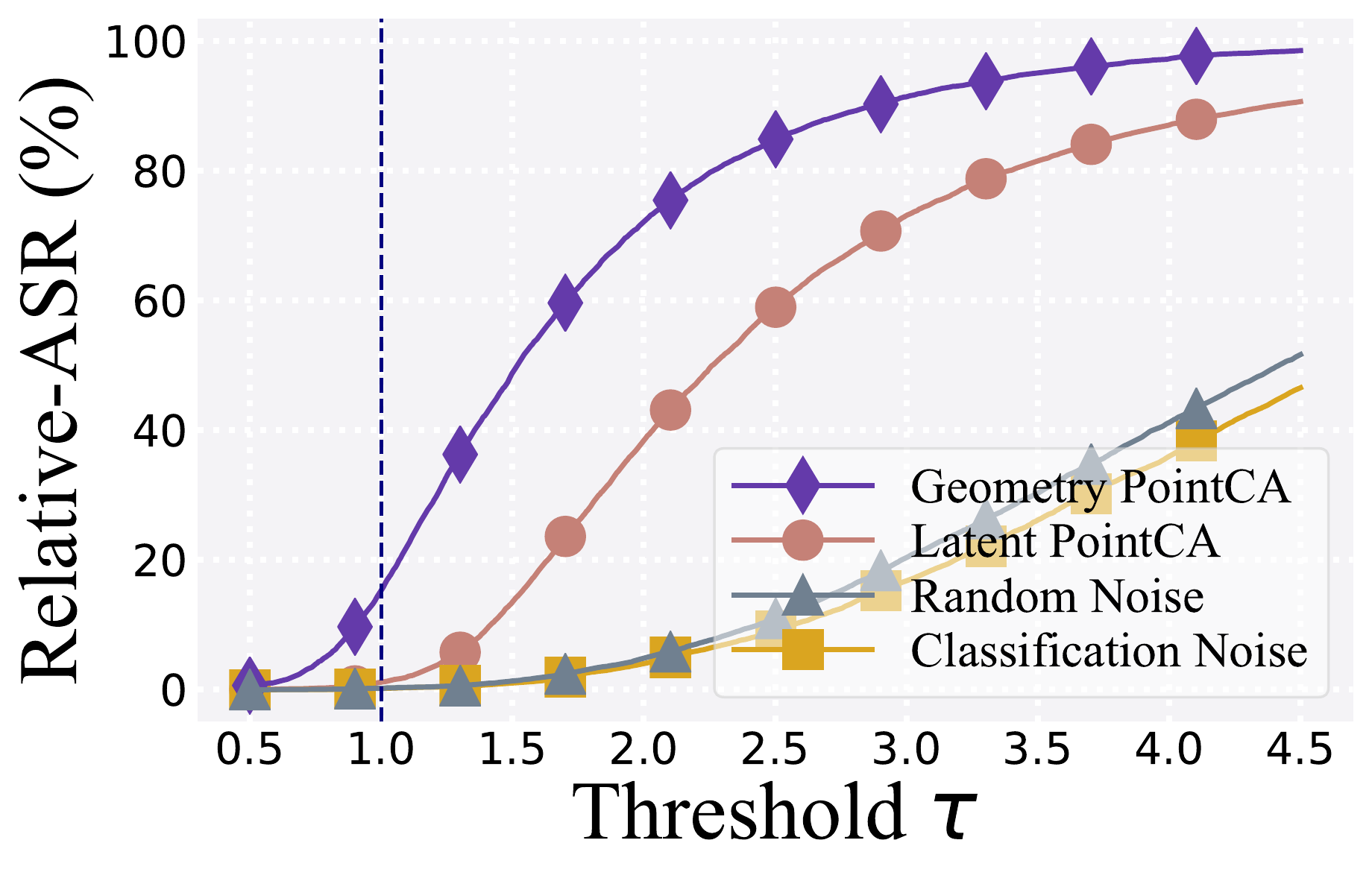}   
}
\vspace{-1em}
\caption{Relative attack success rates under different thresholds $\tau$}
\label{fig:RASR}
\vspace{-1em}
\end{figure}

The results of Relative-ASR on PCN and RFA are depicted in Fig.~\ref{fig:RASR}.
We can see there is a great gap between our attacks and baselines, which obviously demonstrates the strong attack ability of PointCA. Meanwhile nearly 30$\%$ of adversarial examples on PCN and 15\% generated on RFA in geometry PointCA have a T-NRE below 1. The majority of examples generated by PointCA have a T-NRE below 2, further proving our attack's effectiveness.

\noindent\textbf{Semantic evaluation.}
We train three classifiers: PointNet, PointNet++, and DGCNN, with modelnet10 dataset. Subsequently, the complete point clouds reconstructed on our adversarial examples are fed into these models to obtain the classification results. In this way, we can comprehensively evaluate our attack methods in a point cloud processing pipeline rather than on an immutable independent model.


As shown in Table~\ref{tab:semantic}, the \textit{Accuracy} (ACC) of all the three classifiers decreases drastically, \eg, PointNet++ suffers from the decrease from 77.9$\%$ to 16.7$\%$. It suggests that the completion outputs of partial adversarial point clouds have changed into other shapes with semantics different from the original ones. The \textit{Attack Success Rate} (ASR) of latent attack is higher, indicating the feature similarity might be more effectively transmitted by the decoder and has a greater influence on downstream tasks than geometry similarity. Besides, the average ASR of all reconstructed point clouds is about 32$\%$. Although only a part of the outputs mislead the classification models to target labels, most of them cannot be identified as original semantics.


\begin{figure}[t] 
\centering
\subfigure[T-NRE on PCN] 
{
    \label{comparasion_pcn_tnre}
	\centering         
	\includegraphics[width=0.22\textwidth]{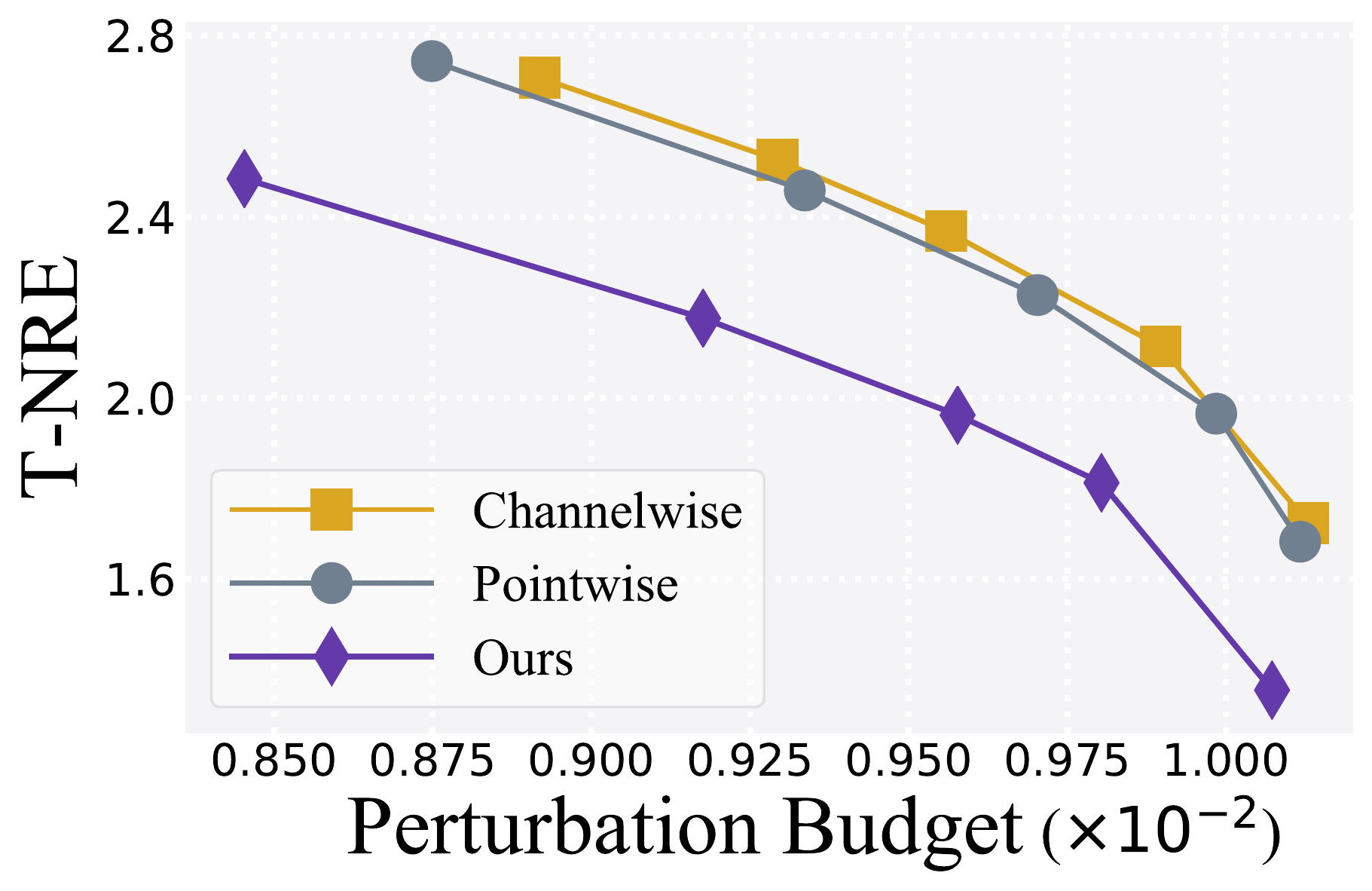}   
}
\subfigure[Outliers on PCN] 
{
    \label{comparasion_pcn_outliers}
	\centering     
	\includegraphics[width=0.22\textwidth]{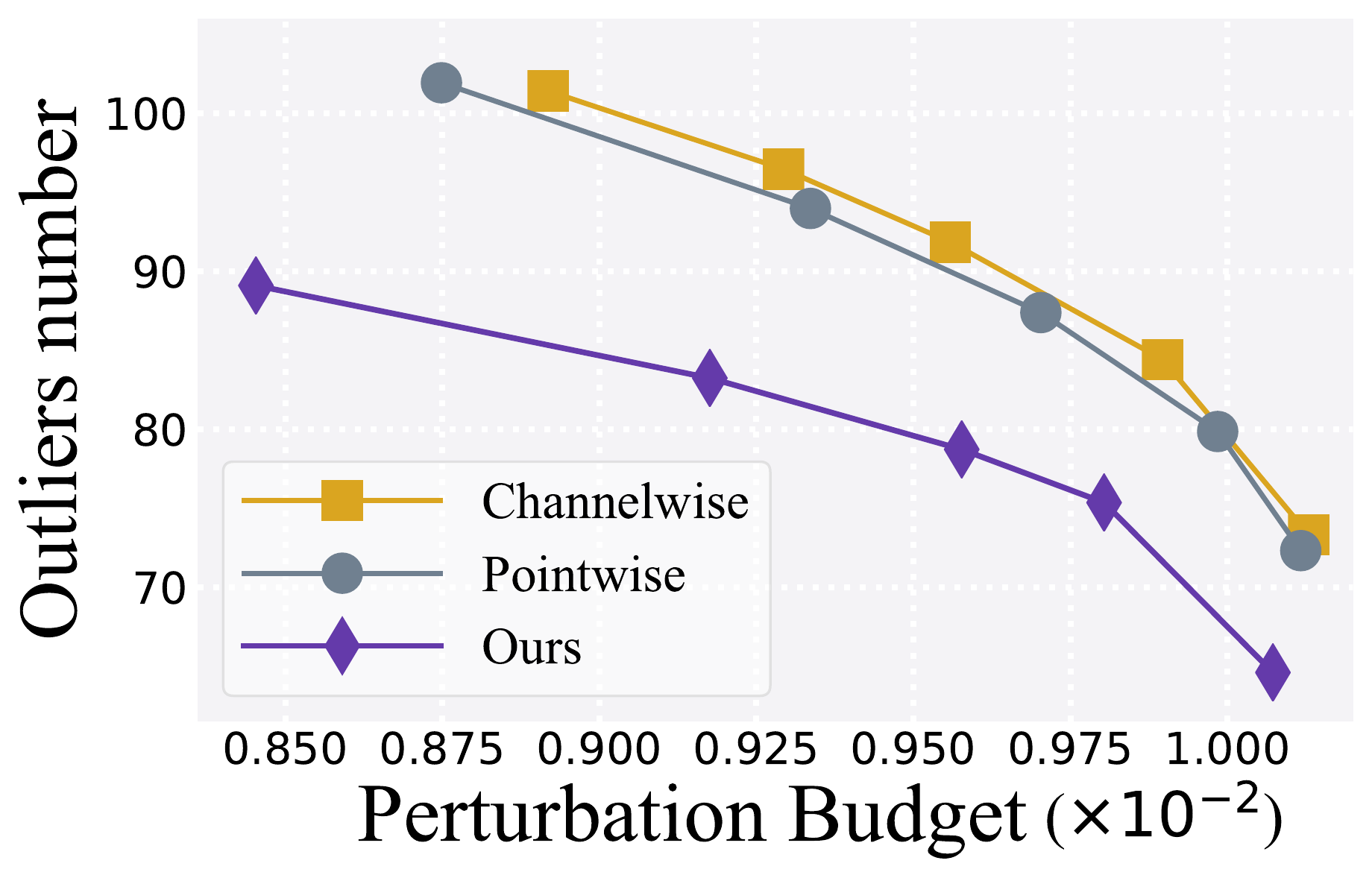}   
}

\subfigure[T-NRE on RFA] 
{
    \label{3}
	\centering          
	\includegraphics[width=0.22\textwidth]{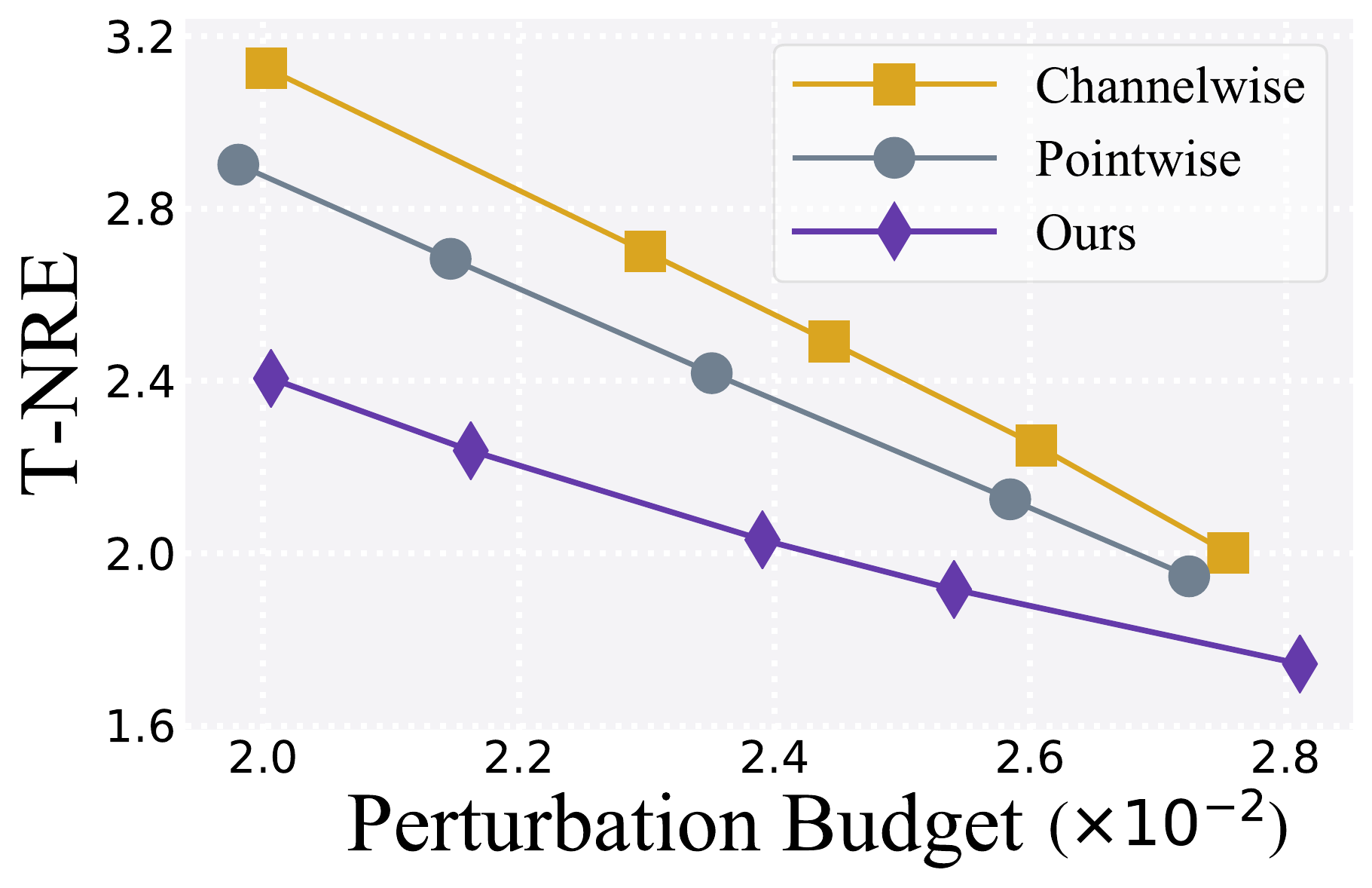}  
}
\subfigure[Outliers on RFA] 
{
    \label{4}
	\centering      
	\includegraphics[width=0.22\textwidth]{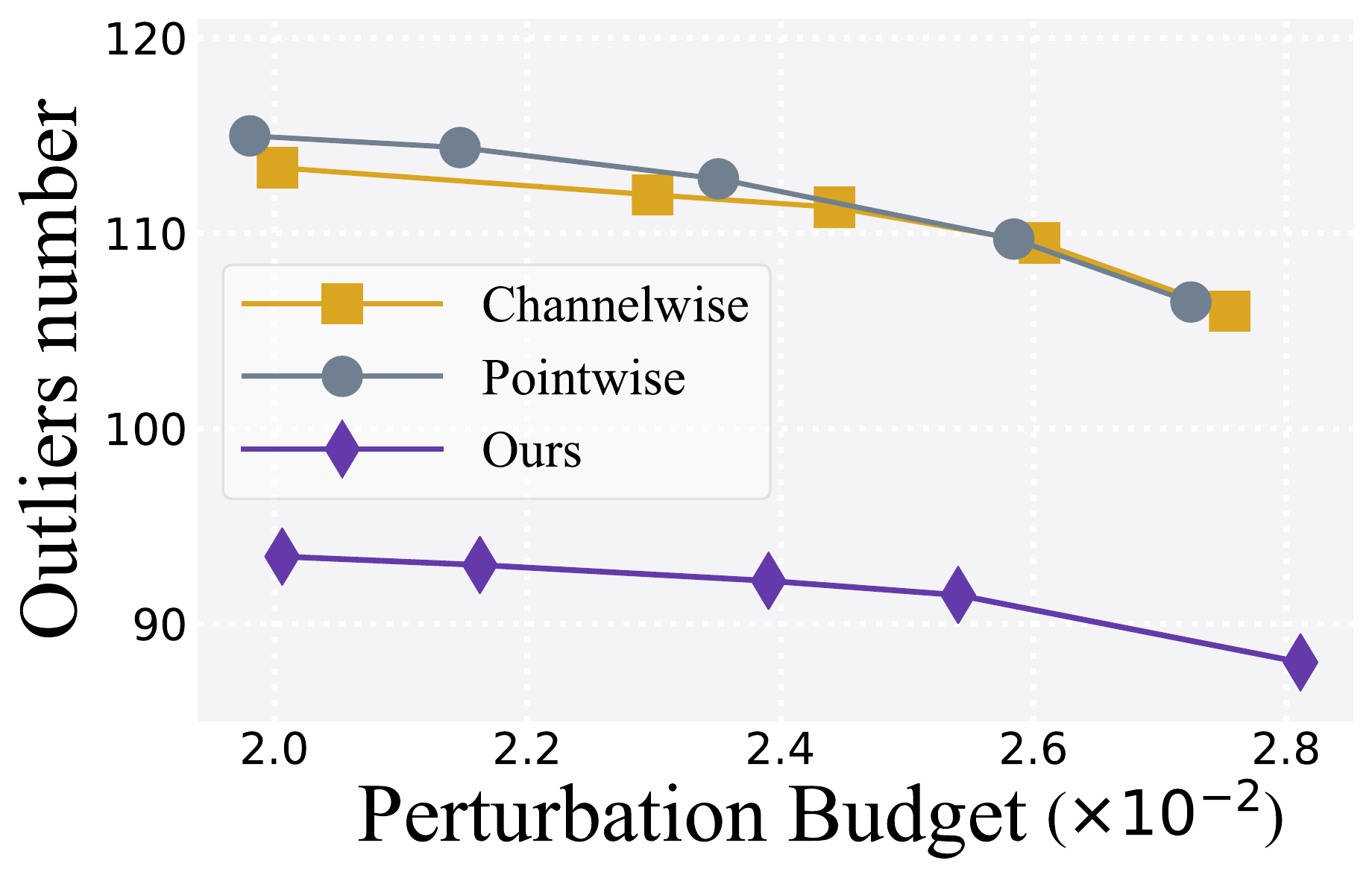}
}

\vspace{-1em}
\caption{Comparison between different perturbation constraint strategies on PCN and RFA
}
\label{fig:comparasion}
\vspace{-1em}
\end{figure}

\begin{figure}[]
    \centering
    \includegraphics[width=0.94\columnwidth]{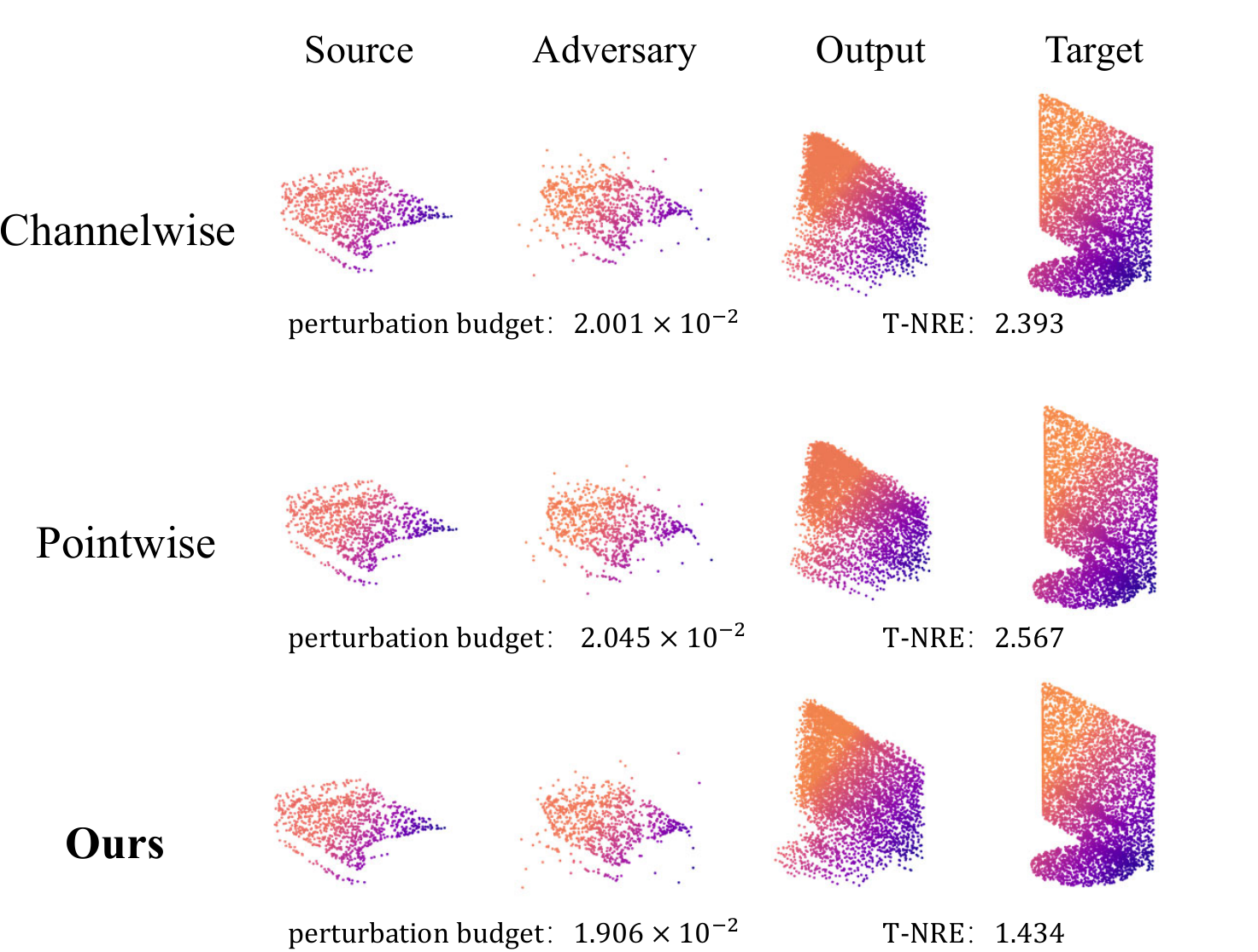}
    \caption{Visualization results of different perturbation constraint strategies. Our adaptive geometric constraint  can obtain a better adversarial completion performance with less perturbation budget. 
    }
    \label{fig:comparision_visual}
    \vspace{-1em}
\end{figure}

\subsection{Comparison Study}
\label{sec:Comparision Study}
\textbf{Implementation details.} 
In this section, we compare our adaptive geometric constraint with two popular perturbation constraint strategies in classification attacks: channelwise $l_{\infty}$-norm clip~\cite{hamdi2020advpc} and pointwise $l_{2}$-norm clip~\cite{ma2020efficient}, whose basic ideas of constraining the perturbation through clipping method are similar to ours.
To ensure a fair comparison, we use \textit{perturbation budget} and \textit{outliers number}  to analyze the adaptive constraint.

\noindent\textbf{Analysis.}
The results on PCN and RFA are shown in Fig.~\ref{fig:comparasion}. Notably, compared with channelwise $l_{\infty}$-norm and pointwise $l_{2}$-norm constraints, our method has an obvious advantage in generating a high quality of adversarial examples, \ie,  a better T-NRE and fewer outliers under the same perturbation budget. 
It also verifies our insight that not all points contribute equally to the final attack performance. 
The nonuniform and locally diverse data distributions of partial point clouds lead to an anisotropic search process in generating adversarial examples, thus an adaptive geometric constraint strategy is necessary in point cloud completion attack.
Although channelwise $l_{\infty}$-norm and pointwise $l_{2}$-norm constraints incur more perturbation budget, they are unable to take full advantage of the budget since some perturbations are useless for boosting an effective attack. Fig.~\ref{fig:comparision_visual} visualizes the examples generated under three perturbation constraints for comparison. 

\begin{table}[t]
\renewcommand\arraystretch{1.16}
\resizebox{\columnwidth}{!}{%
\begin{tabular}{ccccccccc}
\toprule[1.5pt]
\multicolumn{2}{c}{\textbf{Defense}} & \textbf{None} & \multicolumn{2}{c}{\textbf{SRS}} & \multicolumn{2}{c}{\textbf{OR}} & \multicolumn{2}{c}{\textbf{SOR}} \\ 
\cmidrule(r){4-5} \cmidrule(r){6-7} \cmidrule(r){8-9}
Model & $\boldsymbol{\eta}$ & S-RE & S-RE$(\uparrow)$  & S-NRE$(\uparrow)$ & S-RE$(\uparrow)$ & S-NRE$(\uparrow)$ & S-RE$(\uparrow)$ & S-NRE$(\uparrow)$ \\ \hline
\multirow{4}{*}{PCN} & clean & 0.012 & 0.0122 & 1.0042 & 0.0124 & 1.0211 & 0.0142 & 1.1704 \\
 & 1.5 & 0.034 & 0.0310 & 2.7645 & \textbf{0.0203} & \textbf{1.7595} & 0.0173 & 1.4666 \\
 & 2.5 & 0.038 & 0.0352 & 3.1638 & 0.0198 & 1.7115 & \textbf{0.0174} & \textbf{1.4726} \\
 & 5 & 0.042 & \textbf{0.0382} & \textbf{3.4471} & 0.0194 & 1.6747 & 0.0173 & 1.4692 \\ \hline
\multirow{4}{*}{RFA} & clean & 0.010 & 0.0108 & 1.0248 & 0.0109 & 1.0363 & 0.0126 & 1.1995 \\
 & 1.5 & 0.026 & 0.0248 & 2.5256 & 0.0217 & 2.1788 & 0.0207 & 2.0627 \\
 & 2.5 & 0.029 & 0.0283 & 2.9104 & 0.0242 & 2.4477 & 0.0232 & 2.3238 \\
 & 5 & 0.032 & \textbf{0.0310} & \textbf{3.1991} & \textbf{0.0259} & \textbf{2.6362} & \textbf{0.0248} & \textbf{2.5082} \\ \hline
\multirow{4}{*}{GRNet} & clean & 0.011 & 0.0113 & 1.0001 & 0.0113 & 1.0002 & 0.0129 & 1.1416 \\
 & 1.5 & 0.027 & 0.0267 & 2.5132 & 0.0267 & 2.5149 & 0.0254 & 2.3597 \\
 & 2.5 & 0.029 & 0.0291 & 2.7492 & 0.0267 & 2.5876 & 0.0271 & 2.5198 \\
 & 5 & 0.030 & \textbf{0.0301} & \textbf{2.8509} & \textbf{0.0301} & \textbf{2.8519} & \textbf{0.0277} & \textbf{2.5840} \\ \hline
\multirow{4}{*}{VRCNet} & clean & 0.010 & 0.0104 & 1.0092 & 0.0103 & 1.0010 & 0.0111 & 1.0832 \\
 & 1.5 & 0.028 & 0.0275 & 2.9079 & 0.0240 & 2.5138 & 0.0220 & 2.2938 \\
 & 2.5 & 0.031 & 0.0304 & 3.2242 & 0.0264 & 2.7734 & 0.0243 & 2.5479 \\
 & 5 & 0.032 & \textbf{0.0314} & \textbf{3.3409} & \textbf{0.0271} & \textbf{2.8567} & \textbf{0.0251} & \textbf{2.6290} \\ 
 \bottomrule[1.5pt]
\end{tabular}%
}
\caption{Evaluating PointCA under three defenses. Higher S-RE $(\uparrow)$ and higher S-NRE $(\uparrow)$ imply better attack strength.}
\label{tab:defense}
\vspace{-1em}
\end{table}

\subsection{Evaluation against Defenses}
\label{sec:Defense}
\textbf{Implementation details.}
We analyze the adversarial point clouds generated by Geometry PointCA under three mainstream defenses: \textit{Simple Random Sampling} (SRS), \textit{Outlier Removal} (OR), and \textit{Statistic Outlier Removal} (SOR)~\cite{huang2022shape,ShiCX0YH22}.  

\noindent\textbf{Analysis.}
The results in Table~\ref{tab:defense} show that although adversarial attacks can be somewhat mitigated, these defenses cannot fully recover the original shapes. SOR defense achieves a relatively good result, while in most cases, the reconstructed results of SOR denoised adversarial point clouds still have at least 50\% errors compared with normally restored samples. More results under diverse defense settings are depicted in our supplementary.

\section{Conclusion}\label{conclusion}
In this paper, we propose the first adversarial attack towards point cloud completion model, namely PointCA. 
The representation similarity in the  geometry space and the latent space is exploited to generate adversarial point clouds. 
We further design an adaptive geometric constraint depending on local density information for each point to improve the imperceptibility of PointCA. The comprehensive experiments verify the effectiveness and efficiency of our attack.

\section{Acknowledgments}\label{acknowledgments}
Shengshan’s work is supported in part by the National Natural Science Foundation of China (Grant No. U20A20177). Minghui's work is supported in part by the National Natural Science Foundation of China (Grant No. 62202186).







\appendix

\section{Supplementary Contents}
\begin{figure}[t]
    \setlength{\belowcaptionskip}{-0.5cm}   
    \centering
    \includegraphics[width=0.45\textwidth]{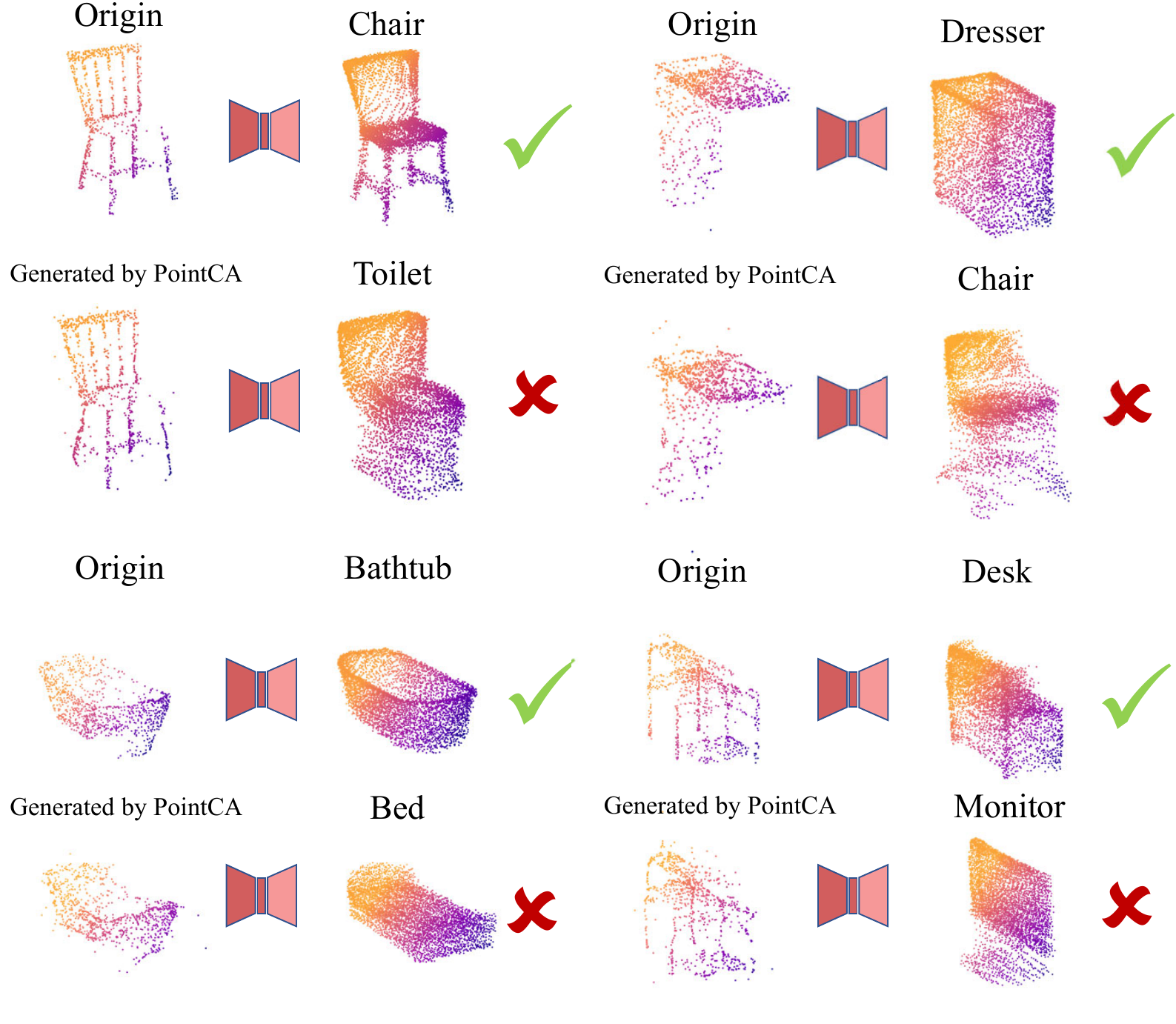}
    \vspace{-1em}
    \caption{Additional visualization of adversarial point clouds generated by PointCA. After adding imperceptible noises, the point clouds will be misleadingly completed as different objects.}
    \label{fig:instruction}
\end{figure}

\begin{itemize}
\item Fig.~\ref{fig:instruction}: Additional visualization of adversarial point clouds generated by PointCA.

\item Sec.A: Detailed experiment settings for evaluation metrics, training of completion models, and the implementation of PointCA. 

\item Sec.B: More supplemental evaluating results for Relative-ASR, comparison, defenses, transferability, and ablations.

\item Sec.C: More visualization results about local neighborhood density distribution information and adversarial point clouds generated on different completion models.

\end{itemize}

\section{A \quad Experiment Settings}
\label{sec:experiment}
\subsection{A.1 Evaluation Metrics}
The similarity function measures the difference between the output point cloud and the target point cloud. Since both point clouds are unordered, the loss needs to be invariant to permutations of the points. For a comprehensive study, we choose two candidates of permutation invariant functions which are introduced by Fan \etal~\shortcite{fan2017point} – \textit{Chamfer Distance} (CD) and \textit{Earth Mover’s Distance} (EMD). CD pays more attention to the difference between global structures. On the other hand, EMD is more discriminative to the local details and density distribution~\cite{wu2021density}.

\noindent\textbf{Chamfer distance.}
Based on two point cloud sets: $S_1$ and $S_2$, the algorithm of CD finds the nearest neighbor point set for each point and sums the squared distances up, which is depicted as:
\vspace{-0.5em}
\begin{equation}\label{eq:Chamfer Distance}
\begin{aligned}
D_{chamfer}\left(S_{1}, S_{2}\right)
=&\frac{1}{\left|S_{1}\right|}\sum_{x_{1}\in S_{1}}
\min_{x_{2}\in S_{2}}\left\|x_{1} - x_{2}\right\|_{2}\\
+&\frac{1}{\left|S_{2}\right|}\sum_{x_{2}\in S_{2}}
\min_{x_{1}\in S_{1}}\left\|x_{2} - x_{1}\right\|_{2}.
\end{aligned}
\end{equation}

Note that $S_1$ and $S_2$ do not need to have the same size to calculate CD. There are two variants for CD which are denoted as CD-P and CD-T~\cite{wang2020cascaded}. Specifically, CD-P takes the square root operation and is divided by 2. Considering that point cloud coordinates are usually normalized to $[-1,1]$ or a unit sphere for computational convenience, the calculated result of CD-P is usually larger than the value of CD-T under the same point cloud pair. 

In order to maintain consistent with previous work such as PCN~\cite{yuan2018pcn}, all the \textit{Chamfer Distance} metrics are measured through the CD-P method in our work. Although the specific values of the \textit{perturbation budget}, \textit{target reconstruction error} (T-RE) and \textit{target normalized reconstruction error} (T-NRE) can be further reduced if computed by CD-T method, which seems to benefit us. However, we think the difference between two \textit{Chamfer Distance} variants does not influence the correctness of the current results. 

\noindent\textbf{Earth mover's distance.}
The EMD between point cloud sets $S_1$ and $S_2$ is formulated as:
\begin{equation}\label{eq:Earth mover's Distance}
\begin{aligned}
D_{earth\;mover's}\left(S_{1}, S_{2}\right)
=&\min_{\phi \colon S_{1} \rightarrow  S_{2}}\frac{1}{\left|S_{1}\right|}\sum_{x \in S_{1}}\left\|x - \phi \left( x \right)\right\|_{2},
\end{aligned}
\end{equation}
where $\phi$ represents a bijection $\phi: S_1 \rightarrow S_2$, which minimizes the average distance between corresponding points.

\subsection{A.2 Dataset Generation}
We use the back-projected depth camera~\cite{yuan2018pcn} to generate 10 partial point clouds of 1024 points from 10 random viewpoints on each complete object, and set the number of ground truth point clouds to be 16384.
thus the completion dataset totally contains 33910 point cloud pairs for the training and 9080 point cloud pairs for the testing.
Despite the fact that there are a few real-world reconstruction datasets~\cite{dai2017scannet}, their ground truths contain missing regions and are not detailed enough to be used in completion tasks since the scanner's view is limited~\cite{yuan2018pcn,yu2021pointr}.

\subsection{A.3 Training of Victim Completion Models}
Because of the distinction between our completion Modelnet10 and other point cloud completion datasets, some custom modifications are necessary to the parameters of the completion model structures so that the completion models can be fully trained.

\noindent\textbf{PCN.} 
The chamfer version of PCN (PCN-CD) is selected as our victim model. We follow the same settings in~\cite{wang2021unsupervised} to train the model for 50 epochs with a batch size of 32. The fine loss of victim PCN on the test set is 0.01256 calculated in CD-P method.

\noindent\textbf{RFA.} 
The size of the input partial point cloud in our completion Modelnet10 is 1024, so the number of coarse point clouds is set to 1024. And the grid size and upratio are both set as 4, thus leading to an output fine point cloud whose size is 1024 $\times\, 4^{2}$ = 16384, which is consistent with the size of ground truth. We train the RFA model with the default parameters in the original work. The fine loss of victim RFA on the test set is 0.01091 calculated in the CD-P method and $5 \times10^{-4}$ when computed in the CD-T method.

\noindent\textbf{GRNet.} 
We change the sizes of both random point sampling and latent feature vector to 1024 to fit the input size of our partial point cloud. The GRNet model has been trained for 350 epochs with the base learning rate of $3\times 10^{-4}$. The fine loss of victim GRNet is $5.99\times 10^{-4}$, which is competitive with the results in official paper. Note that the official releasing code of GRNet implements the CD-T method, so its value may seem smaller. Whereas we compute the \textit{Chamfer Distance} in attacks through the CD-P method for consistency across multiple victim models.

\noindent\textbf{VRCNet.} 
The VRCNet model has been trained according to the author's default settings and the number of input point cloud is set as 1024. Evaluated on the test set, the CD-P loss of victim VRCNet is 0.01055, the CD-T loss is $5.54\times 10^{-4}$ and the F-Score is 0.6749. Given that the partial point clouds in our dataset only contain half as many points as the official MVP dataset, the model's fundamental performance is adequate to show the validity of our attack strategy.

\subsection{A.4 Baseline Implementation Details}
\noindent\textbf{Random noise}: We add random Gaussian noise to the input partial point clouds with a standard deviation of 0.02.

\noindent\textbf{Classification noise}: 
Firstly, we train a Pointnet++ classification model on the partial point cloud dataset of Modelnet10, then employ the method in~\cite{xiang2019generating} to generate the adversarial examples on classification model, whose attack success rate can reach nearly $100\%$. These adversarial examples produced on PointNet++ are subsequently fed to the completion models and get the evaluation results. 

All these baseline noises are generated under the same local geometric density adaptive constraints with our geometry and latent attack.

\subsection{A.5 Attack Implementation Details}
A gradient-based algorithm is employed to construct perturbative point data such that the similarity loss in different representation spaces can be minimized between the adversarial point cloud and the target point set. We also implement a variable step size strategy. The specific parameter settings are depicted in Table~\ref{tab:parameters}. $Geometry / Latent$ represents the parameters under the Geometry PointCA and Latent PointCA respectively. 
\vspace{-0.5em}
\begin{table}[h]
\centering
\renewcommand\arraystretch{1.2}
\resizebox{0.47\textwidth}{!}{%
\begin{tabular}{ccccc}
\toprule[1.3pt]
\textbf{Victim Model}      & \textbf{PCN}      & \textbf{RFA}      & \textbf{GRNet}   & \textbf{VRCNet}    \\ \hline
Number of Iterations         & 200      & 200      & 200    & 200  \\
hyperparameter $\lambda$    & 1000   & 20   & 0.05  & 20 \\
Base Step Size $\beta$           & 0.01/0.5     & 0.5/0.75      & 0.003     & 0.002/0.01 \\
Decay Rate     & 0.7/0.5     & 0.6     & 0.5    & 0.5/0.7  \\
Decay Step     & 20       & 20       & 50     & 30/20  \\
\bottomrule[1.3pt]
\end{tabular}
}
\caption{Parameters of attack setting on four target models}
\label{tab:parameters}
\vspace{-1em}
\end{table}

\subsection{A.6 Defense Implementation Details}
\noindent\textbf{Simple Random Sampling.} 
The simple random sampling is a subset of points chosen at random from a larger group, and each point is taken with the same probability. The proportion of total points that are arbitrarily lost is determined by \textbf{the Drop rate}~\cite{zhou2019dup}.

\noindent\textbf{Outlier Removal.}
Relocating outlying points that may be produced as a result of adversarial perturbations is another method defending against adversarial attacks.
The mean euclidean pointwise distance $d_{i}=\frac{1}{l}\sum_{j=1}^{l}d_{i,j}$ between each central point and its \textit{k-Nearest Neighbors} ($k$NN) points is used to identify outliers~\cite{liu2019extending}. Extremely high mean distance points who exceed the setting \textbf{Threshold} are thought to be outliers and eliminated. This method makes the assumption that any outlier point must be the product of adversarial perturbations because each point on a natural shape should be consistently sampled along the surface.

\noindent\textbf{Statistic Outlier Removal.} 
Statistic Outlier Removal method also leverages $k$NN to compute the average pairwise distance for each point, however the mean $\mu=\frac{1}{m}\sum_{i=1}^{m}d_{i}$ and the standard deviation $\sigma=\sqrt{\frac{1}{m}\,\sum_{i=1}^{m}\,({d}_{i} - {\mu})^{2}}\;$ of all these distances are combined to determine a filter threshold as $\mu + \alpha\cdot\sigma$ to trim the input, where \textbf{ $\boldsymbol{\alpha}$ serves as an interval hyperparameter}, so the denoised adversarial point cloud can be defined as $\mathbf{X}_{sor}^{P^{\prime}} = \{x_{i}|{d}_{i}< \mu + \alpha\cdot\sigma \}$. The number of outliers can be calculated by the difference in the size of point clouds before and after denoised~\cite{zhou2019dup}.

According to the efforts of the past~\cite{huang2022shape,ShiCX0YH22,zhou2019dup}, we set the Drop rate as 0.3 in SRS, the Threshold as 0.05 in OR ,the interval hyperparameter $\alpha$ as 1.1 and the defense neighborhood size $K$ of $k$NN as 2 in our main paper.

\section{B \quad More Qualitative Results}
\label{sec:results}
\subsection{B.1 Additional Relative-ASR Evaluation}
\textbf{Implementation details.} We leverage T-NRE as a reference to dynamically assess the \textit{relative attack success rate} (Relative-ASR). If one attack's T-NRE is below this threshold $\tau$, we can roughly judge the attack is successful. 

\begin{figure}[!] 
\setlength{\belowcaptionskip}{-0.5cm}
\centering
\subfigure[Relative-ASR on GRNet] 
{
    \label{rasr_grnet}
	\centering         
	\includegraphics[width=0.22\textwidth]{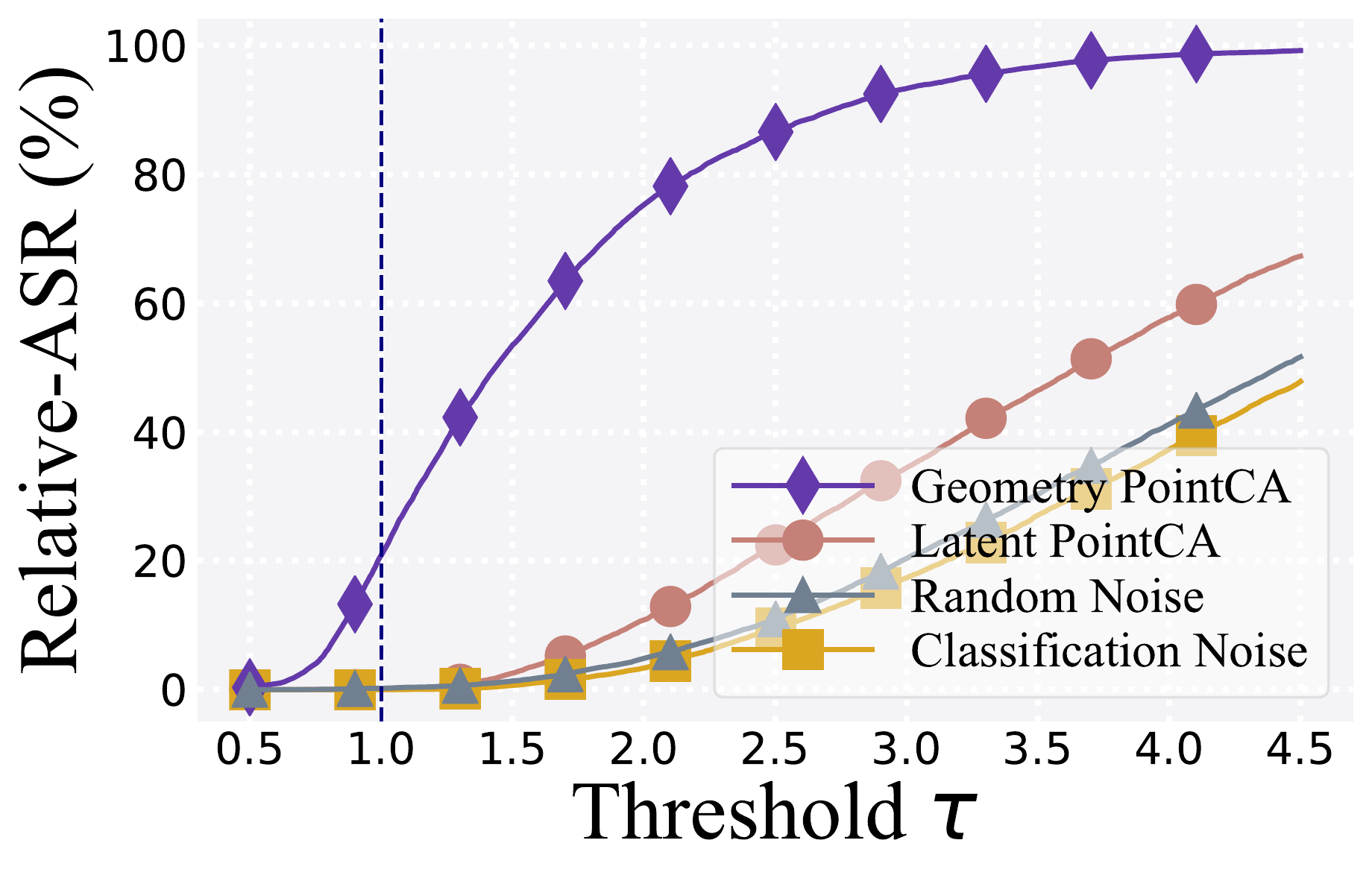}  
}
\subfigure[Relative-ASR on VRCNet] 
{
    \label{rasr_vrcnet}
	\centering     
	\includegraphics[width=0.22\textwidth]{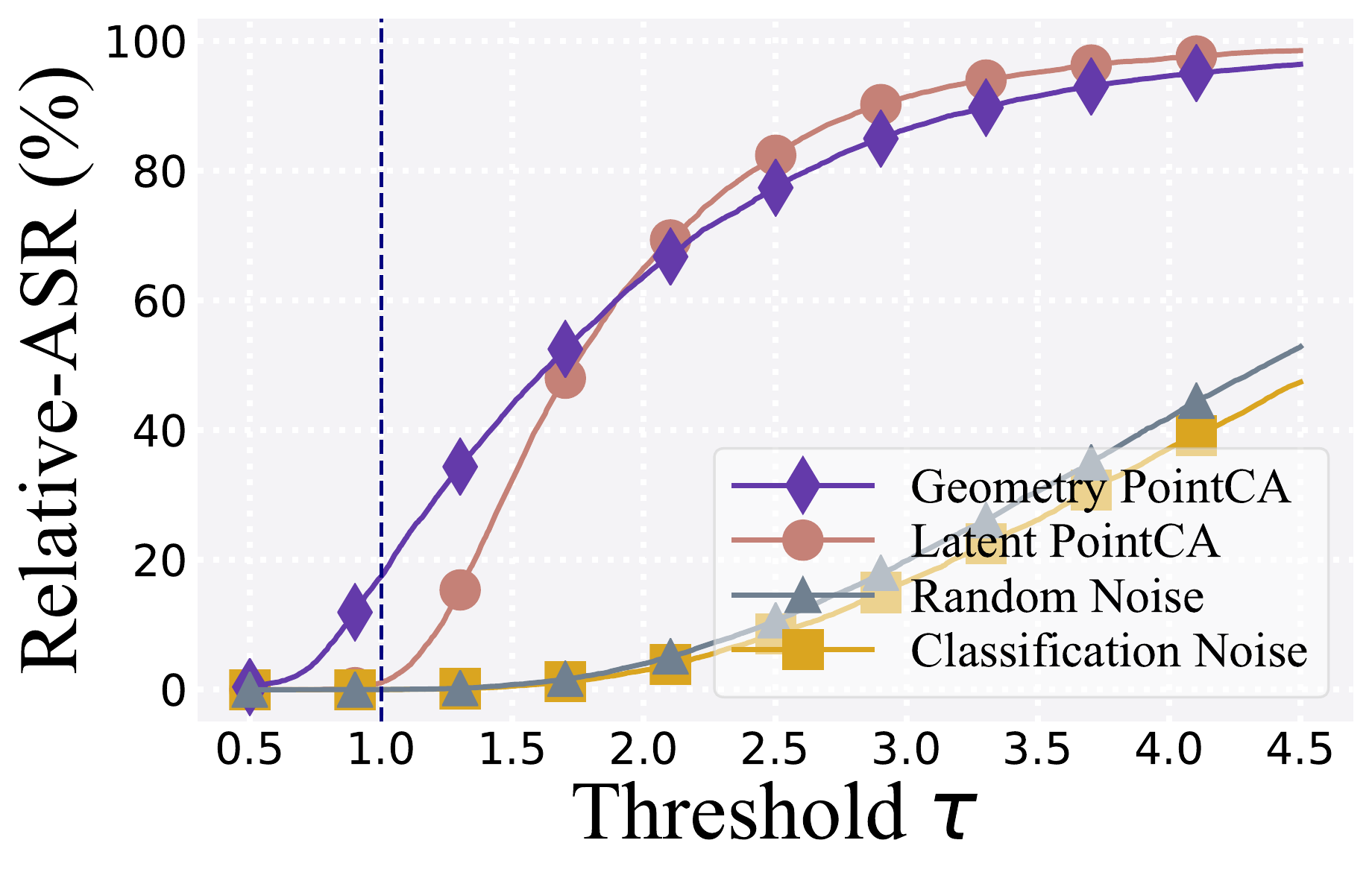}   
}
\vspace{-1em}
\caption{Relative attack success rates under different thresholds $\tau$}
\label{fig:Relative-ASR}
\end{figure}

\noindent\textbf{Analysis.} 
The Relative-ASR results evaluated on GRNet and VRCNet are depicted in Fig.~\ref{fig:Relative-ASR}. We can observe that the outcomes of our attack methods continue to substantially exceed the baselines on both models, demonstrating the power and applicability of our PointCA. 

\subsection{B.2 Additional Comparison Study}
\textbf{Implementation details.} Apart from the PCN and RFA in the main paper, we also compare our adaptive geometric constraint with two popular perturbation constraint strategies on GRNet and VRCNet for a comprehensive research. 

\begin{figure}[t] 
\setlength{\belowcaptionskip}{-0.5cm}  
\centering
\subfigure[T-NRE on GRNet] 
{
    \label{comparison_grnet_tnre}
	\centering         
	\includegraphics[width=0.22\textwidth]{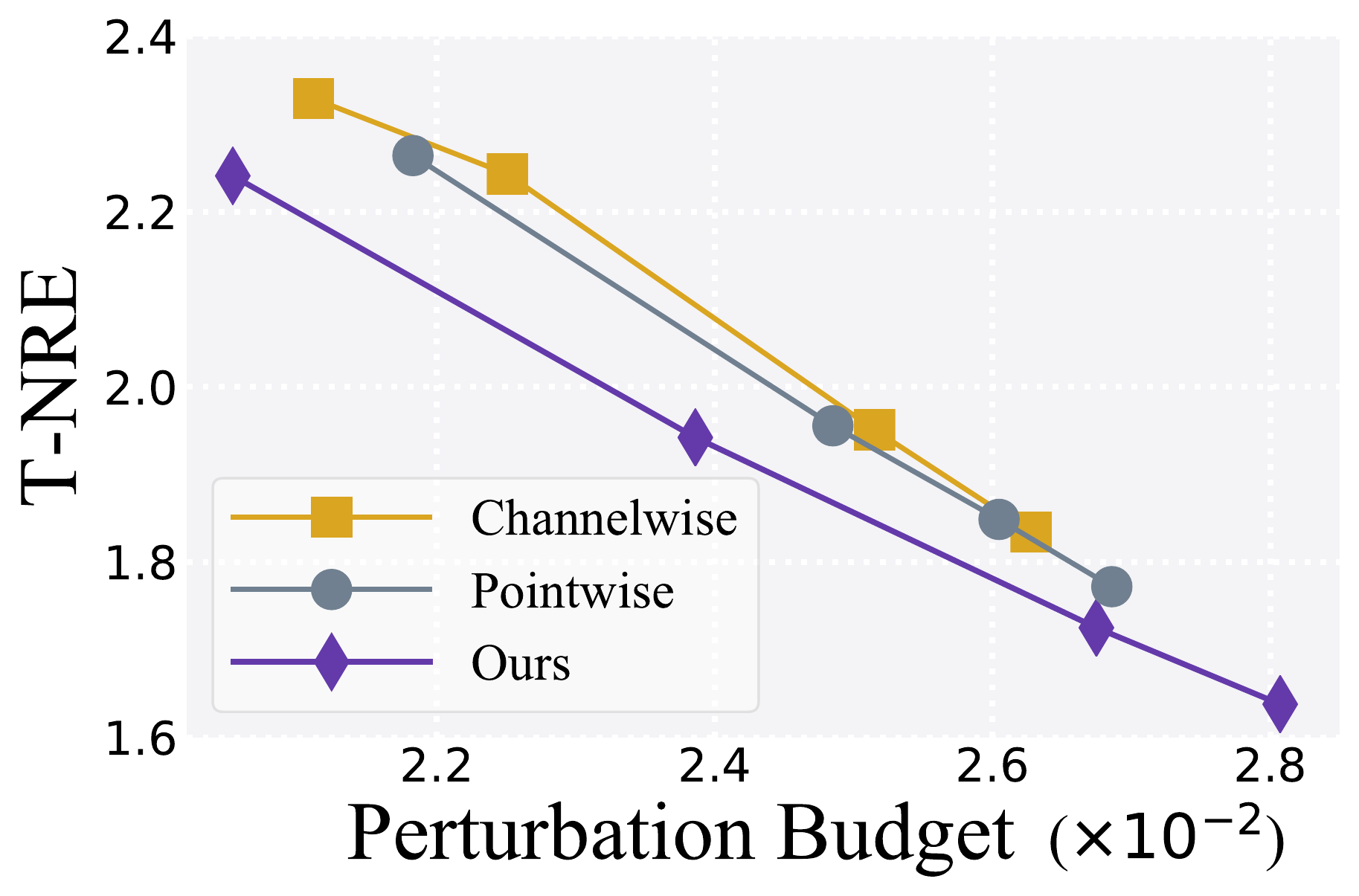}  
}
\subfigure[Outliers on GRNet] 
{
    \label{comparison_grnet_outliers}
	\centering     
	\includegraphics[width=0.22\textwidth]{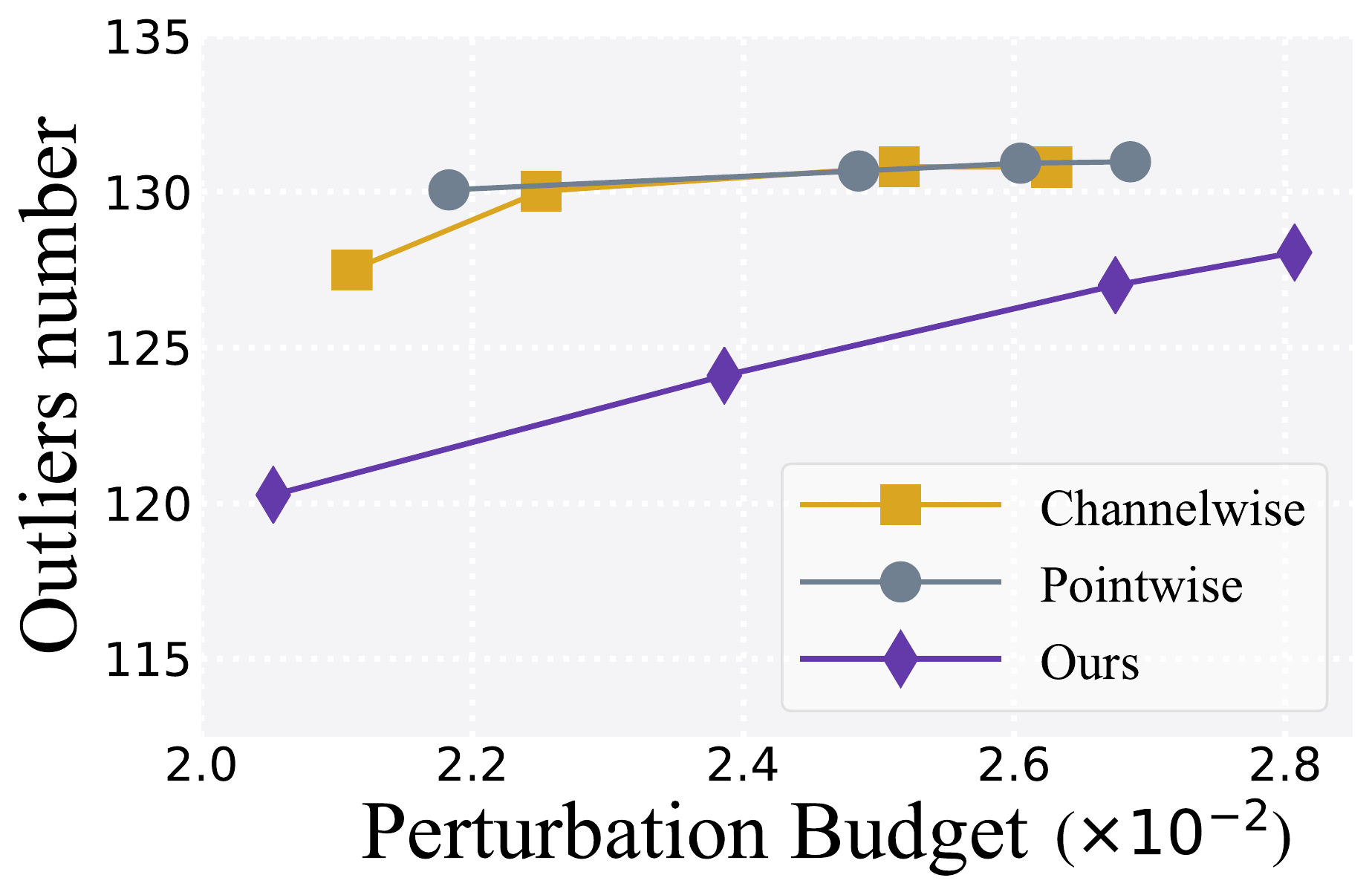}   
}
\subfigure[T-NRE on VRCNet] 
{
    \label{comparison_vrcnet_tnre}
	\centering     
	\includegraphics[width=0.22\textwidth]{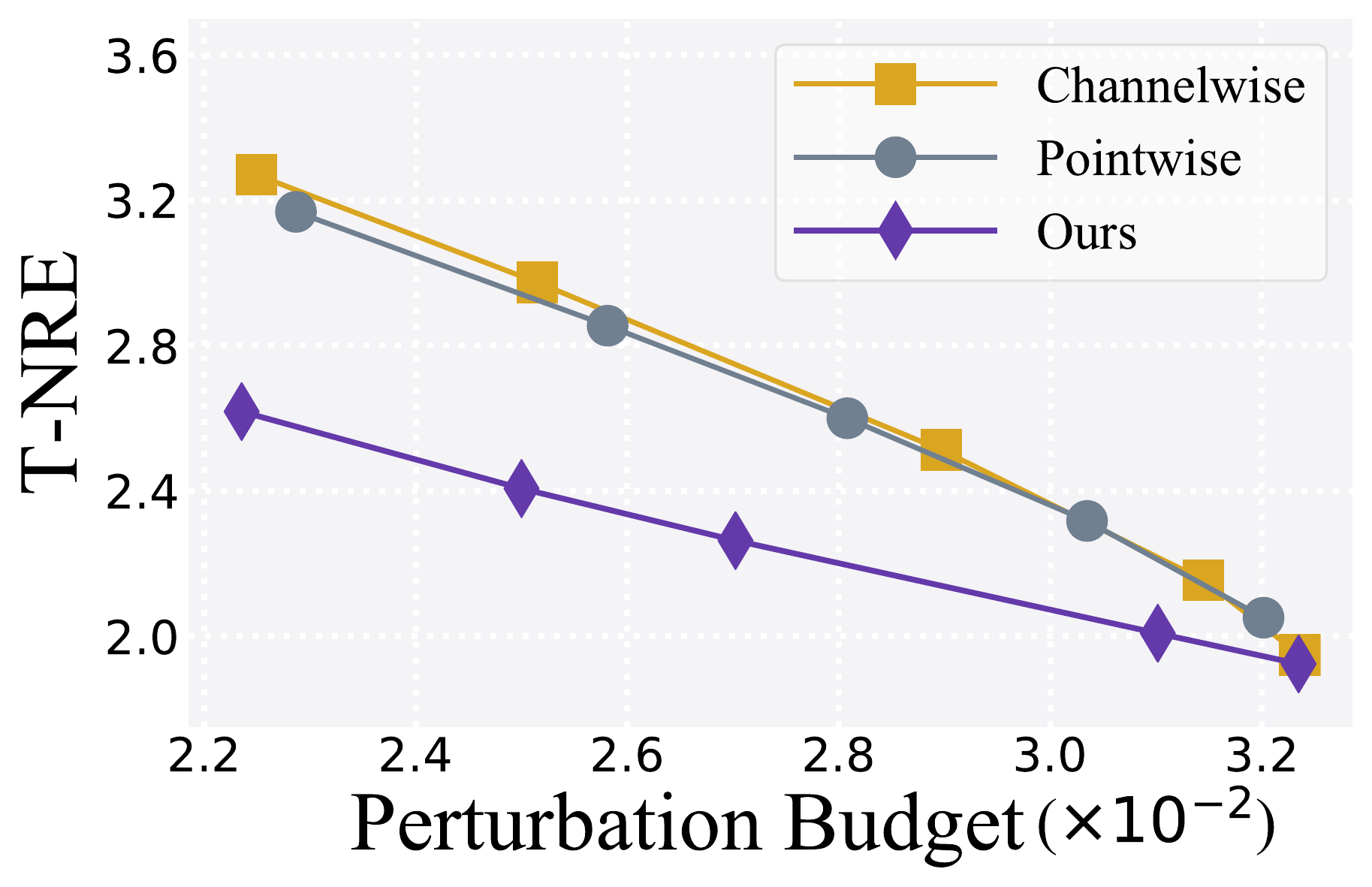}   
}
\subfigure[Outliers on VRCNet] 
{
    \label{comparison_vrcnet_outliers}
	\centering     
	\includegraphics[width=0.22\textwidth]{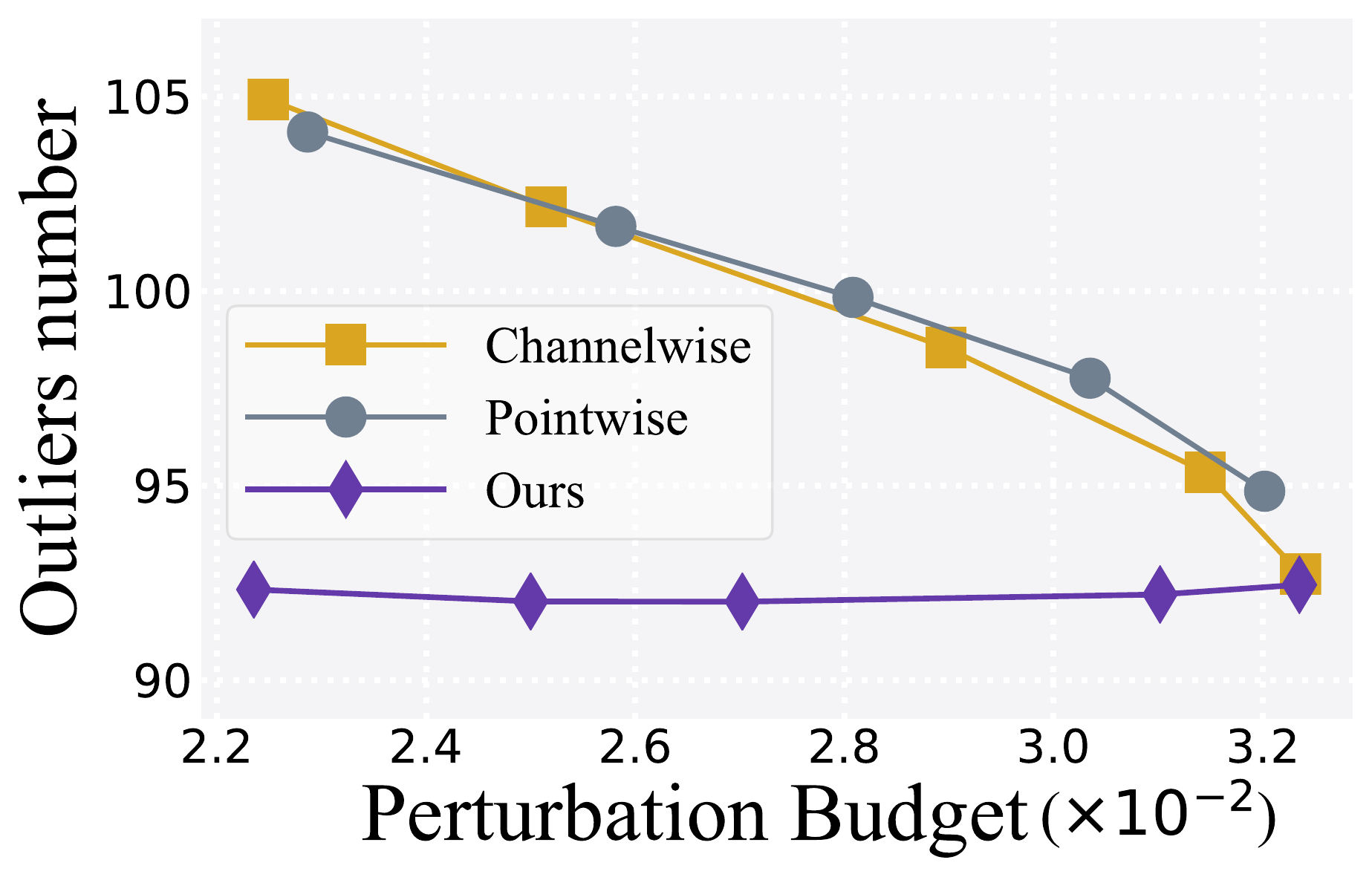}   
}
\vspace{-1em}
\caption{Comparison between different perturbation constraint strategies on GRNet and VRCNet}
\label{fig:comparison}
\end{figure}

\noindent\textbf{Analysis.} 
The results are shown in Fig.~\ref{fig:comparison}. The evaluating results on GRNet and VRCNet are consistent with the results on other two models.
Compared with channelwise $l_{\infty}$-norm and pointwise $l_{2}$-norm constraints, our method obtains a better T-NRE and fewer outliers under the same perturbation budget, which further confirms the correctness of our design.


\subsection{B.3 Additional Evaluations against Defenses}
\textbf{Implementation details.}
We investigate the Geometry PointCA-generated adversarial point clouds with three widely used defenses: \textit{Simple Random Sampling} (SRS), \textit{Outlier Removal} (OR) and \textit{Statistic Outlier Removal} (SOR). To evaluate the performance of our method against defenses in a more comprehensive and detailed manner, we conduct trials on different attack parameters under various defense settings.

\begin{figure*}[t] 
\centering
\subfigure[Drop rate=0.1] 
{
    \label{defense_srs_Rate0.1}
	\centering         
	\includegraphics[width=0.32\textwidth]{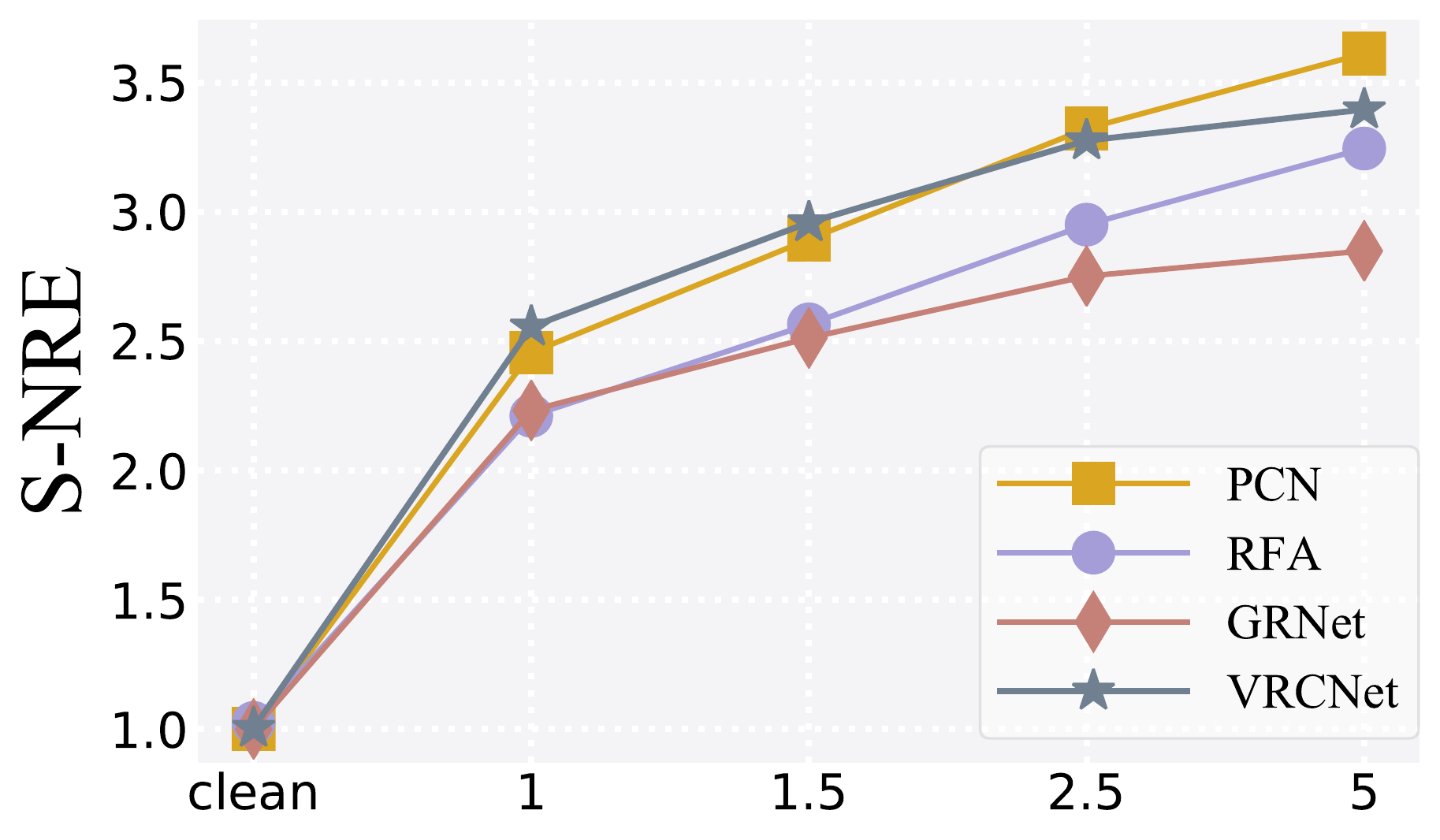}  
}
\subfigure[Drop rate=0.2] 
{
    \label{defense_srs_Rate0.2}
	\centering     
	\includegraphics[width=0.32\textwidth]{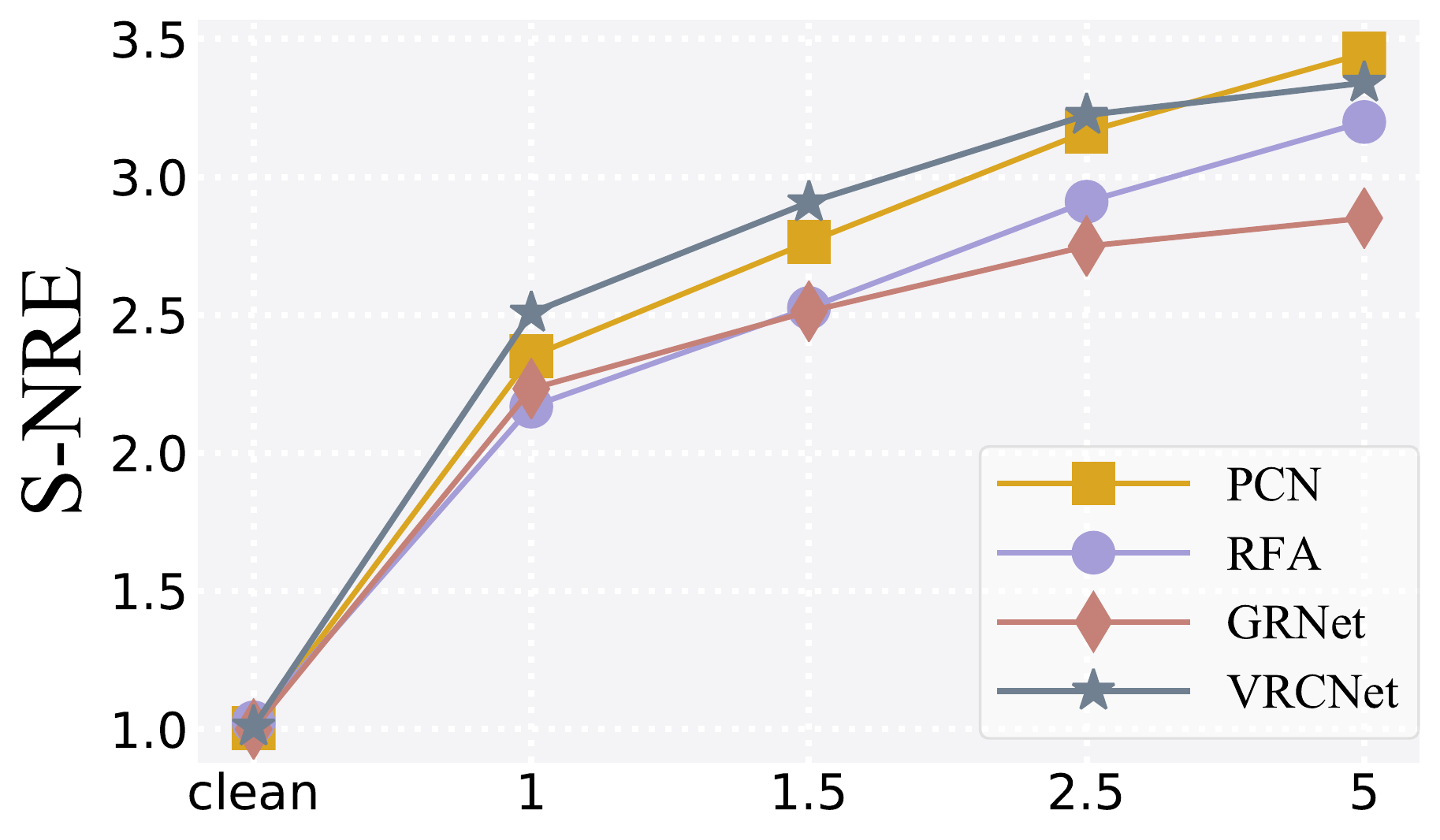}   
}
\subfigure[Drop rate=0.3] 
{
    \label{defense_srs_Rate0.3}
	\centering     
	\includegraphics[width=0.32\textwidth]{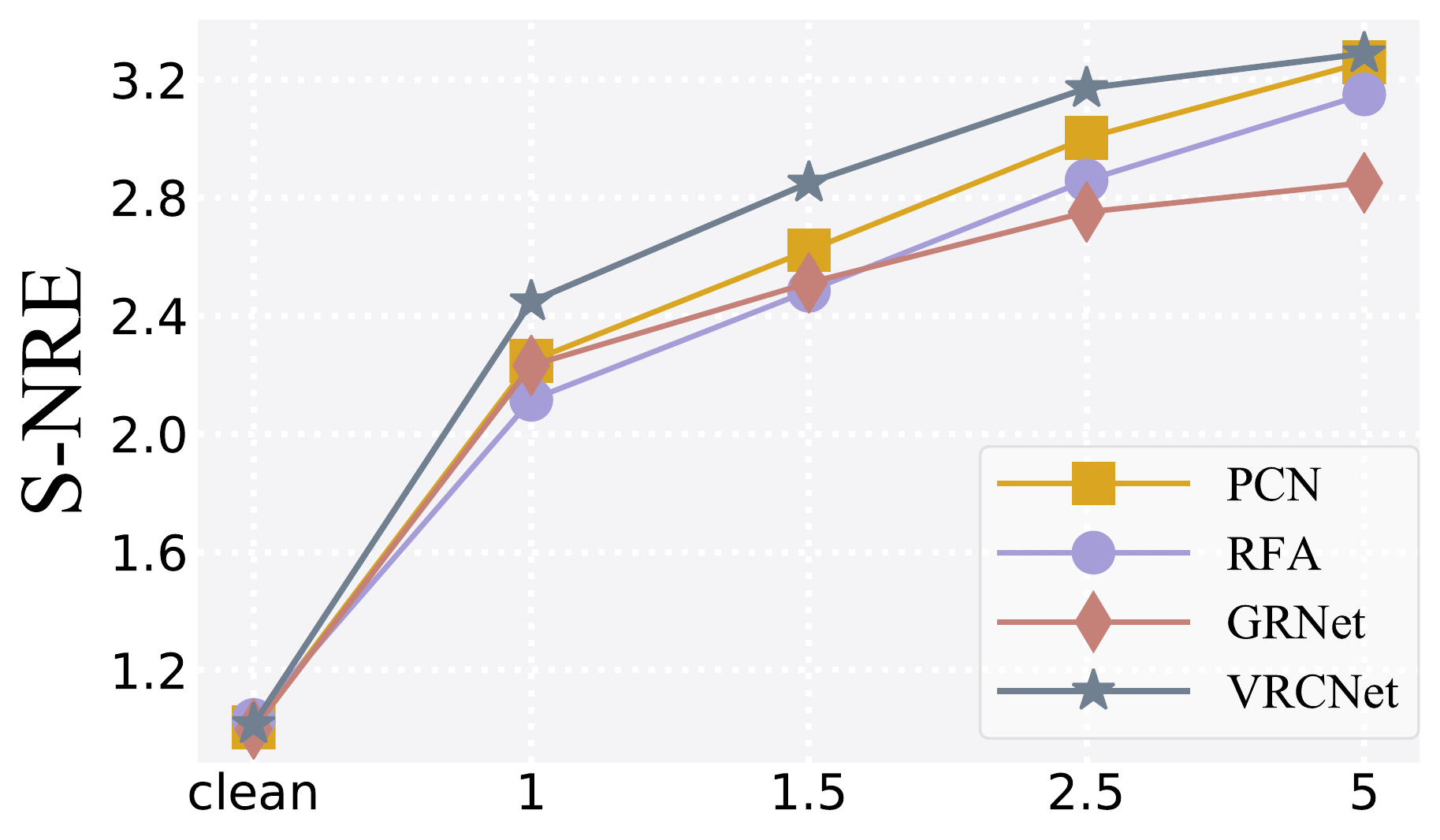}   
}
\vspace{-0.8em}
\caption{Comparison against the SRS defense. The drop rate varies from 0.1 to 0.3 from left to right, and scaling coefficients $\eta$ varies from 1 to 5 along x-axis in each sub-figure.}
\label{fig:defense_srs}
\end{figure*}

\begin{figure*}[!t] 
\centering
\subfigure[Threshold=0.03] 
{
    \label{defense_os_T0.03}
	\centering         
	\includegraphics[width=0.32\textwidth]{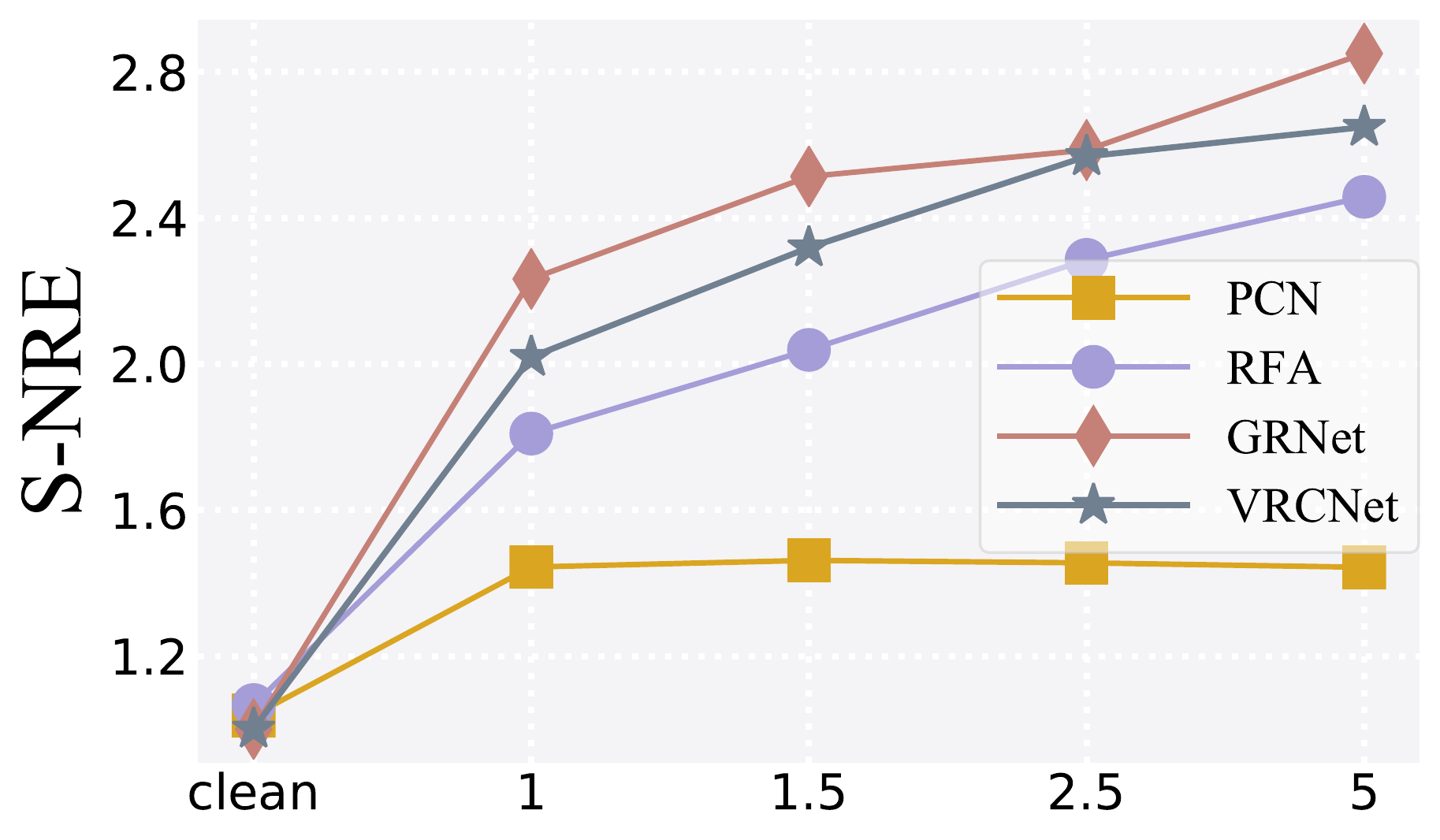}  
}
\subfigure[Threshold=0.05] 
{
    \label{defense_os_T0.05}
	\centering     
	\includegraphics[width=0.32\textwidth]{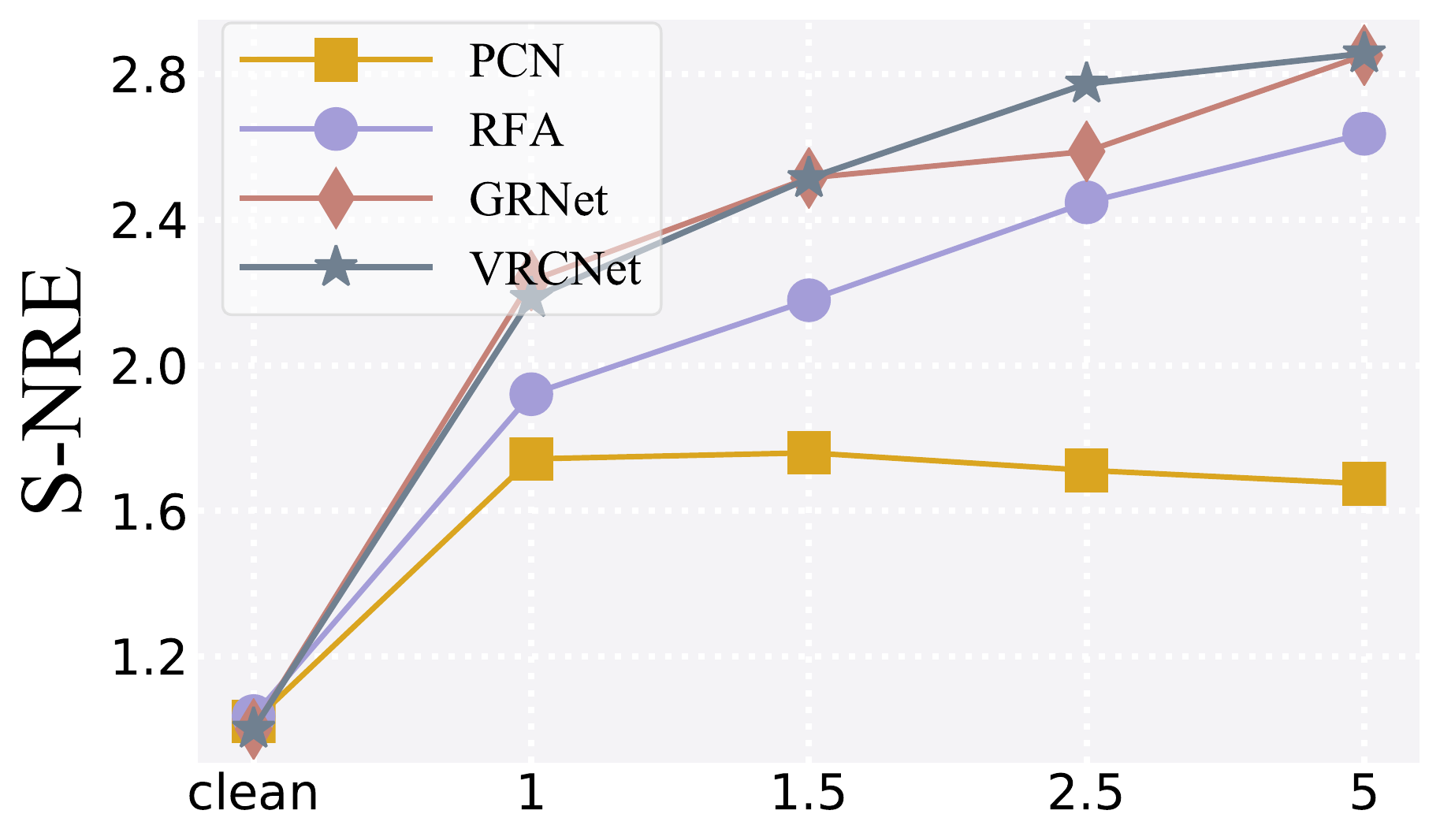}   
}
\subfigure[Threshold=0.07] 
{
    \label{defense_os_T0.07}
	\centering     
	\includegraphics[width=0.32\textwidth]{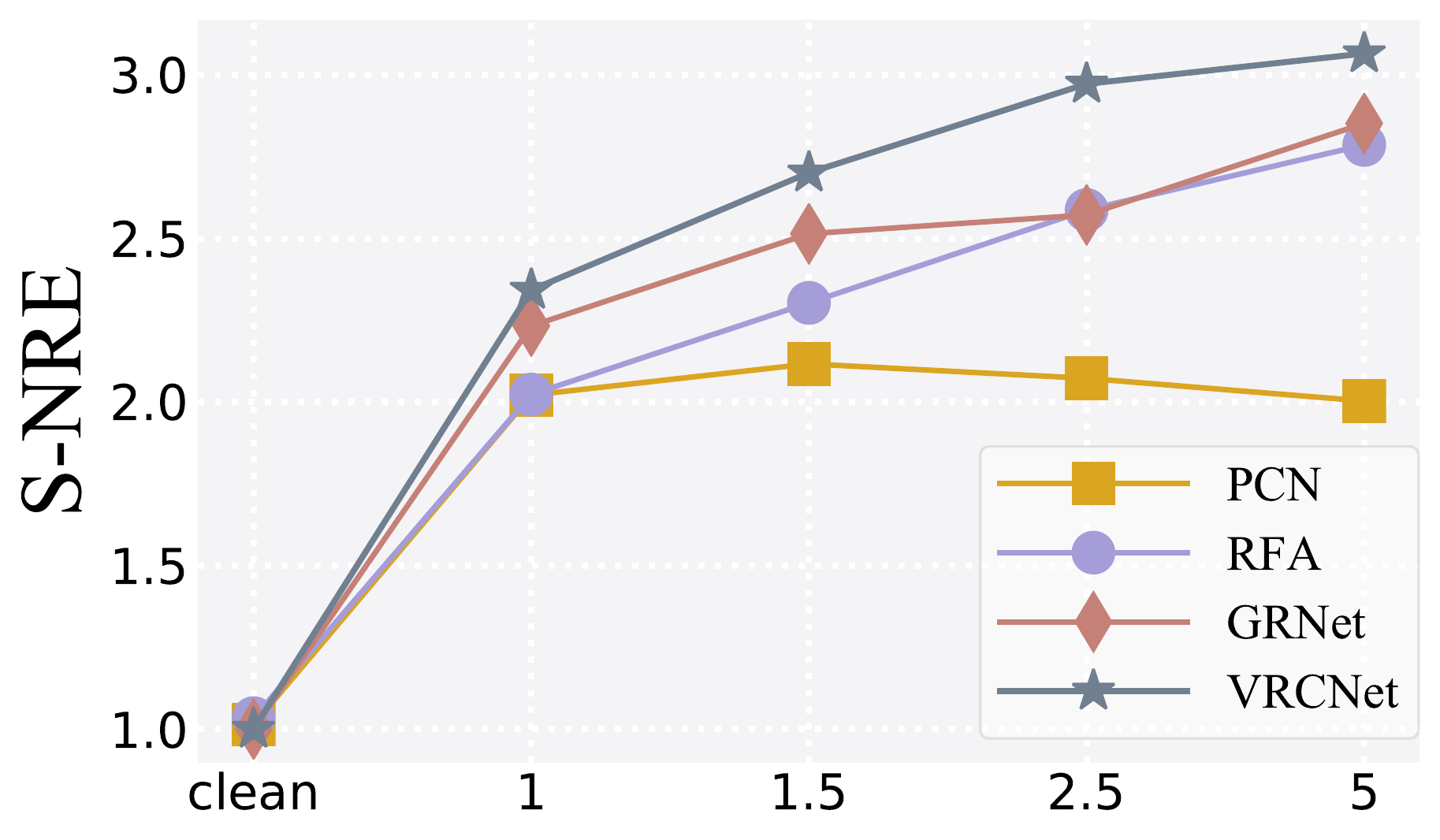}   
}
\vspace{-0.8em}
\caption{Comparison against the OR defense. The threshold varies from 0.03 to 0.07 from left to right, and scaling coefficients $\eta$ varies from 1 to 5 along x-axis in each sub-figure.}
\label{fig:defense_os}
\end{figure*}

\begin{figure*}[!t] 
\centering
\subfigure[$K$=2] 
{
    \label{defense_sor_k2}
	\centering         
	\includegraphics[width=0.32\textwidth]{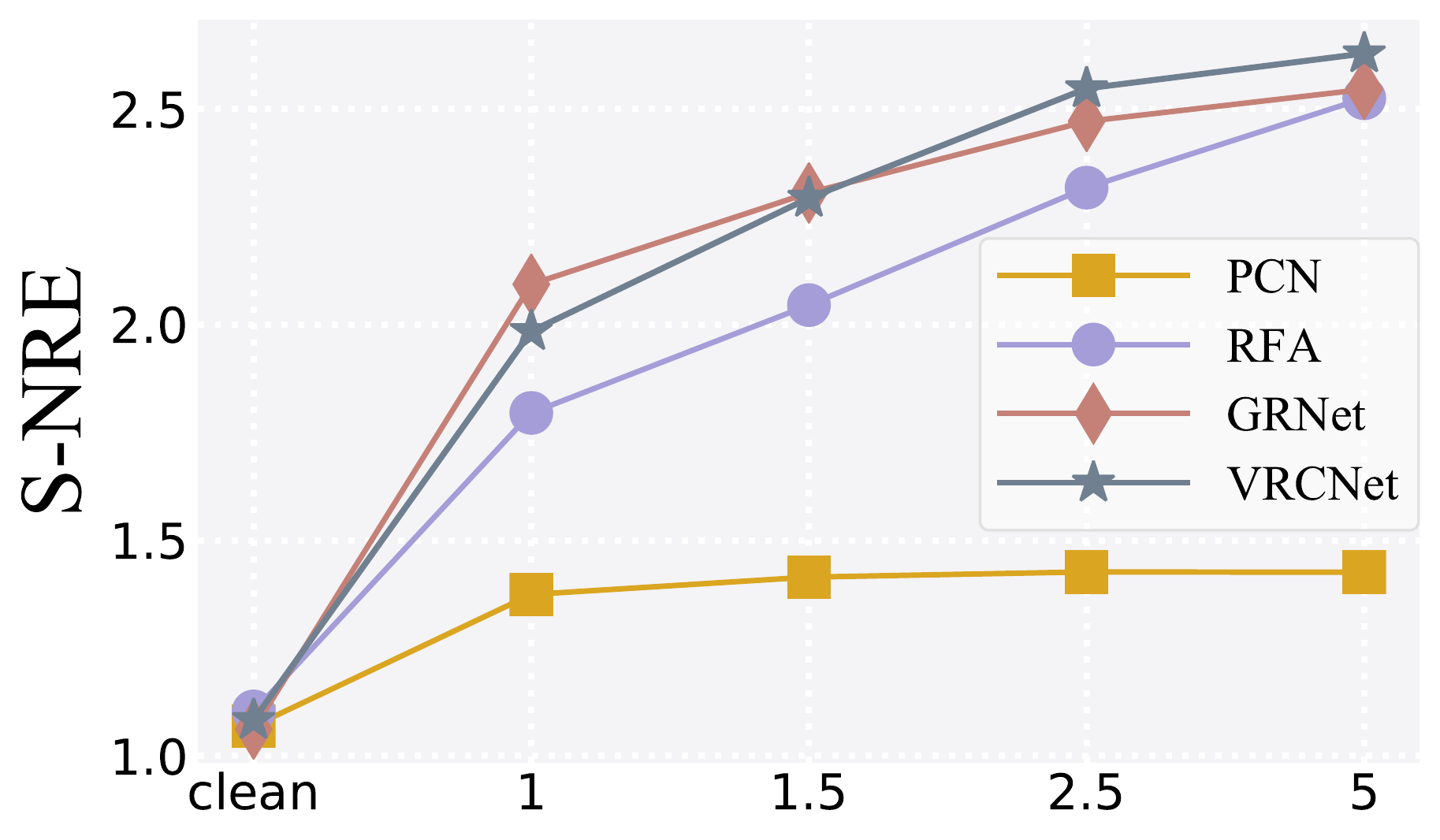}  
}
\subfigure[$K$=8] 
{
    \label{defense_sor_k8}
	\centering     
	\includegraphics[width=0.32\textwidth]{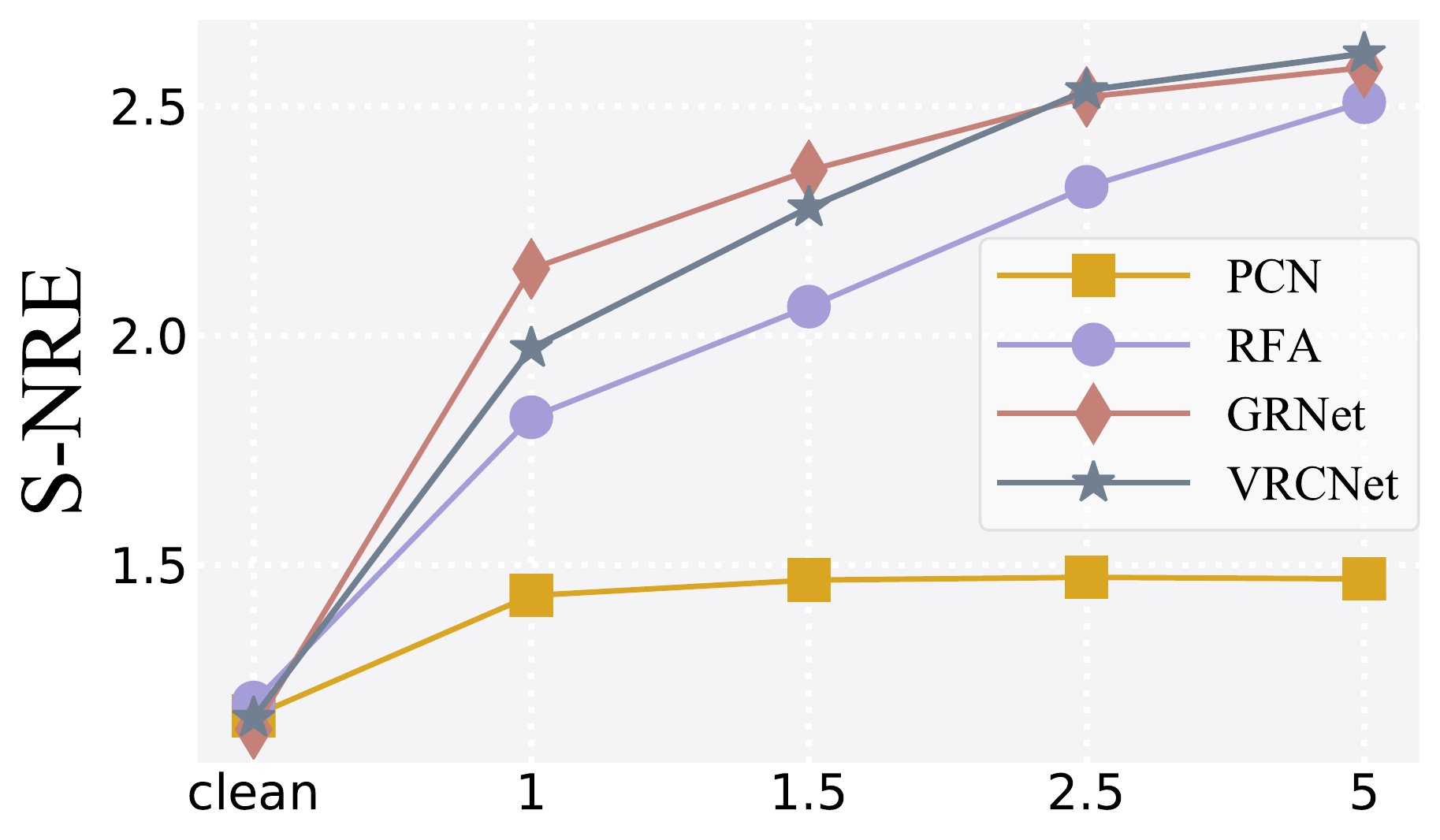}   
}
\subfigure[$K$=10] 
{
    \label{defense_sor_k10}
	\centering     
	\includegraphics[width=0.32\textwidth]{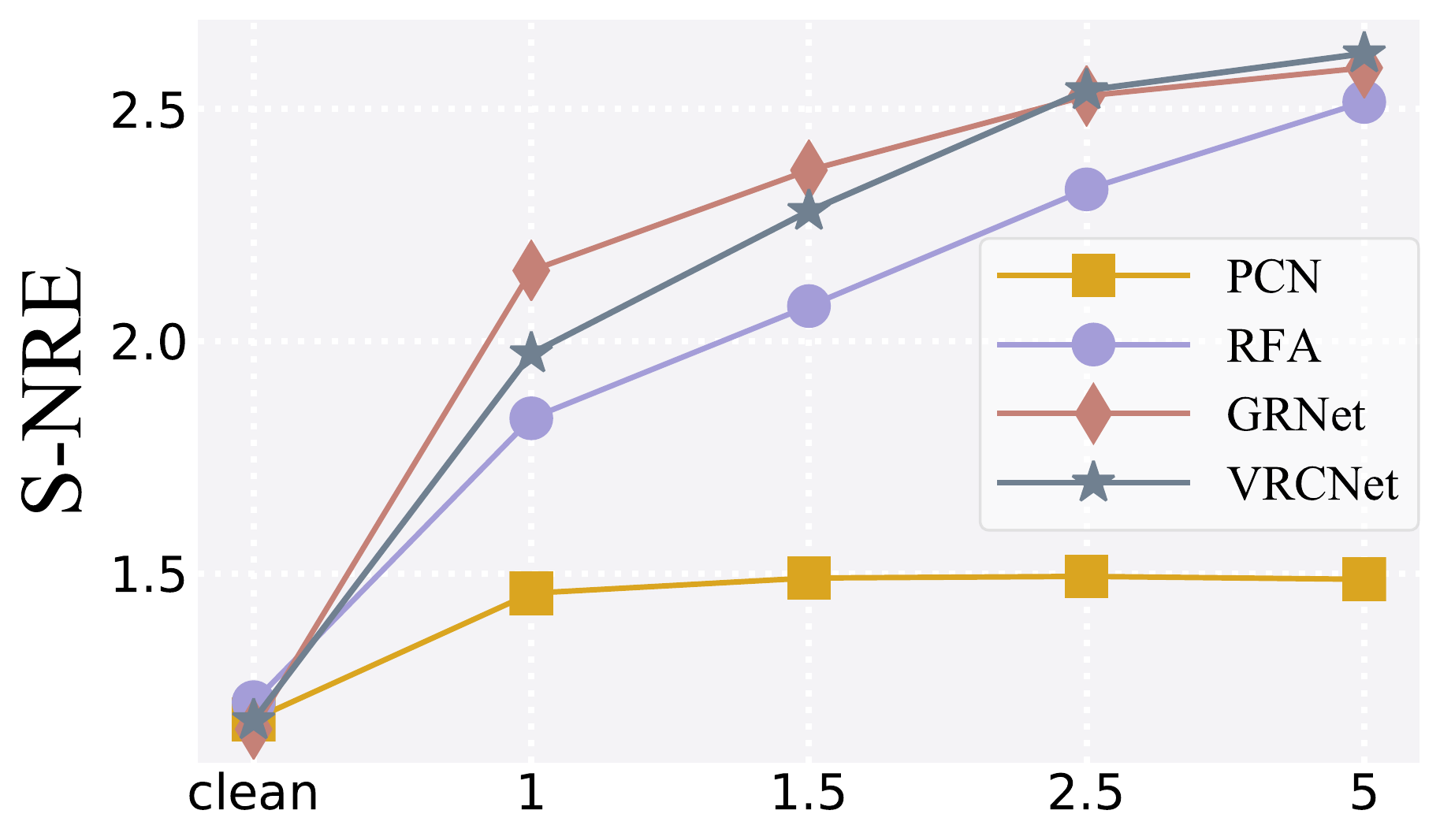}   
}
\vspace{-0.8em}
\caption{Comparison against the SOR defense. The defense neighborhood size $K$ varies from 2 to 10 from left to right, and scaling coefficients $\eta$ varies from 1 to 5 along x-axis in each sub-figure.}
\label{fig:defense_sor}
\end{figure*}

\begin{figure*}[!t] 
\centering
\subfigure[$\alpha$=0.7] 
{
    \label{defense_sor_alpha0.7}
	\centering         
	\includegraphics[width=0.32\textwidth]{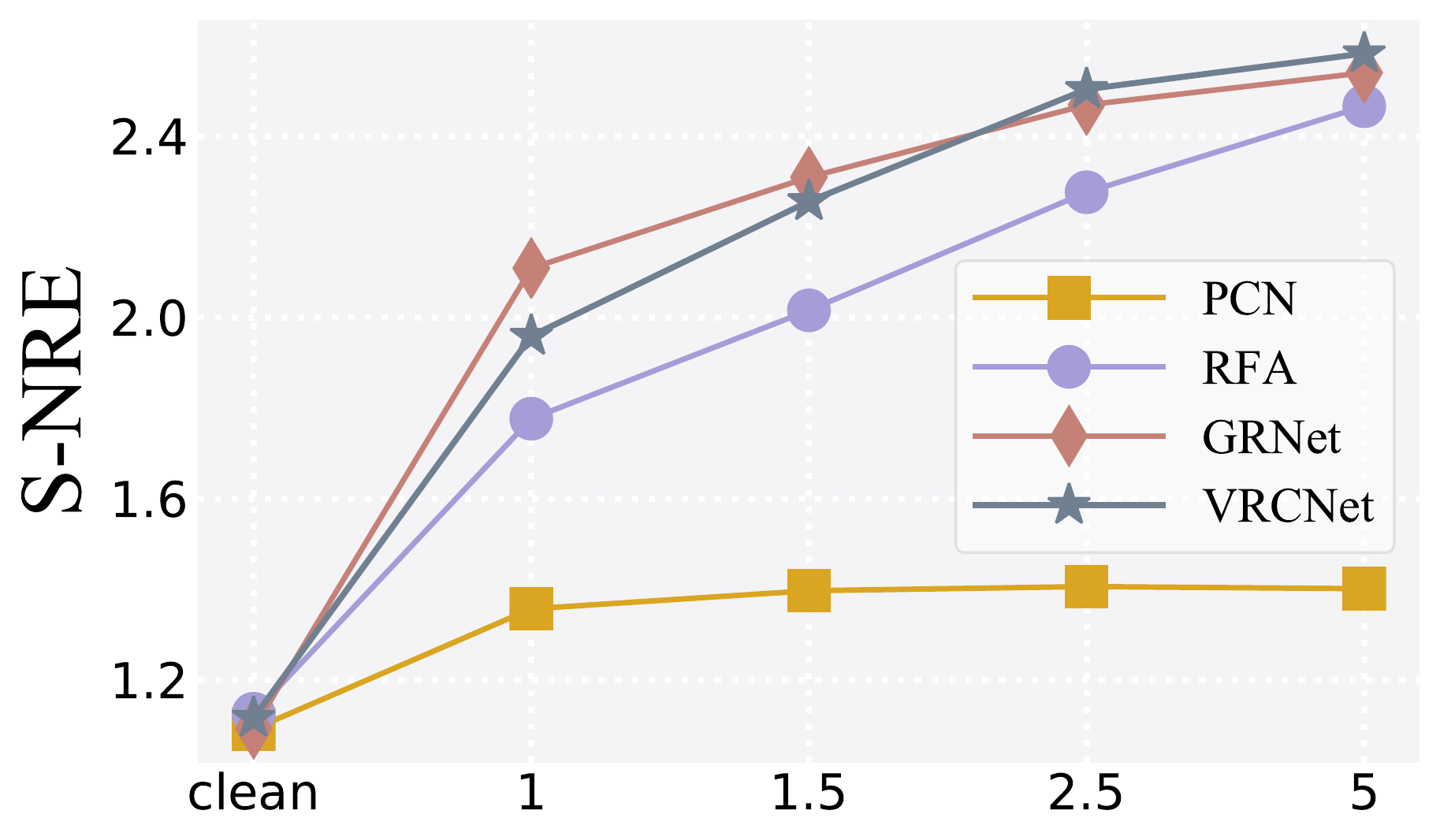}  
}
\subfigure[$\alpha$=1.1] 
{
    \label{defense_sor_alpha1.1}
	\centering     
	\includegraphics[width=0.32\textwidth]{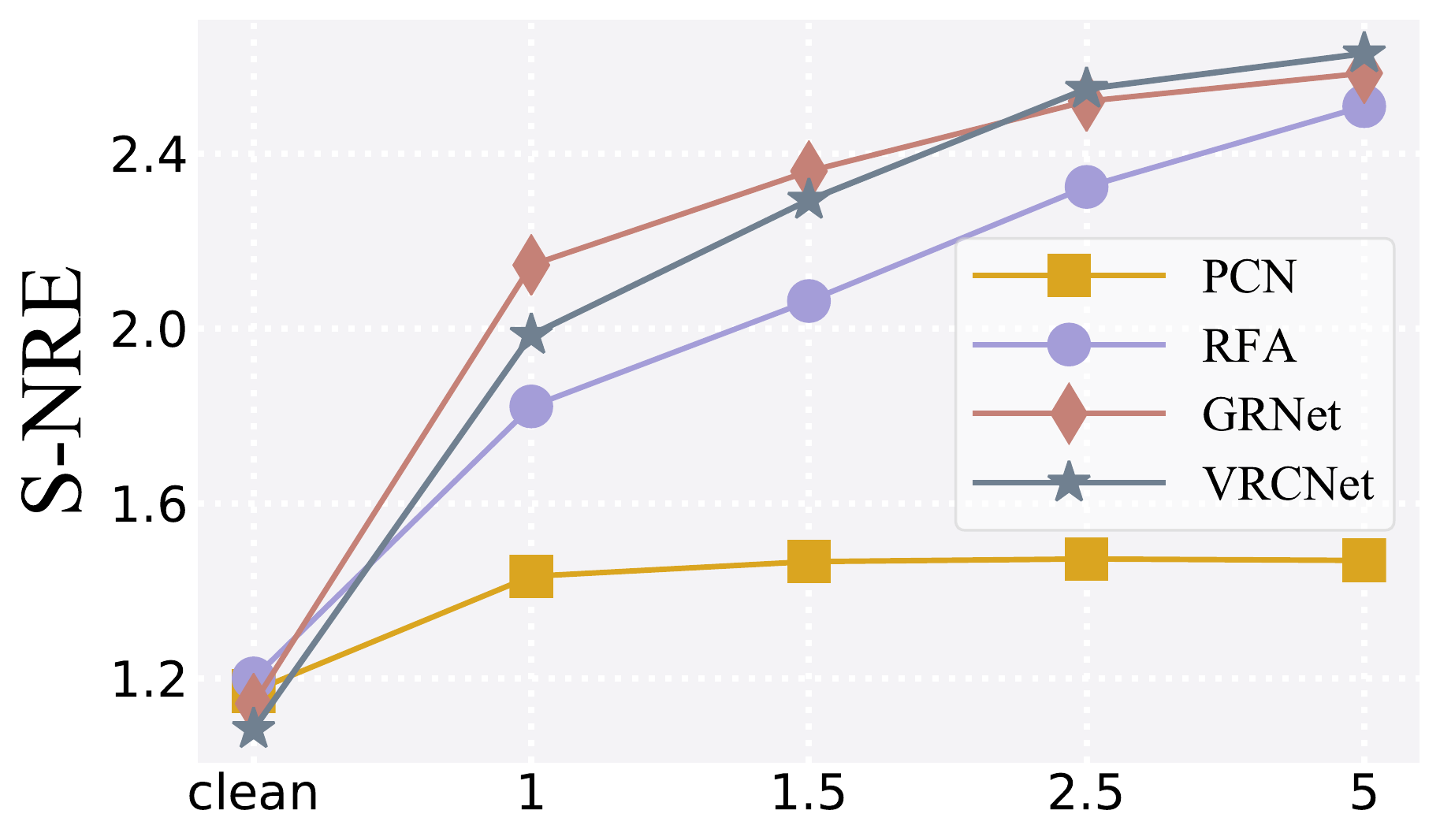}   
}
\subfigure[$\alpha$=1.5] 
{
    \label{defense_sor_alpha1.5}
	\centering     
	\includegraphics[width=0.32\textwidth]{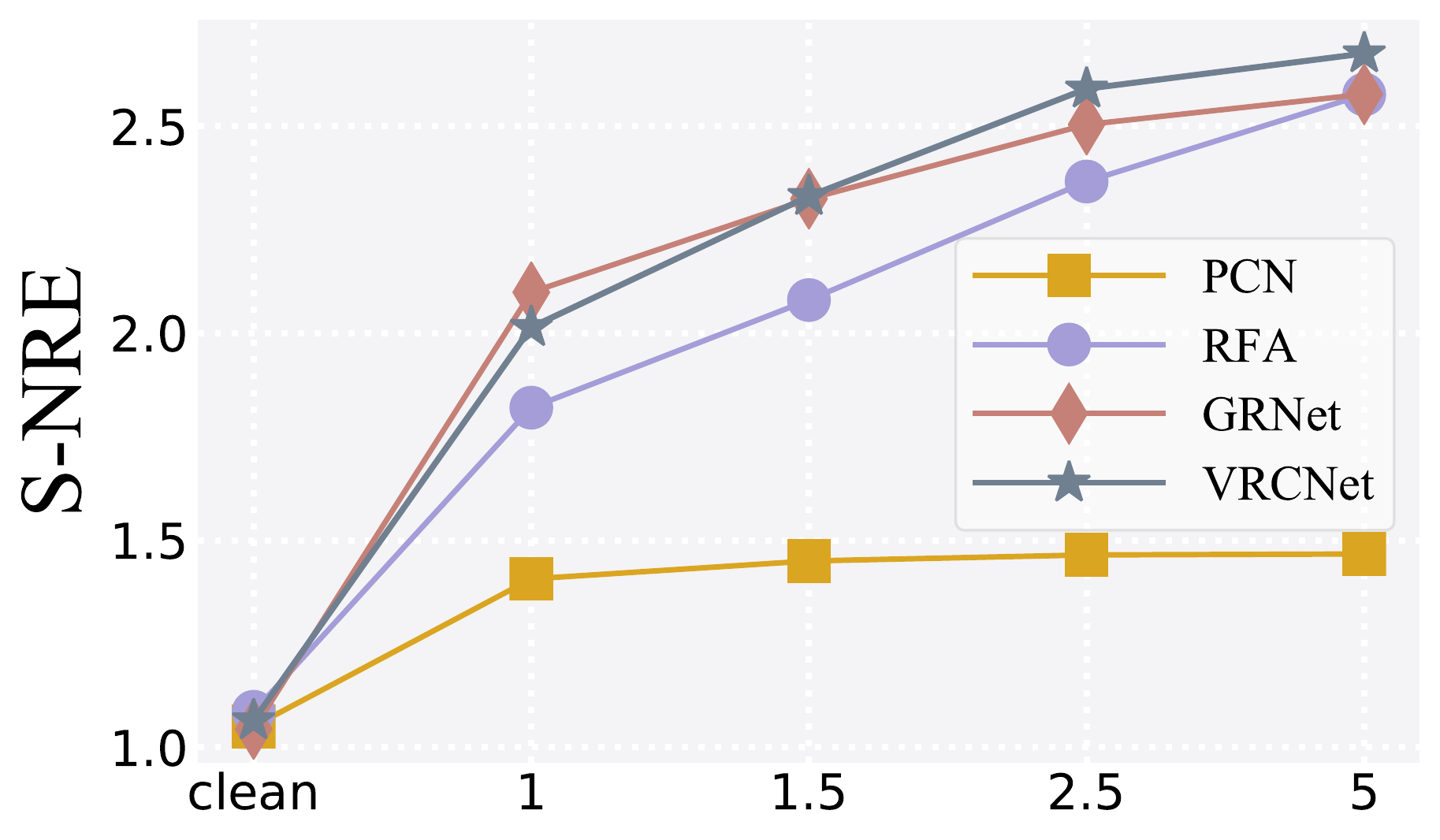}   
}
\vspace{-0.8em}
\caption{Comparison against the SOR defense. The defense interval parameter $\alpha$ varies from 0.7 to 1.5 from left to right, and scaling coefficients $\eta$ varies from 1 to 5 along x-axis in each sub-figure.}
\label{fig:defense_sor2}
\end{figure*}

\noindent\textbf{Analysis.} 
Fig.~\ref{fig:defense_srs} shows the results under different Drop rate settings in SRS defense. Fig.~\ref{fig:defense_os} shows the results under different Threshold in OR defense. Fig.~\ref{fig:defense_sor} and Fig.~\ref{fig:defense_sor2} report the results under different defense neighborhood size $K$ and different interval parameter $\alpha$ in SOR. We can observe that the SOR approach achieves the best outcome, whereas the SRS method has the weakest defense performance. Despite being tested with various defense configurations, our approach still results in incredibly large error for the majority of models, demonstrating the dependability of our approach.

\subsection{B.4 Transferability of PointCA}
\textbf{Implementation details.} The attack ability of adversarial point clouds across multiple models is another aspect worth considering. Under the black box setting, most existing adversaries often have a poor transferable performance to attack other models. We investigate the transferability of PointCA through feeding the target victim model with adversarial examples generated on other two source models. 

\begin{figure}[t]
    \setlength{\belowcaptionskip}{-0.5cm}   
    \centering
    \includegraphics[width=0.42\textwidth]{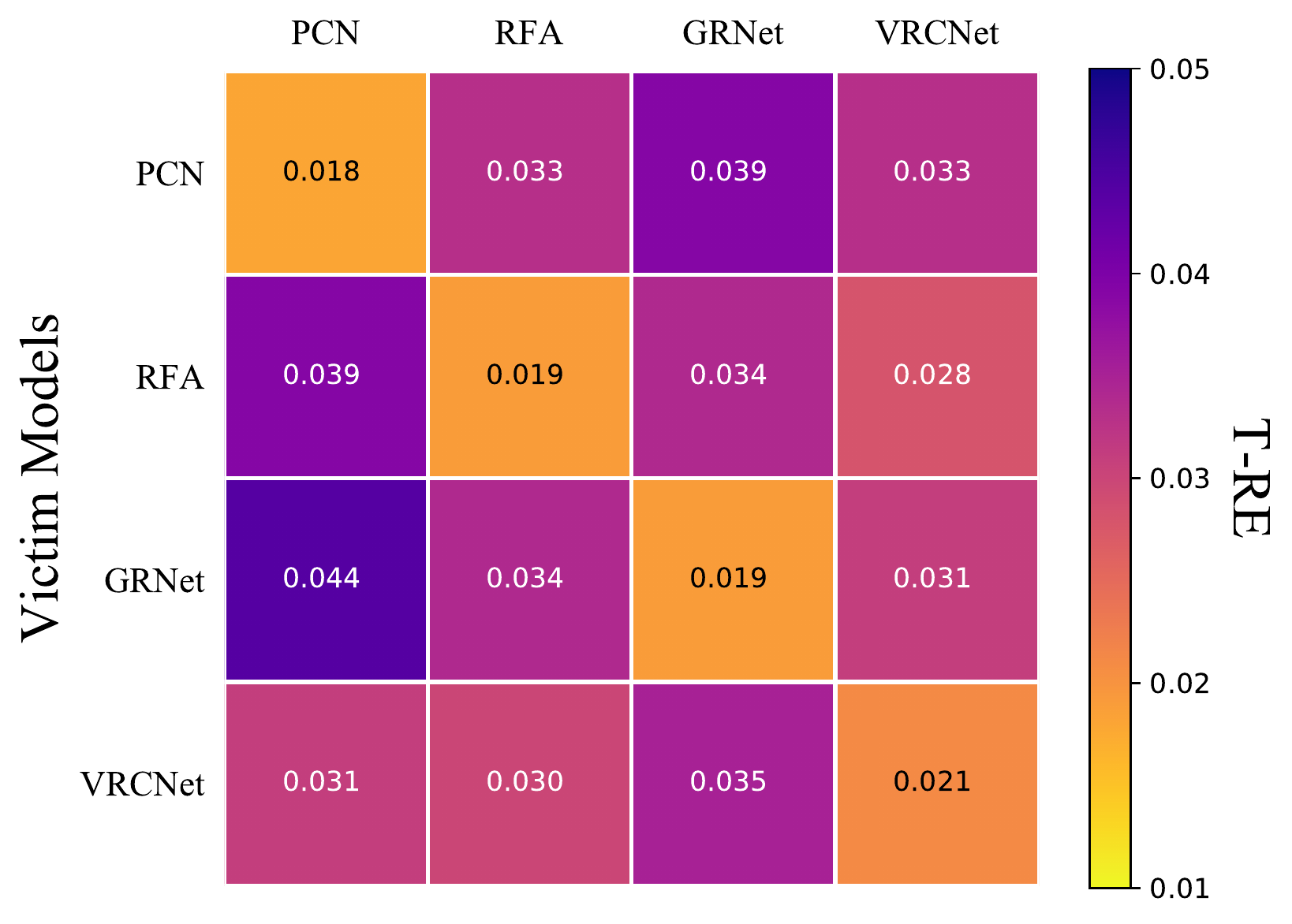}
    \vspace{-1em}
     \caption{Co-occurrence matrix for evaluating the transferability of PointCA on four target models. A smaller value of T-RE indicates a better attack effect. The network exploited in generating the adversarial examples is represented by elements in the same column, while the victim network which the attack is transferred to is represented by elements in the same row.}
    \label{fig:transfer_geo}
\end{figure}

\noindent\textbf{Analysis.} We use the \textit{target reconstruction error} (T-RE) to measure the transferable attack effect. As shown in Fig.~\ref{fig:transfer_geo}, most of the adversarial point clouds can have a certain transferable effect. 

\begin{table*}[t]
\centering\Large
\renewcommand\arraystretch{1.2}
\resizebox{0.95\textwidth}{!}{%
\begin{tabular}{cccccccccc}
\toprule[1.3pt]
\multicolumn{2}{c}{\textbf{Model}}    & \multicolumn{4}{c}{\textbf{PCN}}     & \multicolumn{4}{c}{\textbf{RFA}}  \\ 
\cmidrule(r){3-6} \cmidrule(r){7-10} 
\multicolumn{1}{c}{\textbf{$\eta$}} & t   & Outliers$(\downarrow)$  & $\text{T-RE}_{emd}(\downarrow)$ & $\text{T-NRE}_{cd}(\downarrow)$ & $\text{T-NRE}_{cd}(\downarrow)$ & Outliers$(\downarrow)$  & $\text{T-RE}_{emd}(\downarrow)$ & $\text{T-NRE}_{cd}(\downarrow)$ & $\text{T-NRE}_{cd}(\downarrow)$ \\ \hline
\multicolumn{1}{c}{\multirow{5}{*}{2.5}} & 0 & 79.4753 & 0.1296 & 0.0254 & 1.9881 & 93.5241 & 0.1608 & 0.0268 & 2.4361 \\
\multicolumn{1}{c}{}  & 1.5 & 71.8733 & 0.1221  & 0.0214 & 1.6672  & 92.5507 & 0.1566  & 0.0229 & 2.0841  \\
\multicolumn{1}{c}{}  & 3   & 68.3293 & 0.1186  & 0.0195 & 1.5173  & 91.4828 & 0.1539  & 0.0211 & 1.9159  \\
\multicolumn{1}{c}{}  & 4.5 & 66.5916 & 0.1168  & 0.0186 & 1.4438  & 90.4177 & 0.1523  & 0.0201 & 1.8271  \\
\multicolumn{1}{c}{}  & 6   & \textbf{65.5920} & \textbf{0.1158}  & \textbf{0.0180} & \textbf{1.3991}  & \textbf{89.6543} & \textbf{0.1513}  & \textbf{0.0196} & \textbf{1.7819}  \\ \hline
\multicolumn{1}{c}{\multirow{5}{*}{5}}   & 0 & 68.5820 & 0.1190 & 0.0197 & 1.5330 & 91.6453 & 0.1539 & 0.0212 & 1.9319 \\
\multicolumn{1}{c}{}  & 1.5 & 65.5515 & 0.1158  & 0.0180 & 1.4003  & 89.7653 & 0.1514  & 0.0197 & 1.7877  \\
\multicolumn{1}{c}{}  & 3   & 64.6125 & 0.1148  & 0.0175 & 1.3551  & 88.0403 & 0.1505  & 0.0192 & 1.7430  \\
\multicolumn{1}{c}{}  & 4.5 & 64.2310 & 0.1145  & 0.0172 & 1.3380  & 86.6126 & 0.1499  & 0.0191 & 1.7308  \\
\multicolumn{1}{c}{}  & 6   & \textbf{64.1041} & \textbf{0.1143}  & \textbf{0.0171} & \textbf{1.3295}  & \textbf{85.6540} & \textbf{0.1498}  & \textbf{0.0190} & \textbf{1.7240}  \\ 
\bottomrule[1.3pt]
\end{tabular}%
}
\caption{The attack performance with different auxiliary parameters $t$ in adaptive perturbation constraint. Lower values $(\downarrow)$ indicate better attack strength. \textbf{$\eta$} is a scaling coefficient to set a flexible perturbation limitation.}
\label{tab:uniformity}
\vspace{-1em}
\end{table*}

\subsection{B.5 Ablation Study on Neighborhood Size}
\textbf{Implementation details.} Our adaptive perturbation constraint is based on the local neighborhood partition, so the neighborhood size $k$ plays a key role for the proposed attack.
For the sake of clarifying the internal mechanism of the design, we explore four kinds of neighborhood size $k$ with three scaling coefficients $\eta$ on PCN and RFA. 

\begin{figure}[h] 
\setlength{\belowcaptionskip}{-0.5cm}
\centering
\subfigure[PCN] 
{
    \label{histogram_pcn}
	\centering         
	\includegraphics[width=0.22\textwidth]{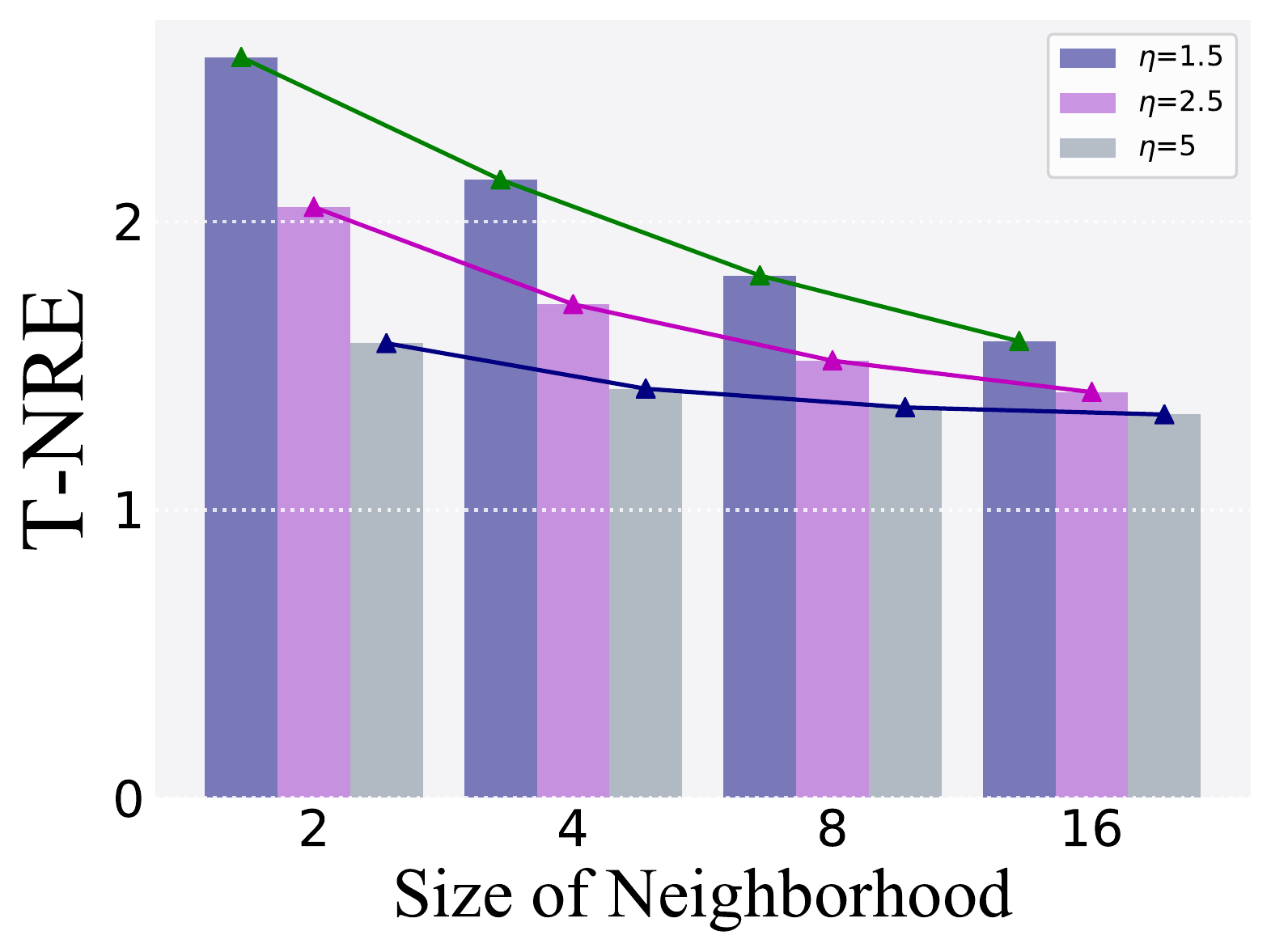}  
}
\subfigure[RFA] 
{
    \label{histogram_rfa}
	\centering     
	\includegraphics[width=0.22\textwidth]{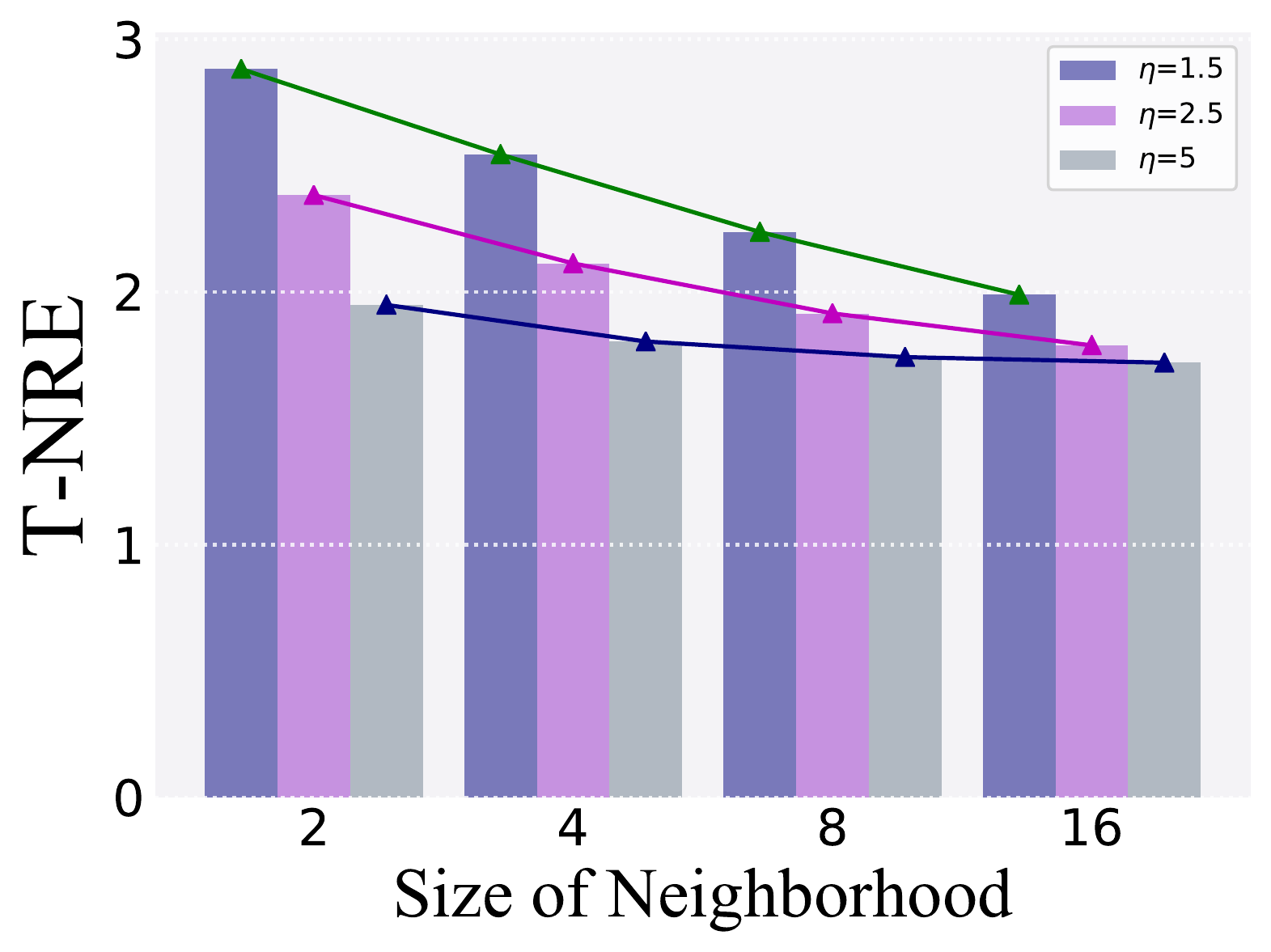}   
}
\vspace{-1em}
\caption{The attack performance with different local neighborhood sizes in adaptive perturbation constraint.}
\label{fig:Neighborhood Size}
\end{figure}

\noindent\textbf{Analysis.}
In Fig.~\ref{fig:Neighborhood Size}, we find that the attack performance can be improved through expanding the size of the neighborhood points set, which is apparent under the settings of $k = \{2,4,8\}$. Whereas, the benefit of using the larger neighborhood seems slow down when setting the size as $k=16$. It shows that the local neighborhood point sets of 8 points are sufficient to explore the local geometric distribution information on the partial point clouds. Therefore, we select the setting of $k$=8 in our experiments to obtain a balance between the computational efficiency and attack quality.

\subsection{B.6 Ablation Study on Uniformity and Sparsity}
\begin{figure}[h] 
\setlength{\belowcaptionskip}{-0.15cm}  
\centering
\subfigure[${d}_{i} = {d}_{j},{\sigma}_{i} \neq {\sigma}_{j}$] 
{
    \label{uniformity}
	\centering         
	\includegraphics[width=0.38\textwidth]{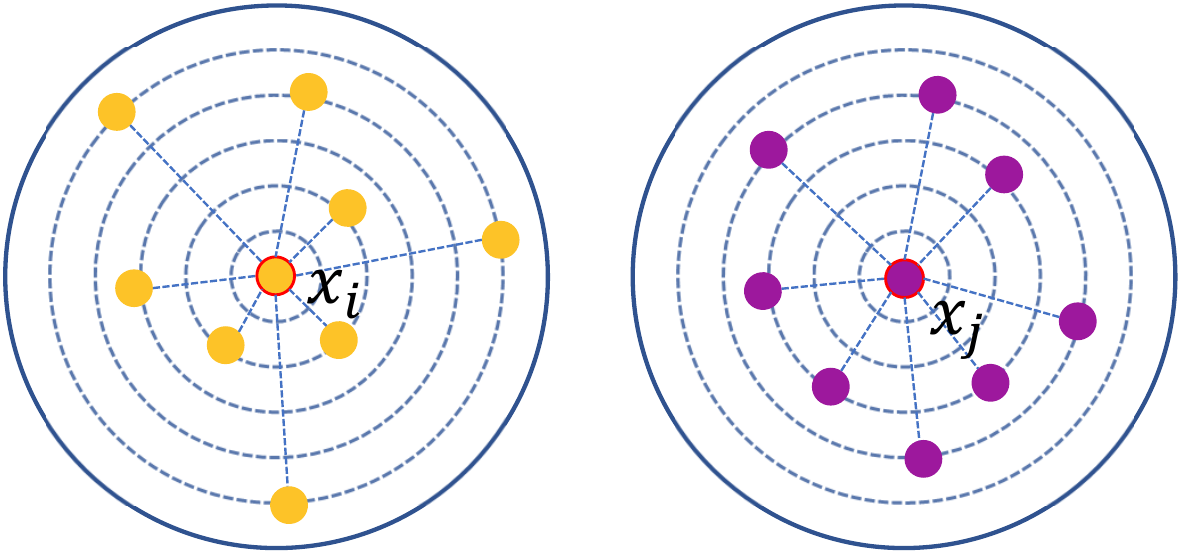}  
}
\subfigure[${d}_{i} \neq {d}_{j},{\sigma}_{i} = {\sigma}_{j}$] 
{
    \label{sparsrity}
	\centering     
	\includegraphics[width=0.38\textwidth]{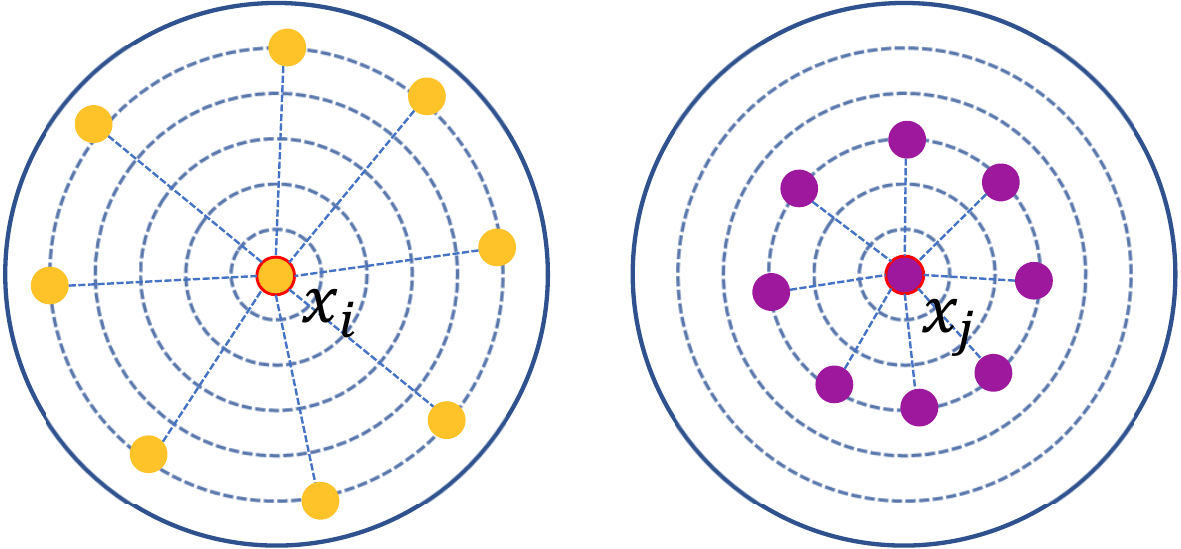}   
}
\vspace{-1em}
\caption{A comparison of uniformity and sparsity in points local neighborhood}
\label{fig:Uniformity and Sparsity}
\end{figure}

\textbf{Implementation details.} 
The density distribution description $\rho_{i}$ for each point contains two elements:
\textit{distribution sparsity} ${d}_{i}$ and \textit{distribution uniformity} $\sigma_{i}$:
\begin{equation}
\begin{aligned}
\rho_{i} =\mathit{d}_{i} + t \cdot \sigma_{i}.
\end{aligned}
\end{equation}
We use an auxiliary parameter $t$ in our adaptive perturbation constraint to balance them. The auxiliary parameter $t$ is empirically set to 3 in all attacks in the main paper. In Fig.~\ref{fig:Uniformity and Sparsity}, We show two cases of sparsity versus uniformity. The point cloud local neighborhood may have the same sparsity but different uniformity. Conversely, the local neighborhoods of a point cloud may be homogeneous but sparse in distinct ways. To explore the specific relative contribution in our adaptive geometric constraint $\epsilon_{i}$ of these two metrics, we evaluate the attack performance under different auxiliary parameters $t$.

\noindent\textbf{Analysis.} 
We can find in Table~\ref{tab:uniformity} that uniformity information can slightly revise the generation process of adversarial point clouds. Nevertheless, it is less important than the sparsity information when increasing the scaling coefficient $\eta$.

\subsection{B.7 Ablation Study on Latent Similarity Loss}
\textbf{Implementation details.} The attack effect of Latent PointCA greatly depends on how effectively we can compare two sets of features in the hidden layer space. Considering prior work, we select the $\ell_{2}$-norm $\left\| \cdot \right\|_{2}$~\cite{kos2018adversarial} and Kullback–Leibler divergence $D_{KL}(\cdot\|\cdot)$~\cite{willetts2021improving} as our latent similarity metric losses. We investigate the attack performances under the different metrics with the consistent learning rate setting to choose the most effective strategy and enhance the effect our Latent PointCA. 

\noindent\textbf{Analysis.} 
Table~\ref{tab:latent_loss} shows the results of different component combinations. It is observed that both KL divergence $D_{KL}(\cdot\|\cdot)$ and $\ell_{2}$-norm $\left\| \cdot \right\|_{2}$ can get the satisfactory results. However, due to the great variance in the magnitude of the hidden layer features of different models, the adjustment of parameters such as learning rate is more complicated and difficult. To reduce the influence of magnitude disparities from the standpoint of distribution similarity, we try to jointly use these two metrics. The best performance is obtained when they are combined, implying that the two components contribute to the improvement of the feature similarity measure in a complimentary way.
\vspace{-0.5em}
\begin{table}[!h]
\renewcommand\arraystretch{1.2}
\resizebox{\columnwidth}{!}{%
\begin{tabular}{cccccc}
\toprule[1.3pt]
KL Divergence & $\ell_{2}$-norm & PCN & RFA & GRNet & VRCNet \\ \hline
\checkmark &  & 1.845 &  2.655 & 4.213 & 1.974\\ 
 & \checkmark & 1.748 & 2.650 & 3.960 & 1.951 \\
\checkmark & \checkmark & \textbf{1.745} & \textbf{2.649} & \textbf{3.958} & \textbf{1.943} \\
\bottomrule[1.3pt]
\end{tabular}%
}
\caption{T-NRE for our Latent PointCA with different similarity metrics. A smaller value means a better attack effect. '\checkmark' indicates that the component is used. }
\label{tab:latent_loss}
\vspace{-0.5em}
\end{table}

\section{C \quad More Visualization Results}
\label{sec:visualization}

\begin{figure*}[htb]
\subfigure[Targeted attack results on RFA]
{
    \centering
    \includegraphics[width=0.92\textwidth]{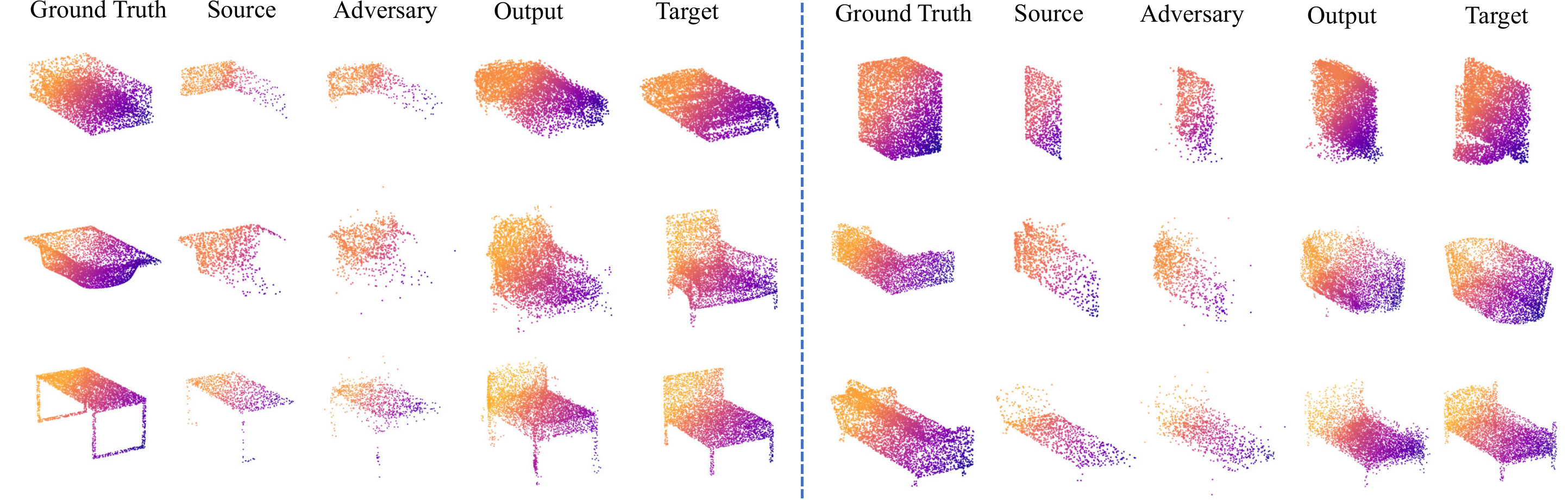}
    \label{fig:rfa_visual}
}

\vspace{10mm}

\subfigure[Targeted attack results on GRNet]
{
    \centering
    \includegraphics[width=0.94\textwidth]{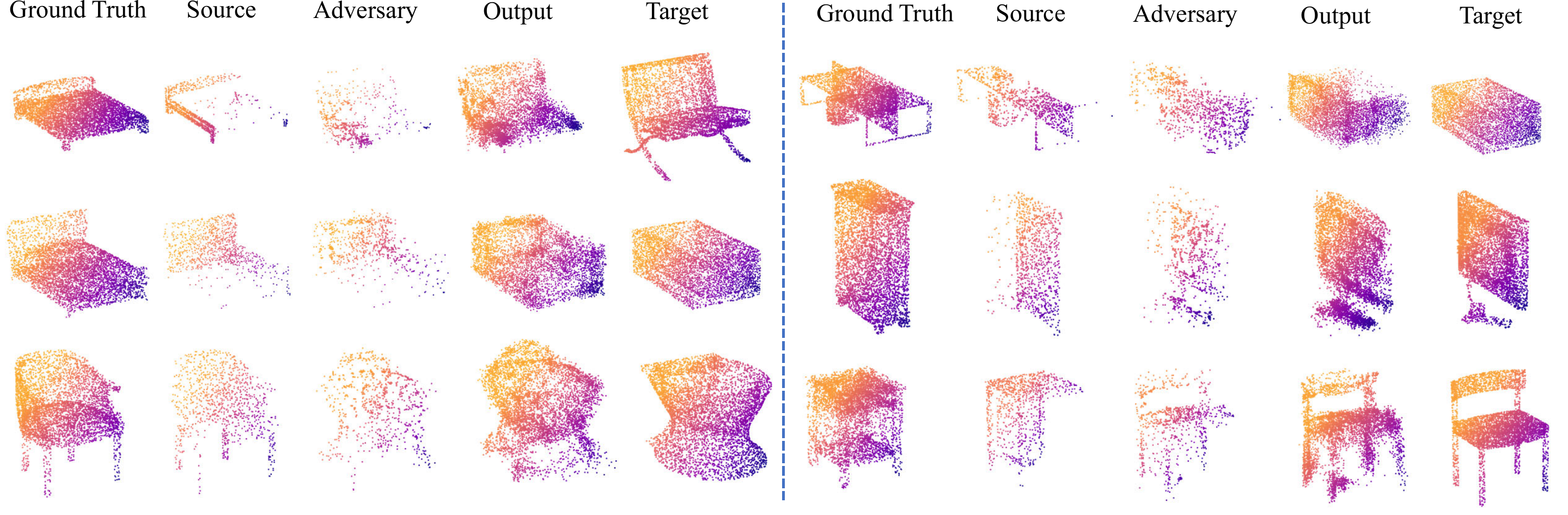}
    \label{fig:grnet_visual}
}

\vspace{10mm}

\subfigure[Targeted attack results on VRCNet]
{
    \centering
    \includegraphics[width=0.96\textwidth]{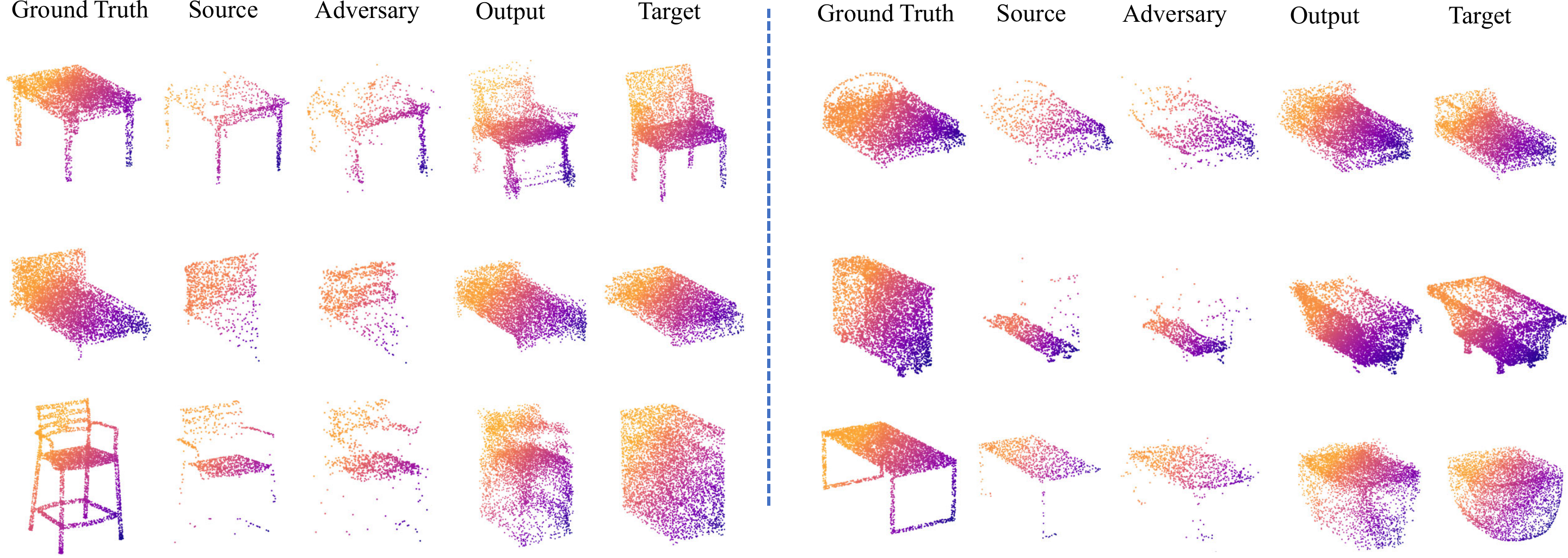}
    \label{fig:vrcnet_visual}
}

\caption{More visualization results of targeted attacks on RFA, GRNet and VRCNet. \textit{Source} is the partial point cloud generated on \textit{Ground Truth} from one view point. Tiny perturbations are added to \textit{Source} to obtain the \textit{Adversary}, whose completion \textit{Output} is very close to the target.}
\label{fig:visual}
\end{figure*}

More visualization results of targeted attacks on RFA, GRNet and VRCNet are exhibited in Fig.~\ref{fig:visual}.

\vspace{6mm}
\vspace{6mm}
\vspace{6mm}

\vspace{10mm}

\FloatBarrier
\bibliography{aaai23}

\begin{thebibliography}{37}
\providecommand{\natexlab}[1]{#1}

\bibitem[{Carlini and Wagner(2017)}]{carlini2017towards}
Carlini, N.; and Wagner, D. 2017.
\newblock Towards evaluating the robustness of neural networks.
\newblock In \emph{Proceedings of the 2017 IEEE Symposium on Security and
  Privacy (SP'17)}, 39--57.

\bibitem[{Dai et~al.(2017)Dai, Chang, Savva, Halber, Funkhouser, and
  Nie{\ss}ner}]{dai2017scannet}
Dai, A.; Chang, A.~X.; Savva, M.; Halber, M.; Funkhouser, T.; and Nie{\ss}ner,
  M. 2017.
\newblock Scannet: Richly-annotated 3d reconstructions of indoor scenes.
\newblock In \emph{Proceedings of the 2017 IEEE/CVF Conference on Computer
  Vision and Pattern Recognition (CVPR'17)}, 5828--5839.

\bibitem[{Fan, Su, and Guibas(2017)}]{fan2017point}
Fan, H.; Su, H.; and Guibas, L.~J. 2017.
\newblock A point set generation network for 3d object reconstruction from a
  single image.
\newblock In \emph{Proceedings of the 2017 IEEE/CVF Conference on Computer
  Vision and Pattern Recognition (CVPR'17)}, 605--613.

\bibitem[{Goodfellow, Shlens, and Szegedy(2015)}]{goodfellow2014explaining}
Goodfellow, I.~J.; Shlens, J.; and Szegedy, C. 2015.
\newblock Explaining and harnessing adversarial examples.
\newblock In \emph{Proceedings of the 3rd International Conference on Learning
  Representations (ICLR'15)}.

\bibitem[{Hamdi et~al.(2020)Hamdi, Rojas, Thabet, and Ghanem}]{hamdi2020advpc}
Hamdi, A.; Rojas, S.; Thabet, A.; and Ghanem, B. 2020.
\newblock Advpc: Transferable adversarial perturbations on 3D point clouds.
\newblock In \emph{Proceedings of the 16th European Conference on Computer
  Vision (ECCV'20)}, 241--257.

\bibitem[{He et~al.(2016)He, Zhang, Ren, and Sun}]{he2016deep}
He, K.; Zhang, X.; Ren, S.; and Sun, J. 2016.
\newblock Deep residual learning for image recognition.
\newblock In \emph{Proceedings of the 2016 IEEE/CVF Conference on Computer
  Vision and Pattern Recognition (CVPR'16)}, 770--778.

\bibitem[{Hu et~al.(2022{\natexlab{a}})Hu, Liu, Zhang, Li, Zhang, Jin, and
  Wu}]{hu2022protecting}
Hu, S.; Liu, X.; Zhang, Y.; Li, M.; Zhang, L.~Y.; Jin, H.; and Wu, L.
  2022{\natexlab{a}}.
\newblock Protecting Facial Privacy: Generating Adversarial Identity Masks via
  Style-robust Makeup Transfer.
\newblock In \emph{Proceedings of the 2022 IEEE/CVF Conference on Computer
  Vision and Pattern Recognition (CVPR'22)}, 15014--15023.

\bibitem[{Hu et~al.(2022{\natexlab{b}})Hu, Zhou, Zhang, Zhang, Zheng, He, and
  Jin}]{hu2022Badhash}
Hu, S.; Zhou, Z.; Zhang, Y.; Zhang, L.~Y.; Zheng, Y.; He, Y.; and Jin, H.
  2022{\natexlab{b}}.
\newblock BadHash: Invisible Backdoor Attacks against Deep Hashing with Clean
  Label.
\newblock In \emph{Proceedings of the 30th ACM International Conference on
  Multimedia (MM'22)}, 678--686.

\bibitem[{Huang et~al.(2022)Huang, Dong, Chen, Zhou, Zhang, and
  Yu}]{huang2022shape}
Huang, Q.; Dong, X.; Chen, D.; Zhou, H.; Zhang, W.; and Yu, N. 2022.
\newblock Shape-invariant 3D Adversarial Point Clouds.
\newblock In \emph{Proceedings of the 2022 IEEE/CVF Conference on Computer
  Vision and Pattern Recognition (CVPR'22)}, 15335--15344.

\bibitem[{Kim et~al.(2021)Kim, Hua, Nguyen, and Yeung}]{kim2021minimal}
Kim, J.; Hua, B.-S.; Nguyen, T.; and Yeung, S.-K. 2021.
\newblock Minimal adversarial examples for deep learning on 3D point clouds.
\newblock In \emph{Proceedings of the 2021 IEEE/CVF International Conference on
  Computer Vision (ICCV'21)}, 7797--7806.

\bibitem[{Kos, Fischer, and Song(2018)}]{kos2018adversarial}
Kos, J.; Fischer, I.; and Song, D. 2018.
\newblock Adversarial examples for generative models.
\newblock In \emph{Proceedings of the 2018 IEEE Symposium on Security and
  Privacy Workshops (SPW'18)}, 36--42.

\bibitem[{Lang, Kotlicki, and Avidan(2021)}]{lang2021geometric}
Lang, I.; Kotlicki, U.; and Avidan, S. 2021.
\newblock Geometric adversarial attacks and defenses on 3D point clouds.
\newblock In \emph{Proceedings of the 2021 International Conference on 3D
  Vision (3DV'21)}, 1196--1205.

\bibitem[{Lang, Manor, and Avidan(2020)}]{lang2020samplenet}
Lang, I.; Manor, A.; and Avidan, S. 2020.
\newblock Samplenet: Differentiable point cloud sampling.
\newblock In \emph{Proceedings of the 2020 IEEE/CVF Conference on Computer
  Vision and Pattern Recognition (CVPR'20)}, 7578--7588.

\bibitem[{Liu, Yu, and Su(2019)}]{liu2019extending}
Liu, D.; Yu, R.; and Su, H. 2019.
\newblock Extending adversarial attacks and defenses to deep 3D point cloud
  classifiers.
\newblock In \emph{Proceedings of the 2019 IEEE International Conference on
  Image Processing (ICIP'19)}, 2279--2283.

\bibitem[{Liu, Jia, and Gong(2021)}]{liu2021pointguard}
Liu, H.; Jia, J.; and Gong, N.~Z. 2021.
\newblock Pointguard: Provably robust 3D point cloud classification.
\newblock In \emph{Proceedings of the 2021 IEEE/CVF Conference on Computer
  Vision and Pattern Recognition (CVPR'21)}, 6186--6195.

\bibitem[{Ma et~al.(2020)Ma, Meng, Wu, Xu, and Zhang}]{ma2020efficient}
Ma, C.; Meng, W.; Wu, B.; Xu, S.; and Zhang, X. 2020.
\newblock Efficient joint gradient based attack against SOR defense for 3D
  point cloud classification.
\newblock In \emph{Proceedings of the 28th ACM International Conference on
  Multimedia (MM'20)}, 1819--1827.

\bibitem[{Pan et~al.(2021)Pan, Chen, Cai, Zhang, Zhao, Yi, and
  Liu}]{pan2021variational}
Pan, L.; Chen, X.; Cai, Z.; Zhang, J.; Zhao, H.; Yi, S.; and Liu, Z. 2021.
\newblock Variational relational point completion network.
\newblock In \emph{Proceedings of the 2021 IEEE/CVF Conference on Computer
  Vision and Pattern Recognition (CVPR'21)}, 8524--8533.

\bibitem[{Ren, Pan, and Liu(2022)}]{ren2022benchmarking}
Ren, J.; Pan, L.; and Liu, Z. 2022.
\newblock Benchmarking and analyzing point cloud classification under
  corruptions.
\newblock \emph{arXiv preprint arXiv:2202.03377}.

\bibitem[{Rubner, Tomasi, and Guibas(2000)}]{rubner2000earth}
Rubner, Y.; Tomasi, C.; and Guibas, L.~J. 2000.
\newblock The earth mover's distance as a metric for image retrieval.
\newblock \emph{International Journal of Computer Vision}, 40(2): 99--121.

\bibitem[{Shi et~al.(2022)Shi, Chen, Xu, Yang, Yu, and Huang}]{ShiCX0YH22}
Shi, Z.; Chen, Z.; Xu, Z.; Yang, W.; Yu, Z.; and Huang, L. 2022.
\newblock Shape Prior Guided Attack: Sparser Perturbations on 3D Point Clouds.
\newblock In \emph{Proceedings of the Thirty-Sixth AAAI Conference on
  Artificial Intelligence (AAAI'22)}, 8277--8285.

\bibitem[{Sun et~al.(2022)Sun, Zhang, Kailkhura, Yu, Xiao, and
  Mao}]{sun2022benchmarking}
Sun, J.; Zhang, Q.; Kailkhura, B.; Yu, Z.; Xiao, C.; and Mao, Z.~M. 2022.
\newblock Benchmarking Robustness of 3D Point Cloud Recognition Against Common
  Corruptions.
\newblock \emph{arXiv preprint arXiv:2201.12296}.

\bibitem[{Sun et~al.(2021)Sun, Chen, Chen, and Wang}]{sun2021local}
Sun, Y.; Chen, F.; Chen, Z.; and Wang, M. 2021.
\newblock Local Aggressive Adversarial Attacks on 3D Point Cloud.
\newblock In \emph{Proceedings of the 2021 Asian Conference on Machine Learning
  (ACML'21)}, 65--80.

\bibitem[{Wang et~al.(2021)Wang, Liu, Yue, Lasenby, and
  Kusner}]{wang2021unsupervised}
Wang, H.; Liu, Q.; Yue, X.; Lasenby, J.; and Kusner, M.~J. 2021.
\newblock Unsupervised point cloud pre-training via occlusion completion.
\newblock In \emph{Proceedings of the 2021 IEEE/CVF International Conference on
  Computer Vision (ICCV'21)}, 9782--9792.

\bibitem[{Wang, Ang~Jr, and Lee(2020)}]{wang2020cascaded}
Wang, X.; Ang~Jr, M.~H.; and Lee, G.~H. 2020.
\newblock Cascaded refinement network for point cloud completion.
\newblock In \emph{Proceedings of the 2020 IEEE/CVF Conference on Computer
  Vision and Pattern Recognition (CVPR'20)}, 790--799.

\bibitem[{Willetts et~al.(2021)Willetts, Camuto, Rainforth, Roberts, and
  Holmes}]{willetts2021improving}
Willetts, M.; Camuto, A.; Rainforth, T.; Roberts, S.; and Holmes, C. 2021.
\newblock Improving VAEs' Robustness to Adversarial Attack.
\newblock In \emph{Proceedings of the 9th International Conference on Learning
  Representations (ICLR'21)}.

\bibitem[{Wu et~al.(2021)Wu, Pan, Zhang, Wang, Liu, and Lin}]{wu2021density}
Wu, T.; Pan, L.; Zhang, J.; Wang, T.; Liu, Z.; and Lin, D. 2021.
\newblock Balanced Chamfer Distance as a Comprehensive Metric for Point Cloud
  Completion.
\newblock In \emph{Proceedings of the 2021 Annual Conference on Neural
  Information Processing Systems (NeurIPS'21)}, 29088--29100.

\bibitem[{Wu et~al.(2015)Wu, Song, Khosla, Yu, Zhang, Tang, and
  Xiao}]{wu20153d}
Wu, Z.; Song, S.; Khosla, A.; Yu, F.; Zhang, L.; Tang, X.; and Xiao, J. 2015.
\newblock 3D shapenets: A deep representation for volumetric shapes.
\newblock In \emph{Proceedings of the 2015 IEEE/CVF Conference on Computer
  Vision and Pattern Recognition (CVPR'15)}, 1912--1920.

\bibitem[{Xiang, Qi, and Li(2019)}]{xiang2019generating}
Xiang, C.; Qi, C.~R.; and Li, B. 2019.
\newblock Generating 3D adversarial point clouds.
\newblock In \emph{Proceedings of the 2019 IEEE/CVF Conference on Computer
  Vision and Pattern Recognition (CVPR'19)}, 9136--9144.

\bibitem[{Xie et~al.(2021)Xie, Wang, Zhang, Yang, Chen, and Wen}]{xie2021style}
Xie, C.; Wang, C.; Zhang, B.; Yang, H.; Chen, D.; and Wen, F. 2021.
\newblock Style-based point generator with adversarial rendering for point
  cloud completion.
\newblock In \emph{Proceedings of the 2021 IEEE/CVF Conference on Computer
  Vision and Pattern Recognition (CVPR'21)}, 4619--4628.

\bibitem[{Xie et~al.(2020)Xie, Yao, Zhou, Mao, Zhang, and Sun}]{xie2020grnet}
Xie, H.; Yao, H.; Zhou, S.; Mao, J.; Zhang, S.; and Sun, W. 2020.
\newblock Grnet: Gridding residual network for dense point cloud completion.
\newblock In \emph{Proceedings of the 16th European Conference on Computer
  Vision (ECCV'20)}, 365--381.

\bibitem[{Yang et~al.(2019)Yang, Zhang, Fang, Ni, Liu, and
  Tian}]{yang2019adversarial}
Yang, J.; Zhang, Q.; Fang, R.; Ni, B.; Liu, J.; and Tian, Q. 2019.
\newblock Adversarial attack and defense on point sets.
\newblock \emph{arXiv preprint arXiv:1902.10899}.

\bibitem[{Yang et~al.(2018)Yang, Feng, Shen, and Tian}]{yang2018foldingnet}
Yang, Y.; Feng, C.; Shen, Y.; and Tian, D. 2018.
\newblock Foldingnet: Point cloud auto-encoder via deep grid deformation.
\newblock In \emph{Proceedings of the 2018 IEEE/CVF Conference on Computer
  Vision and Pattern Recognition (CVPR'18)}, 206--215.

\bibitem[{Yu et~al.(2021)Yu, Rao, Wang, Liu, Lu, and Zhou}]{yu2021pointr}
Yu, X.; Rao, Y.; Wang, Z.; Liu, Z.; Lu, J.; and Zhou, J. 2021.
\newblock Pointr: Diverse point cloud completion with geometry-aware
  transformers.
\newblock In \emph{Proceedings of the 2021 IEEE/CVF International Conference on
  Computer Vision (ICCV'21)}, 12498--12507.

\bibitem[{Yuan et~al.(2018)Yuan, Khot, Held, Mertz, and Hebert}]{yuan2018pcn}
Yuan, W.; Khot, T.; Held, D.; Mertz, C.; and Hebert, M. 2018.
\newblock PCN: Point Completion Network.
\newblock In \emph{Proceedings of the 2018 International Conference on 3D
  Vision (3DV'18)}, 728--737.

\bibitem[{Zhang, Yan, and Xiao(2020)}]{zhang2020detail}
Zhang, W.; Yan, Q.; and Xiao, C. 2020.
\newblock Detail preserved point cloud completion via separated feature
  aggregation.
\newblock In \emph{Proceedings of the 16th European Conference on Computer
  Vision (ECCV'20)}, 512--528.

\bibitem[{Zhou et~al.(2020)Zhou, Chen, Liao, Chen, Dong, Liu, Zhang, Hua, and
  Yu}]{zhou2020lg}
Zhou, H.; Chen, D.; Liao, J.; Chen, K.; Dong, X.; Liu, K.; Zhang, W.; Hua, G.;
  and Yu, N. 2020.
\newblock LG-GAN: Label Guided Adversarial Network for flexible targeted attack
  of point cloud based deep networks.
\newblock In \emph{Proceedings of the 2020 IEEE/CVF Conference on Computer
  Vision and Pattern Recognition (CVPR'20)}, 10356--10365.

\bibitem[{Zhou et~al.(2019)Zhou, Chen, Zhang, Fang, Zhou, and Yu}]{zhou2019dup}
Zhou, H.; Chen, K.; Zhang, W.; Fang, H.; Zhou, W.; and Yu, N. 2019.
\newblock DUP-Net: Denoiser and Upsampler Network for 3D adversarial point
  clouds defense.
\newblock In \emph{Proceedings of the 2019 IEEE/CVF International Conference on
  Computer Vision (ICCV'19)}, 1961--1970.

\end{thebibliography}
\end{document}